\documentclass{article}

\usepackage[main, final, nonatbib]{neurips_2026}

\usepackage[utf8]{inputenc} 
\usepackage[T1]{fontenc}    
\usepackage{hyperref}       
\usepackage{url}            
\usepackage{booktabs}       
\usepackage{amsfonts}       
\usepackage{nicefrac}       
\usepackage{microtype}      
\usepackage{xcolor}         
\usepackage{amsmath} 
\usepackage{graphicx}
\usepackage{comment}
\usepackage{subcaption}
\usepackage{amsthm}
\newtheorem{result}{Result}

\usepackage[sortcites=true,sorting=nyt,backend=biber, giveninits=true]{biblatex}

\title{Emergence of Distortions\\ in High-Dimensional Guided Diffusion Models}

\author{
Enrico Ventura\textsuperscript{1} \quad
Beatrice Achilli\textsuperscript{1} \quad
Luca Ambrogioni\textsuperscript{2} \quad
Carlo Lucibello\textsuperscript{1,3} \\
\\
\textsuperscript{1}Department of Computing Sciences, Bocconi University, Milan, Italy \\
\textsuperscript{2}Donders Institute for Brain, Cognition and behavior, Radboud University, Nijmegen, The Netherlands \\
\textsuperscript{3}Bocconi Institute for Data Science and Analytics, Bocconi University, Milan, Italy \\
}

\begin{document}

\maketitle

\begin{abstract}
Classifier-free guidance (CFG) is the de facto standard for conditional sampling in diffusion models, yet it often reduces sample diversity. Using tools from statistical physics, we analyze the emergence of generative distortions induced by CFG, namely the mismatch between the CFG sampling distribution and the true conditional distribution. We study this phenomenon in analytically tractable settings with exact score functions, characterizing its dependence on data dimensionality and the number of classes. For high-dimensional Gaussian mixtures, we use dynamic mean-field theory to show that distortions persist when the number of classes scales exponentially with the data dimension, whereas they vanish in the sub-exponential regime due to a dynamical phase transition. Moreover we found that, in the infinite-class limit, distortions remain unavoidable regardless of dimensionality because of the increasing density of classes. Finally, we show that standard CFG schedules cannot prevent variance shrinkage, and we propose a theoretically grounded guidance schedule incorporating a negative-guidance window that improves both class separability and sample diversity in real-world latent diffusion models.

\end{abstract}

\section{Introduction}

Generative diffusion models \cite{sohldickstein2015deep} are undoubtedly the most employed tools for generating images \cite{ho2020denoising, song2019generative, song_score-based_2021} and videos \cite{ho2022video, singer2022make, blattmann2023videoldm}. 
Diffusion models (DMs) synthesize images through a stochastic dynamical denoising process \cite{sohldickstein2015deep}. 
While unconditional DMs achieve remarkable sample quality, most practical applications require controllable generation, where samples are drawn from a conditional distribution specified by auxiliary information such as class labels or prompts. The consolidated, yet still unclear, ineffectiveness of direct use of trained conditional scores \cite{bradley2025, rombach2022high} motivated the adoption of guidance mechanisms to amplify the conditioning signal. 
Classifier-free guidance (CFG) has emerged as the standard approach for conditional sampling in DMs, due to its simplicity and empirical effectiveness \cite{ho_classifier-free_2022, jiao_unified2025}. By interpolating between unconditional and conditional score functions, CFG allows practitioners to control the strength of conditioning through a single parameter. Increasing this parameter typically improves class separability and alignment with the conditioning signal, but often induces a noticeable loss of diversity in the generated samples \cite{lu_whatCFGdoes2024, ho_classifier-free_2022, biroli_meta}.
Despite its widespread use, a principled understanding of how and why CFG modifies the target conditional distribution is still missing. In particular, it remains unclear whether the observed loss of sample diversity occurs in high dimensions, since recent work shows that in some settings, guidance-induced distortions only arise as a finite-dimensional effect \cite{biroli_meta}. In this work, we address this question by characterizing the notion of generative distortion in large dimensions and for a large number of classes using a statistical physics approach.

\subsection{Related Works} 

\textbf{Empirical Works and Guidance Schedules.} 
Early empirical analysis has highlighted that stronger CFG improves prompt-alignment but reduces diversity and overall perceptual quality \cite{ho_classifier-free_2022, saharia_photorealistic22}. Later work has tried to overcome pathological distortions by curbing spatial non-uniformity \cite{shen_rethinking2024}, including negative-prompt interactions \cite{ban_understanding, ambro_negative2025},
off-manifold drift \cite{chung_cfg++2025} and even employing a less-trained version of the unconditional model \cite{karras}. Nevertheless, many recent efforts focus on using a time modulation of guidance to improve the generation performance of CFG. Existing CFG schedules can be mostly grouped into four classes: \textit{early-high} / \textit{early-low} \cite{weight_schedulers2024, stage_wise}: respectively, higher / lower guidance at large diffusion times ; \textit{intermediate window} \cite{kynkaaniemi, stage_wise}: guidance turned on only at specific intermediate times; \textit{non-linear} \cite{biroli_meta}: guidance that varies non linearly in diffusion time. Recent work has also considered feedback forms of guidance, where the guiding signal depends on both time and current state \cite{koulischer2025feedback, mao2026spatial}.

\textbf{Analytical Works: Low and finite data dimension.}
Most of the analytical work dedicated to understanding the nature of CFG generative distortions consider low or finite-dimensional tractable data models.  \cite{fu_unveil2024} provide an approach based on statistical theory by approximating the guided score from bounds, on the line of \cite{chen_sampling_2022}. 
A closed-form analysis of a uni-modal Gaussian model in \cite{li_towards2025} has described how guidance shifts the mean of the conditional distribution along salient directions of the data. 
Other works have focused on mixtures of Gaussians and characterized how the data modes get deformed by guidance at different layers of complexity. For instance, \cite{bradley2025, lu_whatCFGdoes2024} study distortions after assuming that data are generated by a one-dimensional mixture of two Gaussians, while \cite{wu_mog2024} analyze the case where $d$ can be large but the number of Gaussian modes is finite and does not scale with the data dimension. Furthermore, \cite{li_provable2025} aims to treat general target distributions. Yet the analysis focuses on sample quality measured in a similar way as the Inception Score, and not on sample diversity or class separation. 
\\

\textbf{Analytical Works: Statistical physics in large dimension.}
Recent seminal works in the literature have employed tools from high-dimensional probability and statistical mechanics to provide insights about the performance of DMs under the assumption of high data dimension. For what concerns the sampling capabilities of these models, \cite{raya_spontaneous_2023, biroli_dynamical_2024, achilli2025memorization} have characterized the way DMs reconstruct the target distribution along the backward diffusion process in terms of a sequence of ergodicity-breaking events in the diffusion potential.   However, these works do not study the effects of these transitions on guided trajectories. 
The analysis in \cite{biroli_meta}, which also inspires a large part of our work, employs such tools to study the effects of guidance when data are generated by a mixture of two well-separated Gaussians in high dimensions. They prove that, in this setting, CFG cannot distort the data distribution. 

Appendix \ref{app:further} provides further detailed comparison between our contributions and the existing literature summarized above. 

\subsection{Our contributions}
The works reviewed above primarily focus on simple, low-dimensional or finite-dimensional settings. An exception is \cite{biroli_meta}, which demonstrates that such finite-dimensional effects become negligible in high-dimensional regimes when the number of classes remains fixed. This naturally raises the question of whether insights from low-dimensional studies remain relevant for understanding dynamics in more realistic, high-dimensional scenarios. In this paper, drawing on methods from statistical physics, we show that distortions in both the mean and variance persist in the high-dimensional regime when the number of classes is sufficiently large.

As novel contributions to the existing literature: 
\begin{enumerate}
\renewcommand{\labelenumi}{(\roman{enumi})}
    \item We analyze the case of an arbitrary number of separated classes in high dimensions, mapping the system into a so-called Random Energy Model (REM) and studying its dynamic phase transition through a mean-field approach. Our heuristic analysis shows that: distortions must asymptotically persist when classes are dense, i.e. their number scales exponentially with the data dimension; distortions vanish in the sub-exponential regime. We also analyze the limit of infinite continuous classes, and rigorously demonstrate that in this regime distortions are always present regardless the data dimension.  
    \item We heuristically show that vanilla CFG and existing guidance schedules inevitably reduce diversity, and we propose a new prescription that alternates positive and negative guidance. We extend our analytical framework to derive a distortion phase diagram, showing that our schedule can simultaneously enhance class separability and diversity. Finally, we validate the theoretical predictions on real datasets. 
    
\end{enumerate}

\section{Generative Diffusion and Classifier-free Guidance (CFG)}
In this paper, we will consider a variance-exploding \cite{song_score-based_2021} forward process where the data $\mathbf{x}_0 \sim p_{0}(\mathbf{x})$ evolves according to the equation $\text{d}\mathbf{x}_t= \text{d}\mathbf{W}_{t}$
where $\text{d}\mathbf{W}_{t}$ is standard Brownian motion in dimension $d$. The solutions of the previous equation have marginal density
\begin{equation}
p_t(\mathbf{x}_t) = \mathbb{E}_{\mathbf{x}_0\sim p_0}\left[\frac{1}{(2 \pi t)^{d/2}}e^{- \frac{\|\mathbf{x}_t - \mathbf{x}_0\|}{2t}^2}\right].
\end{equation}
The \emph{target distribution} $p_{0}(\mathbf{x})$ is then recovered by reversing the diffusion process \cite{anderson1982reverse}. We initialize this reverse (or \textit{backward}) process from $\mathbf{x}_{T} \sim\mathcal{N}\left(0,T\cdot I_d\right)$ at some large time $T$. The SDE running back in time used for generation reads
\begin{equation}
\label{eq:back}
\text{d}\mathbf{x}_t= -s_t(\mathbf{x}_t) \text{d}t+\text{d}\mathbf{W}_{t},
\end{equation}

where $s_t(\mathbf{x}_t)=\nabla_{\mathbf{x}}\log p_{t}(\mathbf{x}_t)$ is called \textit{score function}. From a set of training points $\{\mathbf{y}^1, \dots, \mathbf{y}^N \} \stackrel{\text{iid}}{\sim} p_0$, we can train a neural approximation of $s_t(\mathbf{x}_t)$ using the denoising score matching objective \cite{hyvarinen2005estimation, vincent_connection_2011, ho2020denoising}.

A crucial need for DM users is to sample conditionally on a given context, e.g. a text prompt for text-to-image generation. The most direct option is to train a model to fit the conditional score function $\nabla_{\mathbf{x}}\log p_t(\mathbf{x}_t|\mathbf{c})$ and run through that the reverse process. 
Neural approximations though, especially in the presence of a large conditioning space, often provide weak estimations and yield poor adherence to the conditioning and low quality samples. 
This motivated the use of \emph{guidance} methods that bias sampling toward higher conditional likelihood \cite{bradley2025, ho_classifier-free_2022}. Early work used \emph{classifier guidance} \cite{dhariwal2021diffusion}, adding the gradient of a noise-conditioned classifier $\nabla_{\mathbf{x}}\log p_\phi(\mathbf{c}| \mathbf{x}_t)$ to the vanilla diffusion score to steer samples. While effective, it requires a separate classifier and extra backpropagation at every step, increasing compute and making results sensitive to classifier errors and artifacts at high guidance strength.
Practitioners have then found the way to re-express the classifier score in terms of the original conditional score function $\nabla_{\mathbf{x}}\log p_t(\mathbf{x}_t|\mathbf{c})$ which involves no external classifier and labeling pre-process. They hence introduced a novel drift function that reads
\begin{equation}
\tilde{s}_{t}(\mathbf{x}_t|\mathbf{c})= (1+w)\nabla_{\mathbf{x}}\log p_{t}(\mathbf{x}_t|\mathbf{c})-w\nabla_{\mathbf{x}}\log p_{t}(\mathbf{x}_t),
\label{eq:CFG_score}
\end{equation}
where the \textit{guidance level} $w$ controls the degree
of conditioning: $w = -1$ reproduces unconditional diffusion, $w = 0$ reproduces conditional diffusion with no guidance, $w > 0$ reinforces conditioning. This method is called Classifier-free Guidance (CFG)
for generative diffusion  \cite{ho_classifier-free_2022}.

\section{Generative distortions in CFG}

We are interested in computing the deviation of the distribution
$\tilde{p}_{0}(\mathbf{x}|\mathbf{c})$ induced by CFG from the true target distribution $p_{0}(\mathbf{x}|\mathbf{c})$, assuming that we have access to the true score functions to derive the CFG score defined in Eq. \eqref{eq:CFG_score}.
For this purpose we choose tractable synthetic data models where all relevant distributions are Gaussian. 
Our choice is driven by the strong empirical evidence that trained DMs tend to approximate Gaussian kernel estimators during inference \cite{wang_vastola24, shah_learning23, pham_emergence25}. Moreover, we seek for a sufficiently interpretable framework allowing to compute generative distortions as a function of the number of classes and data dimension.
These distributional assumptions can be relaxed using techniques discussed in \cite{achili_speciation26}, but we will not consider this non-Gaussian extension here.

In Section \ref{sec:mix}, the data distribution is a mixture of $M$ Gaussians, with each component representing a class.
We identify two different scaling regimes of $M$ with respect to the data dimension $d$, and quantify the distortion in the limit $d \to \infty$. 
In our setting, where classes are identified by $\sqrt{d}$-norm vectors, $M$ must scale exponentially with data dimension for non-vanishing distortions to be measured. Our intention is to use this scaling as a proxy for the large amount of compositional text prompts we can condition on when generating from a latent diffusion model 
\cite{rombach2022high, zhang_prompt26, t2i_bench}. 
Besides this qualitative connection, we use the exponential scaling as an analytical artifact allowing Gaussian modes to be abundant and dense in the data space. Such a density of classes, which was not present in \cite{biroli_meta}, is responsible for distortions to occur in the high-dimensional limit.

In Section \ref{sec:cont} classes are continuous and jointly Gaussian distributed with the data: this setting corresponds to the $M \to \infty$ limit of the one analyzed in Section \ref{sec:mix}. 
The CFG sampling process can be exactly integrated and we can analytically quantify distortions, showing that they must appear independently of the data dimension.


\subsection{Separated Classes, High Dimensions \label{sec:mix}}
Let us consider the full target distribution 
\begin{equation}
\label{eq:uni_pdf}
    p_{0}(\mathbf{x}) = \frac{1}{M}\sum_{\mu = 1}^M \mathcal{N}(\mathbf{x};\mathbf{c}^{\mu}, \sigma^2),\quad \mathbf{c}^{\mu} \sim \mathcal{N}(0,I_d),
\end{equation}
that is an homogeneous mixture of  $M$ Gaussians
in a $d$ dimensional real space. 
In this setup, we are performing CFG by conditioning the process with respect to one class coinciding with one of the modes of the mixture. Specifically we choose $\mathbf{c} \equiv \mathbf{c}^1$, with $p_0(\mathbf{x}|\mathbf{c}^1) = \mathcal{N}(\mathbf{x};\mathbf{c}^{1}, \sigma^2)$. 
The evolving sampling distributions along a variance-exploding forward process read
\begin{equation}
    p_t\left(\mathbf{x}_t\right) = \frac{1}{M}\sum_{\mu=1}^M \mathcal{N}\left(\mathbf{x}_t; \mathbf{c}^{\mu},\sigma^2+t\right),\quad\quad p_t\left(\mathbf{x}_t | \mathbf{c}^1 \right) = \mathcal{N}\left(\mathbf{x}_t; \mathbf{c}^1,\sigma^2+t \right).
\end{equation}
Our goal consists again in finding the probability distribution $\tilde{p}_t(\mathbf{x}_t|\mathbf{c}^1)$ induced by the CFG process. For convenience of analysis, we rewrite the time-reversed SDE as
\begin{equation}
    \label{eq:SDE_pot}
    \text{d}\mathbf{x}_t = \nabla_{\mathbf{x}}V_{\text{eff}}(\mathbf{x}_t)\text{d}t + \text{d}\mathbf{W}_t.
\end{equation}
where we defined the effective time-dependent potential 
\begin{equation}
    V_{\text{eff}}(\mathbf{x}_t) =-\log\left[\frac{p_t(\mathbf{x}_t|\mathbf{c}^{1})^{1+w}}{p_t(\mathbf{x}_t)^w}\right]=V_{\text{cond}}(\mathbf{x}_t)+V_{\text{guided}}(\mathbf{x}_t)
\end{equation}
The potential has been decomposed into the pure conditional part
\begin{equation}
    V_{\text{cond}}\left(\mathbf{x}_t \right) = \frac{1}{2}\frac{\|\mathbf{x}_t-\mathbf{c}^1\|^2}{\sigma^2+t}+\frac{d}{2}\log(2\pi t),
\end{equation}
that is a quadratic potential centered on $\mathbf{c}^1$ which does not depend on $w$, and a guiding term
\begin{equation}
V_{\text{guided}}\left(\mathbf{x}_t \right) = -w\cdot \log(M) +w\log\left(1+\sum_{\mu > 1}^{M}e^{-\frac{1}{2(\sigma^2+t)}\left(\|\mathbf{x}_t-\mathbf{c}^{\mu}\|^2-\|\mathbf{x}_t-\mathbf{c}^1\|^2\right)}\right), 
\end{equation}
that depends linearly on $w$ and also on the non-conditioning classes $\{\mathbf{c}^{\mu}\}_{\mu > 1}$. 
\subsubsection{Analysis of the Guided Dynamics \label{sec:rem}}

Inspired by \cite{biroli_meta}, we analyze the effective diffusion potential through the lens of statistical physics.
We will use a simplified and heuristic \textit{mean-field} description,  replacing $V_{\text{guided}}(\mathbf{x}_t)$ with $\mathbb{E}\left[ V_{\text{guided}}(\mathbf{x}_t)\right]$ where expectation is over the realization of all $\mathbf{c}^\mu$ except for the reference one $\mathbf{c}^1$.

We consider different scaling regimes of $M$ with respect to the ambient space dimensions $d$.  Let the number of centroids be $M = e^{\beta(d)\cdot d}$, where $\beta(d)$ will be further specified later. 
By assuming $d \gg 1$, the approximated guided potential can be written as
\begin{equation}
    V_{\text{guided}}(\mathbf{x}_t)\approx -w\left[d\beta(d)-\log\left(1+e^{d\phi_t(\boldsymbol{\mathbf{x}_t|\boldsymbol{c}^1)}}\right)\right],
\end{equation}   
where $\phi_t(\mathbf{x}_t|\mathbf{c}^1)$ is the free energy function of a Random Energy Model (REM), a celebrated statistical physics model that has already been used in the analysis of DMs \cite{biroli_dynamical_2024, achilli2025memorization, achilli2024losing}. The explicit calculation of $\phi_t$ is given in Appendix \ref{app:rem}. Depending on the choice of  $\beta(d)$, $\sigma^2$, and the value of the diffusion time $t$, the potential $V_{\text{eff}}$ can assume two different quadratic shapes that drive the backward process: 
\begin{itemize}
    \item If $\phi_t(\mathbf{x}_t|\mathbf{c}^1) > 0$, the system is in the   \textit{guided phase}, we have a full deformed potential $V_{\text{eff}}(\mathbf{x}_t) = V_{\text{cond}}(\mathbf{x}_t) + V_{\text{guided}}(\mathbf{x}_t)$ which moves in time.
    \item If $\phi_t(\mathbf{x}_t|\mathbf{c}^1) \leq 0$, the system is in the \textit{conditional phase}, with $V_{\text{eff}}(\mathbf{x}_t) = V_{\text{cond}}(\mathbf{x}_t) - dw\beta(d)$ having a steady minimum corresponding to the class $\mathbf{c}^1$ itself.
\end{itemize}
 

Once the expressions for $V_{\text{eff}}$ and its gradient $\nabla_{\mathbf{x}} V_{\text{eff}}$ are obtained, we can solve the CFG SDE in Eq. \eqref{eq:SDE_pot} piecewise integrating the different phases.
The integration can be performed analytically thanks to the piecewise quadratic nature of $V_{\text{eff}}(\mathbf{x})$. We find the transition from the guided to the conditional phase to occur at a specific time in backward diffusion. This event is known as \textit{condensation} transition in statistical physics and it is related to the separation -- or \textit{speciation} \cite{biroli_dynamical_2024, biroli_meta} -- of the modes in the unconditional Gaussian mixture, as explained in Appendix \ref{app:uncond}.  
We find the speciation time $t_s$ by solving the following implicit equation
\begin{equation}
\label{eq:collapse_CFG2}
    \lim_{d \to \infty}\left[\beta(d) + \zeta_{t_s}(\sigma^2,w)\right] = 0,
\end{equation}
where $\zeta_t$ is the moment generating function of the REM. 
Hence the model starts diffusing in the guided phase and then enters the conditional phase at $t = t_s$.
After integrating the full backward SDE as explained in Appendix \ref{sec:mf_CFG},
we obtain the statistics for the diffusive trajectory down to $t = 0$ and measure how $\tilde{p}_0(\mathbf{x}_0|\mathbf{c}^1)$ deviates from $p_0(\mathbf{x}_0|\mathbf{c}^1)$.
Specifically, we can measure the following two distortion estimators
\begin{equation}
    \delta_{\mu} = \lim_{d\to \infty}\frac{\mathbf{c}^1 \cdot(\boldsymbol{\mu}_{w}(0)-\mathbf{c}^1)}{d},\quad\quad\quad
    \delta_{\sigma^2} =\frac{\sigma_{w}^2(0)-\sigma^2}{\sigma^2}.
\end{equation}
The first estimator measures how the \textit{guided mean} $\boldsymbol{\mu}_w(t=0)$ is shifted with respect to the conditional mean, i.e. the class $\mathbf{c}^1$ itself. The second estimator quantifies how the \textit{guided variance} $\sigma_w^2(t = 0)$ is deviating from the conditional variance $\sigma^2$.

\subsubsection{Sub-exponential number of classes}

Examples of sub-exponential scaling regimes of $M$ are: 
the \textit{polynomial} regime where $M = N^a$ with $a > 1$, so that $\beta = \mathcal{O}\left(\frac{\log d}{d}\right)$; the \textit{finite} classes regime where $M = \mathcal{O}(1)$ and $\beta = \mathcal{O}\left(\frac{1}{d}\right)$. 

We solve condition \eqref{eq:collapse_CFG2} in Appendix \ref{sec:assemb} and find that, if the number of classes scales sub-exponentially in $d$, i.e. $\beta(d)$ vanishes with $d$, the speciation time diverges to infinity as 
\begin{equation}
\label{eq:cond_cond}
    t_s(w,d) = \mathcal{O}\left(\frac{1+w}{\beta(d)}\right).
\end{equation} 
As a consequence, the model spends its entire diffusion time in the conditional phase, and the sampling distribution at $t = 0$ must align with the conditional data distribution, resulting in $\delta_{\mu} = 0, \delta_{\sigma^2} = 0$.

\begin{result}[Distortions vanish in the Sub-exponential Regime] Given the data distribution defined in Eq. \eqref{eq:uni_pdf}, 
if the number of classes in the data-set $M$ is sub-exponential in the data dimension $d$, then 
generative distortions induced by CFG vanish in the limit $d \to \infty$.
\end{result}

This result generalizes what obtained by \cite{biroli_meta} for a mixture of two well-separated Gaussians, i.e. $\beta (d) = \ln(2)/d$.

\subsubsection{Exponential number of classes}

\begin{figure*}[t]
\centering
  \includegraphics[width=0.9\linewidth]{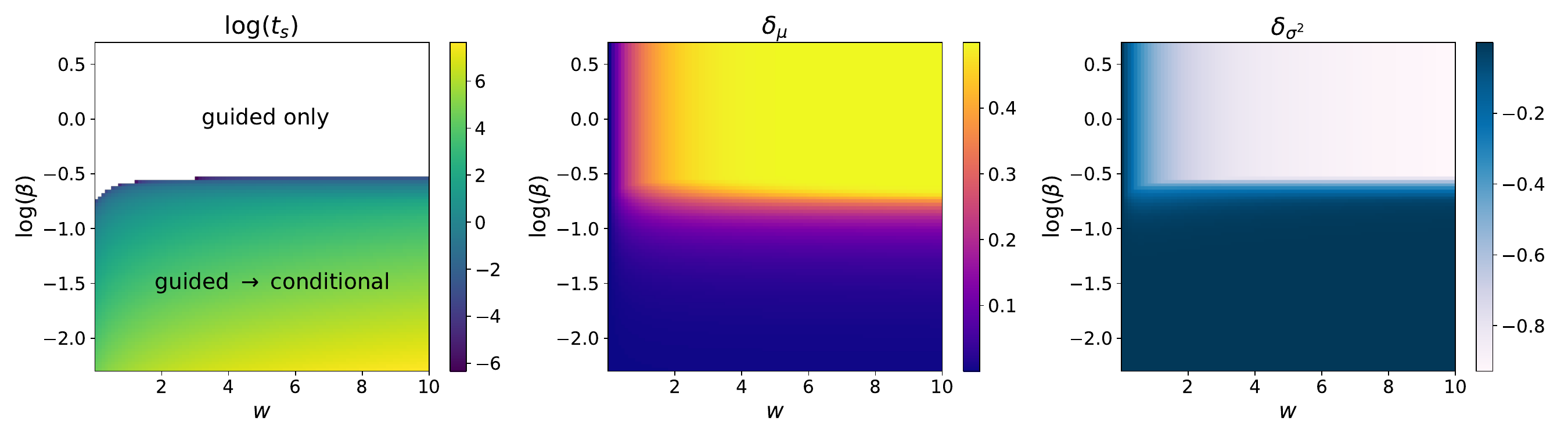}
\caption{Speciation time $t_s$ and the distortion estimators in the exponential regime predicted by the theory as functions of the control parameters $\beta = \log(M)/d$ and $w$, for $\sigma^2 = 0.5$. In the white region, there is no speciation and therefore we have no transition to the conditional phase. This regime displays strong distortion in the conditional sampling as testified by the behavior of $\delta_{\mu}$ and $\delta_{\sigma^2}$. In the small $\beta$ regime instead, where $t_s$ is larger,  distortion is weak.}
\label{fig:distor_exp}
\end{figure*}

Let us now consider mixtures of Gaussians with $ M = e^{\beta d} $, where $ \beta $ is a tunable parameter independent of $ d $. 
When the number of centroids scales exponentially with $ d $, the speciation time satisfies $ t_s = \mathcal{O}(1) $, as obtained by solving the transition condition in Eq.~\eqref{eq:cond_cond} and indeed depends on the full set of parameters $ (w, \beta, \sigma^2) $. 
The underlying mechanism of the emergence of the distortions is tied to the backward SDE: the process evolves within the conditional phase only for $ t < t_s $. If $ t_s $ is well separated from the sampling time, the system has sufficient time to converge to the correct target distribution; conversely, when $ t_s = \mathcal{O}(1) $, convergence is incomplete and distortions persist. In the extreme case $ t_s < 0 $, the transition never occurs, and distortion is unavoidable. 

\begin{figure}[t]
    \centering
    \includegraphics[width=\linewidth]{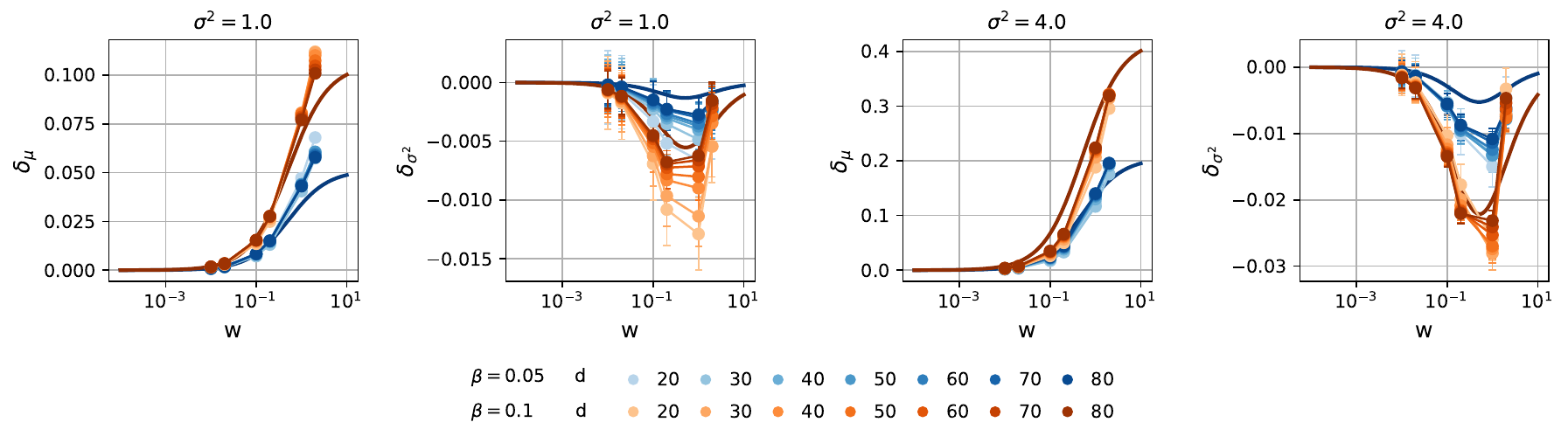}
    \caption{Comparison of the mean-field prediction (full lines) with numerical simulations of CFG (circles) for a mixture of an exponential number of Gaussians as data distribution. The plots show the behavior of distortion estimators $\delta_{\mu}$ and $\delta_{\sigma^2}$, errors are standard deviations of the mean. Theory matches qualitatively and also quantitatively the numerics at increasing $d$. }
    \label{fig:distor_exp_numer}
\end{figure}

This relationship between $ t_s $ and distortion is illustrated in Fig.~\ref{fig:distor_exp}, which reports $ t_s $ alongside the distortion estimators $ \delta_{\mu} $ and $ \delta_{\sigma^2} $ for different values of $ \beta $ and $ w $, at fixed $ \sigma^2 = 0.5 $. 
Similarly, Fig.~\ref{fig:distor_exp_numer} shows the behavior of these estimators as functions of $ w $ for fixed $ \sigma^2 $ and $ \beta $. Notably, the dependence on the guidance strength is non monotonic: $ \delta_{\mu} $ exhibits a maximum (partially visible from the plots), while $ \delta_{\sigma^2} $ displays a clear minimum as $ w $ varies. This behavior can be traced back to the dependence of $ t_s $ on $ w $, which generally increases with the guidance strength, see Fig.~\ref{fig:distor_exp_app} in Appendix~\ref{sec:assemb}. Furthermore, the fact that the positive peak in class alignment and the negative one in sample diversity do not occur at the same guidance level explains the difficulties encountered by practitioners while trying to avoid both an exaggerated overshooting of the mean and mode collapse.
According to our heuristic theory on simple multi-modal data, the best range for the guidance level would be included between the two mentioned stationary points, i.e. $w \in [5, 10]$, which is surprisingly consistent with a common choice in real applications \cite{rombach2022high}.     
Finally, we stress that Fig.~\ref{fig:distor_exp_numer} also includes results from numerical simulations of the guided diffusion, showing close agreement with the predictions of the simplified dynamical mean field theory.

\begin{result}[Distortions persist in the Exponential Regime]
Given the data distribution defined in Eq. \eqref{eq:uni_pdf}, 
if the number of classes in the data-set $M$ is exponential in the data dimension $d$, 
then generative distortions induced by CFG persist in the limit $d \to \infty$.
\end{result}

\subsection{Continuous Classes, Any Dimension \label{sec:cont}}
Let us consider a joint multivariate Gaussian target distribution over $(\mathbf{c},\mathbf{x}) \in \mathbb{R}^{d_1}\times\mathbb{R}^{d_2}$, specifically
\begin{equation}
   \label{eq:multi_joint}p_0(\mathbf{c},\mathbf{x})=\mathcal{N}\!\left(
\begin{pmatrix}\mathbf{c}\\ \mathbf{x}\end{pmatrix};\,
\mathbf{0},\,
\Sigma
\right),\qquad
\Sigma=
\begin{pmatrix}
\Sigma_{cc} & \Sigma_{c x}\\
\Sigma_{x c} & \Sigma_{xx}
\end{pmatrix}, 
\end{equation}

where $\mathbf{c}\in\mathbb{R}^{d_1}$ and $\mathbf{x}\in\mathbb{R}^{d_2}$, with $d_1+d_2=d$ and $\Sigma_{xc}=\Sigma_{cx}^\top$.
In analogy with latent DMs trained on images and their captions \cite{rombach2022high, blattmann2023videoldm}, $\mathbf{c}$ represents a \textit{class} or its representation in a latent space, while $\mathbf{x}$ is the relative data-point, e.g. an \textit{image}.
Let us pin the conditioning class  $\mathbf{c}$, and consider a variance-exploding forward process in the subspace relative to $\mathbf{x}$, namely $\mathbf{x}_0\sim p_0$, $\mathbf{x}_t
=
\mathbf{x}_0
+\sqrt{t}\,\boldsymbol{\epsilon}_t$, $\boldsymbol{\epsilon}_t\sim\mathcal{N}(0,I_d)$.  
Then we have
\begin{equation}
p_t(\mathbf{x}_t | \mathbf{c})
= \mathcal{N}\!\left(\mathbf{x}_t;\, \boldsymbol{\mu},\, \Sigma_{x|c}(t)\right),\quad\quad\quad
p_t(\mathbf{x}_t)
= \mathcal{N}\!\left(\mathbf{x}_t;\, \boldsymbol{0},\, \Sigma_{xx}(t)\right),
\end{equation}
with
$\boldsymbol{\mu}=\Sigma_{xc}\Sigma_{cc}^{-1}\mathbf{c}$, $\Sigma_{xx}(t)=\Sigma_{xx}+tI_{d_2}$, $\Sigma_{x|c}(t)=\Sigma_{x|c}+tI_{d_2}$ and $\Sigma_{x|c}=\Sigma_{xx} -\Sigma_{xc}\Sigma_{cc}^{-1}\Sigma_{cx}$.
The guided score can now be obtained by substituting the corresponding conditional and true score functions into the formula in Eq. \eqref{eq:CFG_score}.
The CFG distribution $\tilde{p}_t(\mathbf{x}_t|\mathbf{c})$ is obtained integrating back in time the SDE
\begin{equation}
    \text{d}\mathbf{x}_t = -A(t)\mathbf{x}_t \text{d}t - B(t)\boldsymbol{\mu}\text{d}t+ \text{d}\mathbf{W}_t,
    \label{eq:SDE2}
\end{equation}
where $A(t) = -(1+w)\left(\Sigma_{x|c}(t) \right)^{-1}+w\left( \Sigma_{xx}(t) \right)^{-1}$ and $B(t) = (1+w)\left(\Sigma_{x|c}(t) \right)^{-1}$.
Let us assume that matrices $\Sigma_{xx}$ and $\Sigma_{x|c}$ commute by construction and that they share a basis of eigenvectors $\{ \boldsymbol{v}^{(i)}\}_{i}^{d_2}$. 
This assumption, known as Common Principal Components Assumption (CPCA) \cite{li_towards2025, flury_84} and used in several application of data analysis, appears to hold in the context of latent DMs, as showed by Appendix \ref{app:cpca}. 

At this point we integrate the SDE and report the passages in Appendix \ref{app:multi}. The resulting marginal density of the guided trajectory is Gaussian with mean and covariance matrix given by
\begin{equation}
\label{eq:guided_mean}
\boldsymbol{\mu}_w(t) = \sum_{i=1}^{d_2} \lambda_i(t)\left(\boldsymbol{v}^{(i),\top}\boldsymbol{\mu} \right)\boldsymbol{v}^{(i)},\quad\quad \Sigma_{w}(t) = \sum_{i=1}^{d_2} \Lambda_i(t) (s_i + t)\ \boldsymbol{v}^{(i)}\boldsymbol{v}^{(i),\top}.
\end{equation}
In our analysis, we have named $\{r_i, s_i\}_i^{d_2}$ the eigenvalues of the matrices $\Sigma_{xx}$ and $\Sigma_{x|c}$.
The full expression of the weights $\lambda_i$ and $\Lambda_i$ as a function of the ratio $s_i/r_i$ are provided in Appendix \ref{app:multi}. We obtain that $\lambda_i \geq 1$ and $\Lambda_i \leq 1\;\; \forall i$.  
Therefore, we can conclude that, at $t = 0$, the mean $\boldsymbol{\mu}$ is always expanded by CFG, while the conditional covariance matrix $\Sigma_{x|c}$ is always contracted for this class of target densities. 
Notice that this result does not depend on the entity of dimensions $d_1$ and $d_2$ themselves. 

\begin{result}[Continuous Classes imply non-vanishing Distortions]
Given the joint distribution of data and classes defined in Eq.~\eqref{eq:multi_joint}, generative distortions induced by CFG are present independently of the dimensionalities $d_1$ and $d_2$.
\end{result}

Consider the Gaussian Mixture defined in Eq. \eqref{eq:uni_pdf}. Then, the joint distribution of variables $(\mathbf{c},\mathbf{x)}$ converges, in the limit of infinite classes $M \to \infty$, to the zero-mean Gaussian defined in Eq. \eqref{eq:multi_joint}, with $d_1 = d_2 = d$ and $\Sigma_{xx} = (\sigma^2 + 1)I_d$, $\Sigma_{cc} = I_d$, $\Sigma_{xc} = I_d$. 
Hence, the general case studied in this Section describes the $M\to \infty$ limit of the separated classes setting, where all classes have fully merged.
Based on this observation, we propose that generative distortions induced by CFG are due to the overlapping supports of the conditional probability distributions relative to the classes in the dataset, rather than dimensionality itself.

\section{Evaluating CFG strategies}
The goal of CFG is to increase the quality of samples and adherence to the conditioning. As already observed, both empirically and theoretically, this procedure might induce a loss of diversity with respect to the true conditional distribution. 
In our analysis, for both the joint Gaussian model with continuous classes and the Gaussian Mixture model, this effect is associated with a contraction of the second cumulant of the samples' distribution. 
Building on our theory, the goal of this Section is to analyze the effects of CFG scheduling, that is, introducing a dependence on $w$ on diffusion time, and to propose a novel procedure that boosts class separation while avoiding loss of diversity. We present here in the main text an analysis based on the Gaussian Mixture with an exponential number of classes. We also performed a similar and more rigorous analysis in the case of the joint Gaussian with continuous classes, reported in Appendix \ref{app:multi_linearCFG}, that shows consistent results. 

\subsection{The sign of \texorpdfstring{$w$}{}  drives distortion}
\label{sec:sign}

Let us now provide a heuristic argument that will lead to a new CFG prescription hindering guidance-induced loss of diversity.

When the target data distribution is a Gaussian mixture, the potential in the \emph{guided phase} is quadratic: $V_{\text{eff}}(\mathbf{x}_t) = \frac{\|\mathbf{x}_t - \mathbf{c}_t^*\|^2}{2{\sigma_t^*}^2} + \text{others}$. 
The position of the minimum and the width of the potential well, read 
\begin{equation}
\label{eq:min}
    \mathbf{c}_t^*= \frac{(1+w)(\sigma^2+t+1)}{\sigma^2+t+1+w}\;\mathbf{c}^1, \quad\quad\quad {\sigma_t^*}^2 = \frac{(\sigma^2+t)(\sigma^2+t+1)}{\sigma^2+t+1+w}.    
\end{equation}
If we compare $\mathbf{c}_t^*$ with $\mathbf{c}^1$, and ${\sigma_t^*}^2$ with the conditional variance $\sigma^2+t$, we realize that $\|\mathbf{c}^*_t\| > \|\mathbf{c}^1\|$ and ${\sigma_t^*}^2 < \sigma^2 + t$, $\forall\; t$ when $w  >0$, while the opposite trend is expected for $w < 0$. This observations, combined with the results from our theory, suggest that positive guidance levels, employed by prior schedules in literature, must always imply mean overshooting and variance shrinkage at $t = 0$.  
We empirically analyze CFG distortions on Stable Diffusion (v1.5) \cite{rombach2022high} where a constant positive guidance level is applied. In Figure \ref{fig:distortion_metrics} we report mean and variance deformations according to standard metrics defined in Appendix \ref{app:metrics}, and show that they support theoretical predictions, i.e. means are shifted away from the class and sample variance reduces when $w$ increases. Figure \ref{fig:fantasy_samples} permits to visualize how images generated by vanilla CFG tend to look similar at high values of $w$, as a consequence of variance shrinkage.  
Further experimental details are provided in Appendix \ref{app:add_exp}.

Based on our observations, we want to design a CFG strategy that simultaneously achieves two objectives: \textit{class separation}, i.e. $\delta_{\mu} > 0$, which should be promoted by exposing the guided trajectory to a \textit{positive} guidance level; and \textit{sample diversity}, i.e. $\delta_{\sigma^2} \geq 0$, which should instead be induced by exposing the guided trajectory to a \textit{negative} guidance level.

\begin{figure}[t]
    \centering

    \begin{subfigure}[t]{0.42\linewidth}
        \centering
        \includegraphics[width=\linewidth]{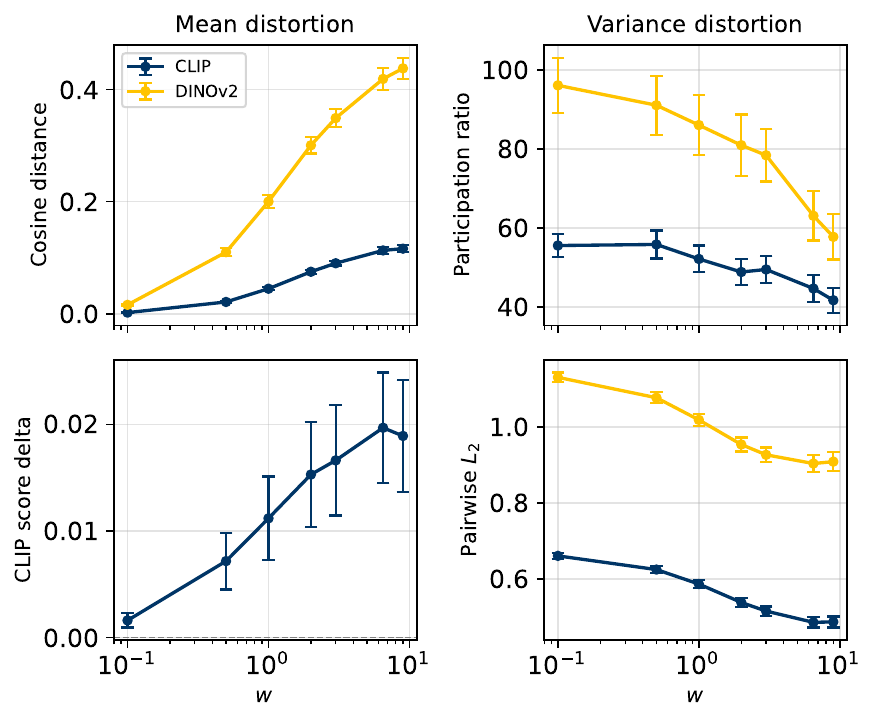}
        \caption{}
        \label{fig:distortion_metrics}
    \end{subfigure}
    \hfill
    \begin{subfigure}[t]{0.5\linewidth}
        \centering
        \includegraphics[width=\linewidth]{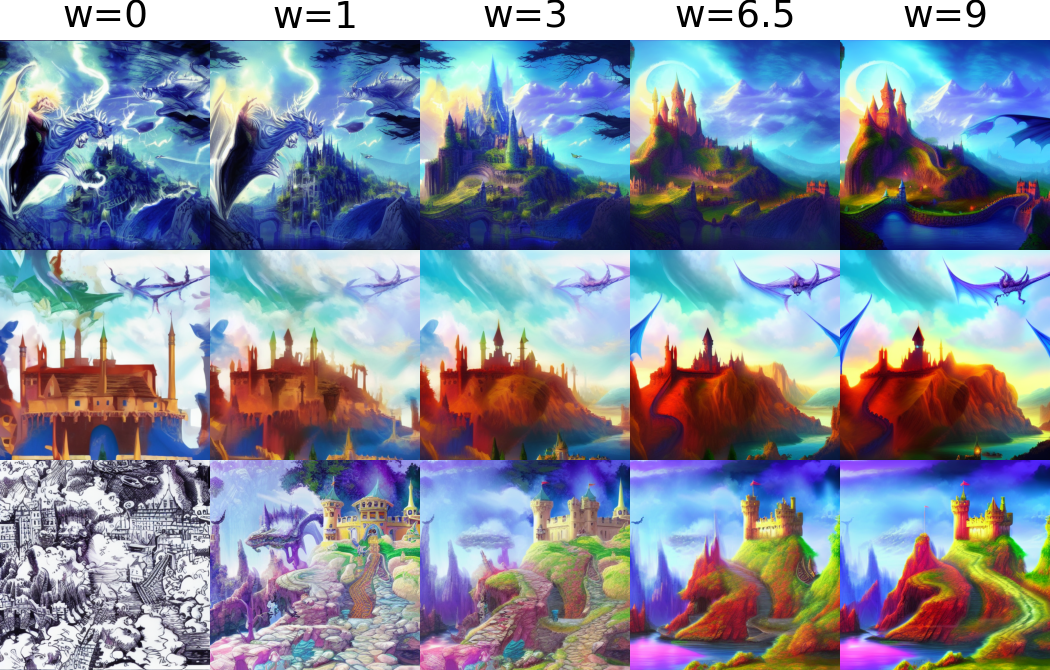}
        \caption{}
        \label{fig:fantasy_samples}
    \end{subfigure}

    \caption{(a) Measures of distortion from a guided Stable Diffusion (v1.5) model, in feature space, as a function of the guidance level $w$, averaged over 50 prompts $\times$ 20 samples, error bars are standard deviations of the mean. Blue circles refer to CLIP feature extractor, yellow ones to DINOv2. 
    (b) Samples generated from the prompt \textit{a fantasy landscape with castles and dragons, vibrant colors, digital art}. Rows refer to different random seeds, columns to guidance levels.}
    
    \label{fig:metrics_distortion}
\end{figure}

\subsection{Increasing Diversity through a Negative Guidance Window}

\begin{figure}[t]
    \centering
    \begin{subfigure}[t]{0.35\textwidth}
        \centering
        \includegraphics[width=\linewidth]{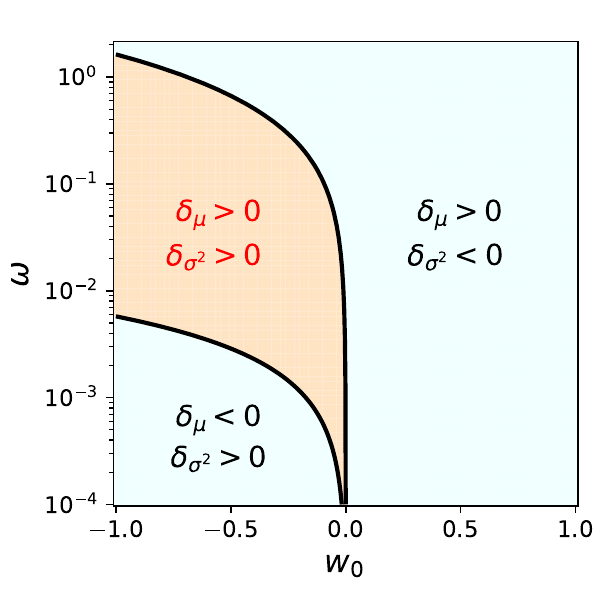}
        \caption{}
        \label{fig:negCFG_distor_exp}
    \end{subfigure}
    \hfill
    \begin{subfigure}[t]{0.59\textwidth}
        \centering
        \includegraphics[width=0.9\linewidth]{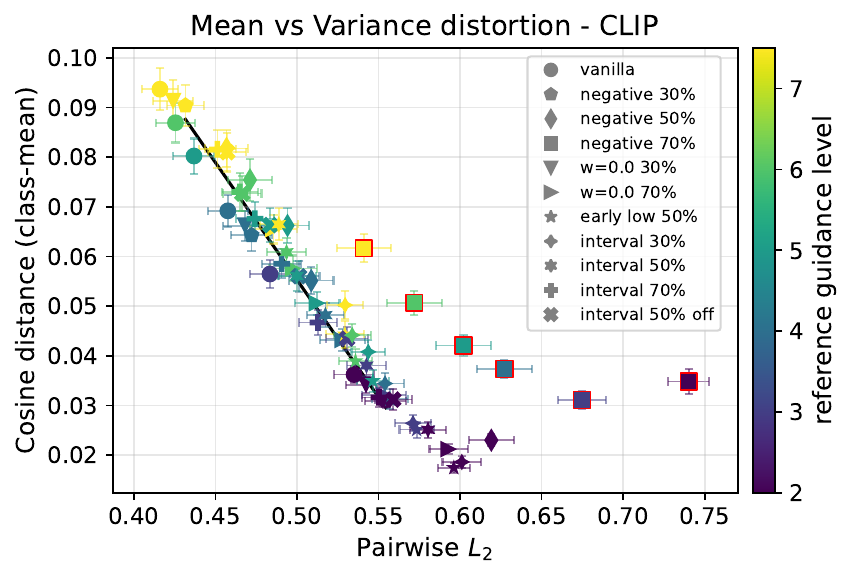}
        \caption{}
        \label{fig:guidance_schedules}
    \end{subfigure}
    \caption{(a): Distortions'phase diagram for the negative-window schedule, as function of $w_0$ and $\omega$. Data are from the Gaussian Mixture model with $\sigma^2 = 0.75$ and $\beta > 2$. 
    The orange region signals a gain in both sample diversity and class separability, other regions show either loss of separability or tendency to shrink the mode variance. (b): We compare several guidance schedules: our negative-window prescription is consistent with standard techniques in  prompt alignment and outperforms them in sample diversity. Points are averaged over 50 prompts $\times$ 20 samples, error bars are standard deviations of the mean.  }\label{fig:schedules}
\end{figure}

We hence propose a ``early-high'' CFG schedule represented by the following function
\begin{equation}
    w(t) = w_0 + \omega\cdot t,\qquad w_0\geq-1,\;\omega > 0.
\end{equation} 
According to this procedure, the trajectory is exposed to two different signs of the guidance level across the backward process: at large times, the guided diffusive trajectory is pushed to expand the means and contract the variances; when $0 < t <\text{max}(0, -\frac{w_0}{\omega})$, the system is induced to reduce means and expand variances. 

By substituting $w(t)$ into the backward SDE in Eq.~\eqref{eq:SDE_pot} we obtain the typical guided trajectories, their statistics and a pair of distortion estimators $\delta_{\mu}$ and $\delta_{\sigma^2}$. Detailed calculations are reported in Appendix~\ref{app:mog_linear_CFG}. For simplicity, we consider $\beta$ large enough that the dynamics stays in the \emph{guided}-only phase.
We fix $\sigma^2 = 0.75$ and $\beta \gg 1$ and vary $w_0$ and $\omega$ to construct the phase diagram depicted in Figure~\ref{fig:negCFG_distor_exp}. 
We identify a region in the phase diagram where $\delta_{\mu} > 0$ and $\delta_{\sigma^2} > 0$, which lies in the  $w_0 < 0$ half-plane, as predicted by our heuristic argument. There rest of the diagram is associated with mismatched signs of the estimators: at $w_0 < 0$ and small $\omega$, the negative-guidance window is too large, and even the mean collapses; at $w_0 > 0$ and high values of $\omega$, guidance is always positive and variances must shrink.

\begin{result}[Negative Guidance improves Diversity] 
In contrast to standard CFG schedules leveraging positive guidance levels only, early-high schedules featuring a negative-guidance window can simultaneously achieve $\delta_{\mu}>0$ and $\delta_{\sigma^2}>0$.
\end{result}

Figure~\ref{fig:guidance_schedules} reports mean and variance distortions evaluated on images generated through time-dependent guidance schemes in Stable Diffusion (v1.5).
The solid black line signals the trend followed by vanilla CFG, i.e. fixed guidance level, and also standard guidance routines such as early-high / early-low and intermediate window schedules \cite{stage_wise, weight_schedulers2024}. 
However, the schedules highlighted in red, which correspond to a  negative guidance window of $70\%$ time-steps width, are shifted to the right of the line, signaling a clear gain in diversity and a good degree of class separation. Interestingly, a net improvement appears when applying $50-70 \%$ of window width. Here we adopt cosine similarity to evaluate mean shift, and pairwise distance to address variance shrinkage, accordingly with prior literature \cite{stage_wise, biroli_meta} and also to be consistent with the phase diagram in panel (a). 
Nevertheless, Figures \ref{fig:schedule-sweep-clip}, \ref{fig:schedule-sweep-dino}, \ref{fig:clip_scores} in Appendix \ref{app:experiments} evaluates additional standard metrics such as Human Preference Score (HPS), CLIP score and Vendi Score and show that our negative-window schedule is the only one allowing to preserve class separation while reaching a sample diversity which is higher than the conditional setting. This is consistent with the theory, that predicts a positive, yet slightly milder, gain in class alignment, and a positive gain in sample diversity too.  
For the purpose of our analysis, we are not formally addressing the visual quality of samples generated by our proposed scheme. Nevertheless, the examples shown in Figures \ref{fig:auroras} and \ref{fig:castles} appear flawless to the human eye.    

Further experimental details can be found in Appendix~\ref{app:experiments}.

\section{Conclusions}
We studied classifier-free guidance (CFG) through the lens of generative distortion, defined as the deviation between the CFG sampling distribution and the true conditional distribution. 
In the context of high-dimensional multi-modal data, we identified a sharp regime distinction: distortion persists when the number of modes is exponential in dimension, but vanishes for sub-exponential mode counts, generalizing previous results \cite{biroli_meta} and ruling out asymptotic alignment of CFG with conditional diffusion in multi-modal settings. 
As a limit of the multi-modal setting, we evaluated Gaussian data conditioned on continuous classes, and showed that distortions cannot vanish, independently of the data dimension. Summing up all previous results, we suggest that distortions are not a dimensional effect, but instead arise whenever classes are not sufficiently separated. 
We eventually showed that standard CFG schedules using a positive guidance level cannot prevent variance shrinkage, and we proposed a principled schedule incorporating a negative-guidance window that can simultaneously preserve class separability and enhance diversity. We derived a phase diagram of the distortions and tested our schedule on real data, validating the theoretical predictions.

\textbf{Limitations \& Future Work.} Even though our primary objective was to develop a tractable, yet sufficiently general, analytical framework for studying guidance distortions, our analysis is restricted to simple Gaussian data distributions, which may not faithfully capture the complexity of real-world data.
Extending this framework to more general target distributions, in line with \cite{achili_speciation26}, as well as deriving an asymptotically exact mean-field theory for CFG dynamics, constitutes a natural direction for future work.
Regarding the proposed negative-window schedule, our intent was to isolate and demonstrate the effect of negative guidance on the sampling process, rather than to provide a comprehensive empirical evaluation. Therefore, a systematic analysis of negative guidance levels within CFG, including their broader impact and optimal tuning, is left for future investigation.

\section*{Acknowledgements}
The authors thank Giulio Biroli and Luca Saglietti for fruitful discussions. CL and EV acknowledge the European
Union - Next Generation EU fund, component M4.C2, investment 1.1 - CUP J53D23001330001.

\printbibliography


\newpage
\appendix

\section*{\Huge \bfseries Appendices}
\vspace{5em}

\section{Further Related Works}
\label{app:further}

\subsection{Main CFG Theoretical Contributions}

Table~\ref{tab:theo_comp} summarizes a comparison of theoretical works in the literature that analyze generative distortions induced by CFG. 
The predominant data models considered are Gaussian distributions and Gaussian mixtures with $M$ modes. We indicate, for each work, whether class alignment and diversity are explicitly evaluated, as well as the dimensional regime under study. \cite{fu_unveil2024, li_provable2025} adopt more general statistical frameworks, and in particular \cite{li_provable2025} provides a theoretical analysis of the Inception Score as an aggregate measure of sample quality.

Our analysis generalizes many of the data model employed so far and studies, under the lens of statistical physics, the unexplored yet realistic setting of an extensive number of data classes in a high-dimensional ambient space.  

\begin{table}[h]
\centering
\resizebox{\linewidth}{!}{
\begin{tabular}{lcccccccc}
\toprule
Ref. & Gaussian & GMM & GMM & General \& & Finite $d$ & $d \to \infty$ & Class & Diversity \\
 & (linear score) &  $M=\mathcal{O}(1)$ &  $M$ scaling with $d$ & Statistical Theory &   &   & Alignment  &   \\
\midrule
\cite{li_towards2025}, Li et al. (2025)                & x &   &   & & x &   & x &   \\
\cite{biroli_meta}, Pavasovic et al. (2025)         &   & x &   &  &  & x & x & x \\
\cite{stage_wise}, Jin et al. (2025)             &   & x &  & & x &   & x & x \\
\cite{li_provable2025}, Li \& Jiao (2025)     &   &   &   & x &   &   &   &   \\
\cite{lu_whatCFGdoes2024}, Chidambaram et al. (2024)       &   & x &   &  & x &   & x & x \\
\cite{wu_mog2024}, Wu et al. (2024)               &   & x &   &  & x &   & x & x \\
\cite{bradley2025}, Bradley \& Nakkiran (2024)      & x & x &   &  & x &   & x &   \\
\cite{fu_unveil2024}, Fu et al. (2024)      &   &   &   & x &   &   &   &   \\
\textbf{Ours}                  & \textbf{x} & \textbf{x} & \textbf{x} &  & \textbf{x} & \textbf{x} & \textbf{x} & \textbf{x} \\
\bottomrule
\end{tabular}
}
\vspace{0.5cm}
\caption{Comparison of data models, dimensional assumptions, and core analytical objectives in theoretical studies of guided diffusion models.}
\label{tab:theo_comp}
\end{table}

\subsection{Comparison with Pavasovic et al. (2026)}
The recent work by \cite{biroli_meta} has been the first one to apply statistical physics to study the sampling dynamics of a classifier-free guided DM, in the same style as in \cite{raya_spontaneous_2023, biroli_dynamical_2024}. The procedure, that we also implement, consists in writing the score function in terms of the gradient of an effective potential, and then integrating the backward CFG SDE. In the same style as  \cite{raya_spontaneous_2023, biroli_dynamical_2024}, the authors adopt data generated by a mixture of two Gaussians and study the overlap, in time, of the particle position with the centroids. They show that, in the limit $d \to \infty$, the potential undergoes a transition from the distorted to the conditional shape, as we also find, at a diverging time $t_s \to \infty$, implying no final distortion at sampling time $t = 0$. 

The theory contained in our Section \ref{sec:rem} generalizes their study for a mixture of two Gaussians to a mixture of $M$ Gaussians, where $M = \exp\left( \beta(d)\cdot d\right)$. 
We hence included, as a new control parameter, the density of classes $\beta(d)$. In our case, the resulting diffusion potential can be mapped in the partition function of a so-called Random Energy Model (REM) \cite{Derrida1981, lucibello_exponential_2023}, undergoing a condensation phase transition at a time that we also named $t_s$. 
When $\beta(d) = 1/d$, and $d \to \infty$, we recover the result of \cite{biroli_meta}. Furthermore, when $\beta(d)$ is not vanishing in the same limit, the transition time $t_s$ does not diverge, implying the distortion effects to accumulate along the diffusive trajectory and show up at sampling time $t = 0$. 
As an additional novelty, with respect to \cite{biroli_meta}, we use statistical mechanics to evaluate CFG schedules: beforehand, the proposal of new CFG prescriptions was suggested by theory, yet not modeled by it. 
Note that \cite{biroli_meta} work in the variance-preserving schedule, while we adopted the variance-exploding one. Passing from one formulation to the other does not change the qualitative behavior of the observables under consideration, such as the speciation time $t_s$. 

\subsection{Comparison with Jin et al. (2025)}

The first difference between \cite{li_towards2025} and our work is the choice of the data models: in Section \ref{sec:mix}, we consider the unconditional data distribution to be a Gaussian mixture, and the conditional one to coincide with one selected Gaussian in the sum; \cite{li_towards2025} model the conditional distribution itself to be a weighted Gaussian mixture, as each class is a cluster of multiple hierarchical sub-classes. 
The analysis provided in \cite{li_towards2025} sheds light on the way the CFG diffusive trajectory is distorted to first align with the center of mass of the class, then suppress weaker modes in the cluster, and eventually reduce the intra-mode variance. 
Despite the dynamical phenomenology is not only consistent with both our analysis and \cite{biroli_meta}, but even richer, in terms of details, we decided to focus on the way the number of classes, which is a global feature of the unconditional distribution, affects the nature of the distortions, in both low and high-dimensional settings. 

\subsection{Comparison with Li et al. (2025)}
While developing our analysis, we found that the concurrent work from \cite{li_towards2025} evaluated the same data model that we considered in Section \ref{sec:cont}. We now proceed to confront our approach with theirs, as well as with the comparison among the results. 
The main difference between the two works is that \cite{li_towards2025} adopt a ODE formulation conditioned to the initialization of the diffusive trajectory. This approach allows the authors to obtain an insightful and detailed description of how the conditional mean is shifted, hence monitoring sample quality and prompt-alignment, yet not explicitly quantifying variance distortion. 

Our approach, based instead on the SDE formulation, provides for a closed expression of mean-shift, which is consistent with their result and, additionally, an explicit quantification of variance shrinking. Our analysis, in fact, wants to disentangle the effects of CFG on class separation and sample diversity.   
Furthermore, in Appendix \ref{app:multi_linearCFG}, we extend our analytical derivation of the distortions in the uni-modal setting to evaluate the effect of linear CFG schedules. 

\subsection{Comparing Guidance Schedules}
Despite some common image generates, such as Stable Diffusion, imply CFG with a constant guidance level $w$, many practitioners have developed time-dependent prescriptions that can improve conditional data generation, for what concerns quality, class separation and diversity \cite{weight_schedulers2024}.

Guidance schedules that start with \textit{high} / \textit{low} guidance levels at large diffusion times are what \cite{stage_wise} refer to as \textit{early-high} /\textit{early-low}  prescriptions. \cite{stage_wise} shows that, while the first type tends to boost prompt alignment while degrading diversity, the second one  mitigates loss of global diversity but reduces fine-grained variation.
A successful class of guidance schedules gathers prescriptions where CFG is applied only at \textit{intermediate} times. For instance, \cite{kynkaaniemi} suggests to apply CFG only in a limited intermediate interval, while \cite{stage_wise} and \cite{malarz25} suggest to progressively increase guidance until reaching a peak and then decrease it. These methods appear to preserve prompt-alignment while mitigating the loss of sample diversity. A possible explanation of such an effect could be the fact that DMs are more susceptible to class bifurcation at mid-stage during backward diffusion \cite{biroli_dynamical_2024, raya_spontaneous_2023, stancevic26, achili_speciation26}
Another section of works in literature proposes more complicated \textit{non-linear} schedules: e.g. \cite{biroli_meta} improves generation applying a power-law CFG while \cite{gao23} propose a cosine-shaped guidance level.  

Another technique we would like to mention, is the so-called \textit{negative guidance} or \textit{negative prompting} \cite{ban_understanding, ambro_negative2025}. This method, that resonates with unlearning procedures in machine learning, consists into guiding diffusion towards a positive class, while trying to avoid, usually through a repulsive force, an antagonist class. In this way, engineers are capable of generating data that do not contain specific features or traits.

All the methods mentioned above rely on applying a net positive guidance level. To the best of our knowledge, we are the first ones to propose to apply total negative guidance along the diffusive process, and to obtain positive results in terms of both class separation and sample diversity. 

\section{Multivariate Gaussian Data \label{app:multi}}

Consider the joint class-data distribution 

\begin{equation}
   p_0(\mathbf{c},\mathbf{x})=\mathcal{N}\!\left(
\begin{pmatrix}\mathbf{c}\\ \mathbf{x}\end{pmatrix};\,
\mathbf{0},\,
\Sigma
\right),\qquad
\Sigma=
\begin{pmatrix}
\Sigma_{cc} & \Sigma_{c x}\\
\Sigma_{x c} & \Sigma_{xx}
\end{pmatrix}, 
\end{equation}

defined on a $d$ dimensional ambient space. 
where $\mathbf{c}\in\mathbb{R}^{d_1}$ and $\mathbf{x}\in\mathbb{R}^{d_2}$ with $d_2+d_1=d$ and $\Sigma_{xc}=\Sigma_{cx}^\top$.
We consider a variance-exploding forward process \cite{achilli2024losing, ventura2024spectral}, namely
\begin{equation}
\begin{pmatrix}\mathbf{c}_0\\ \mathbf{x}_0\end{pmatrix}\sim p_0,\qquad
\begin{pmatrix}\mathbf{c}_t\\ \mathbf{x}_t\end{pmatrix}
=
\begin{pmatrix}\mathbf{c}_0\\ \mathbf{x}_0\end{pmatrix}
+\sqrt{t}\,\boldsymbol{\epsilon}_t,\qquad
\boldsymbol{\epsilon}_t\sim\mathcal{N}(0,I_d).
\end{equation}
We call $p_{t}(\mathbf{c}_t,\mathbf{x}_t)$ the joint distribution of variables $(\mathbf{c}_t,\mathbf{x}_t)$,
that is 

\begin{align}
p_{t}(\mathbf{c}_t,\mathbf{x}_t) &=\int \text{d}p_{0}(\mathbf{c}_0,\mathbf{x}_0)\ p(\mathbf{c}_t,\mathbf{x}_t|\mathbf{c}_0,\mathbf{x}_0)
\\&=\int \text{d}\mathbf{c}_0\text{d}\mathbf{x}_0\ \mathcal{N}(\mathbf{c}_0,\mathbf{x}_0;0,\Sigma)\ \mathcal{N}(\mathbf{c}_t,\mathbf{x}_t;\mathbf{x}_0,tI_{d})=\mathcal{N}(\mathbf{c}_t,\mathbf{x}_t;0,\Sigma+tI_{d})
\end{align}
We now pin a vector of features $\mathbf{c}$ in the $d_1$-dimensional subset of the ambient space, and we call it \textit{class}.
The sampling distribution conditioned with respect to such class reads

\begin{align}
p_t(\mathbf{x}_t| \mathbf{c})
&=\int p_0(\mathbf{x}_0| \mathbf{c})\,p(\mathbf{x}_t| \mathbf{x}_0)\,d\mathbf{x}_0\\
&=\int \mathcal{N}\left(\mathbf{x}_0;\Sigma_{xc}\Sigma_{cc}^{-1}\mathbf{c},\Sigma_{x| c}\right)
\mathcal{N}\left(\mathbf{x}_t;\mathbf{x}_0,tI_{d_2}\right)\,d\mathbf{x}_0\\
&=\mathcal{N}\left(\mathbf{x}_t;\Sigma_{xc}\Sigma_{cc}^{-1}\mathbf{c},\Sigma_{x| c}+tI_{d_2}\right)
=\mathcal{N}\left(\mathbf{x}_t;\Sigma_{cx}^{\top}\Sigma_{cc}^{-1}\mathbf{c},\Sigma_{x| c}^{t}\right),
\end{align}
with
\begin{equation}
\Sigma_{x|c}^{t}=\Sigma_{xx}-\Sigma_{cx}^{\top}\Sigma_{cc}^{-1}\Sigma_{cx}+tI_{d_{2}}.
\end{equation}
Therefore we can rewrite the full conditional density of $\mathbf{x}_t$
as 

\begin{equation}
p_{t}(\mathbf{c}_t,\mathbf{x}_t|\mathbf{c})=\frac{1}{Z_{t}}\exp-\frac{1}{2}(\mathbf{x}_t-\boldsymbol{\mu})^{\top}(\tilde{\Sigma_{t}})^{-1}(\mathbf{x}_t-\boldsymbol{\mu}),
\end{equation}
where the observables become

\begin{equation}
Z_{t}=(2\pi)^{d/2}\det(\tilde{\Sigma_{t}})^{1/2},
\end{equation}

\begin{equation}
\boldsymbol{\mu}=(\mathbf{c},\Sigma_{cx}^{\top}\Sigma_{cc}^{-1}\mathbf{c}),
\end{equation}

\begin{equation}
\tilde{\Sigma_{t}}=\begin{pmatrix}tI_{d_{1}} & 0\\
0 & \Sigma_{x|c}^{t}
\end{pmatrix}.
\end{equation}

Let us restrict to the $d_2$-dimensional sub-space relative to $\mathbf{x}$. One has
\begin{equation}
    p\left(\mathbf{x}_t | \mathbf{c} \right) = \mathcal{N}\left(\mathbf{x}_t; \boldsymbol{\mu},\Sigma_{x|c}^t \right),
\end{equation}
\begin{equation}
    p\left(\mathbf{x}_t\right) = \mathcal{N}\left(\mathbf{x}_t; \boldsymbol{0},\Sigma_{xx}^t\right)
\end{equation}
where
\begin{align}
\boldsymbol{\mu}&=\Sigma_{cx}^{\top}\Sigma_{cc}^{-1}\mathbf{c}\\
\Sigma_{xx}^t &= \Sigma_{xx} + t I_{d_2}\\ 
\Sigma_{x|c}^t &= \Sigma_{x|c} + t I_{d_2}\\ 
\Sigma_{x|c}&=\Sigma_{xx}-\Sigma_{cx}^{\top}\Sigma_{cc}^{-1}\Sigma_{cx}
\end{align}
The relative score function then become
\begin{equation}
    \nabla_{\mathbf{x}}\log p\left(\mathbf{x}_t | \mathbf{c} \right) = -\left(\Sigma_{x|c}^t\right)^{-1}\left(\mathbf{x}_t - \boldsymbol{\mu} \right),
\end{equation}
\begin{equation}
    \nabla_{\mathbf{x}}\log p\left(\mathbf{x}_t\right) = -\left(\Sigma_{xx}^t\right)^{-1}\mathbf{x}_t .
\end{equation}
As a consequence, the guided score becomes
\begin{equation}
\tilde{s}_{t}(\mathbf{x}_t|\mathbf{c})=
(1+w)\nabla_{\mathbf{x}}\log p_{t}(\mathbf{x}_t|\mathbf{c})-w\nabla_{\mathbf{x}}\log p_{t}(\mathbf{x}_t).
\label{eq:CFG_score2}
\end{equation}
We want to find the probability distribution $\tilde{p}_t(\mathbf{x}_t|\mathbf{c})$ that samples the configurations generated by the following reverse-time SDE
\begin{equation}
    d\mathbf{x}_t = -\tilde{s}_{t}(\mathbf{x}_t|\mathbf{c})dt + dW_t. 
    \label{eq:SDE}
\end{equation}
Equation \eqref{eq:SDE} can be rewritten explicitly as
\begin{equation}
    d\mathbf{x}_t = -A(t)\mathbf{x}_t dt - B(t)\boldsymbol{\mu}dt+ dW_t,
    \label{eq:SDE2_app}
\end{equation}
where 
\begin{equation}
    A(t) = -(1+w)\left(\Sigma_{x|c}^t \right)^{-1}+w\left( \Sigma_{xx}^t \right)^{-1},
\end{equation}
\begin{equation}
    B(t) = (1+w)\left(\Sigma_{x|c}^t \right)^{-1}.
\end{equation}
Integrating Eq.~\eqref{eq:SDE2} backward in time brings to

\begin{align}
    \mathbf{x}_t & = M(t,T)\mathbf{x}_T +\int_t^{T}M(t,t')B(t')\boldsymbol{\mu}\,dt'+\int_{t}^{T}M(t,t')\, dW_{t'}.
    \label{eq:sol-cond-sde}
\end{align}
where we defined the matrix kernel
\begin{align}
M(t_1,t_2) = e^{\int_{t_1}^{t_2} A(t')dt'}.    
\end{align}
Let us now evaluate
\begin{align} \int_{t_1}^{t_2}A(t)dt = \int_{t_1}^{t_2}\left[-(1+w)(\Sigma_{x|c}^{t})^{-1}+w(\Sigma_{xx}^{t})^{-1}\right]dt =  \int_{t_1}^{t_2}\left[-\frac{1+w}{\Sigma_{x|c}+tI_{d_2}}+\frac{w}{\Sigma_{xx}+t I_{d_2}} \right] dt 
    \\ = \log\left[(\Sigma_{x|c}+t_2 I_{d_2})^{-(1+w)}(\Sigma_{x|c}+t_1 I_{d_2})^{1+w}\right]+\log\left[\left(\Sigma_{xx}+t_2 I_{d_2}\right)^{w}\left(\Sigma_{xx}+t_1 I_{d_2} \right)^{-w}\right]
\end{align}
Let us assume for simplicity that $\Sigma_{xx}$ and $\Sigma_{x|c}$ commute. This assumption is named Common Principal Components Assumption \cite{li_towards2025, flury_84} and its use is justified, for our specific context, in Appendix \ref{app:cpca}. 
We get
\begin{align}
  M(t_1,t_2) &= Z(t_1)Z(t_2)^{-1}\\
  Z(t) &= (\Sigma_{x|c}+t I_{d_2})^{1+w}\left(\Sigma_{xx}+t I_{d_2} \right)^{-w}
\end{align}
We are interested in the asymptotic limit $T \rightarrow \infty$,  with appropriate initial condition $\mathbf{x}_T \sim \mathcal{N}(\boldsymbol{\mu},T I_{d_2})$. Since Eq.~\eqref{eq:sol-cond-sde} gives us the law $\tilde{p}_w(\mathbf{x}_t | \mathbf{x}_T)$, we should convolve with $\mathbf{x}_T$ to obtain the marginal for $\mathbf{x}_t$. We can argue though (and check for the $w=0$ case) that for large $T$ the initial condition becomes irrelevant, that is $\tilde{p}_t(\mathbf{x}_t | \mathbf{x}_T) \approx \tilde{p}_t(\mathbf{x}_t)$. In practice, we can focus on solving

\begin{align}
    \mathbf{x}_t & = \int_t^{+\infty}M(t,t')B(t')\boldsymbol{\mu}\,dt'+\int_{t}^{+\infty}M(t,t')\, d\mathbf{W}_{t'}.
    \label{eq:sol-uncond-sde}
\end{align}
We see that $\mathbf{x}_t$ is Gaussian, with mean and covariance given by
\begin{equation}
\boldsymbol{\mu}_w(t) =\int_t^{+\infty}M(t,t')B(t')\boldsymbol{\mu}\,dt'
= Z(t)\int_t^{+\infty}Z^{-1}(t')B(t')\boldsymbol{\mu}\,dt'
\label{eq:tildemu}
\end{equation}
\begin{equation}
    \Sigma_w(t) = \int_{t}^{+\infty}M^2(t,t')\, d t' = Z^{2}(t)\int_{t}^{+\infty}Z^{-2}(t')\, d t' \label{eq:tildesigma}
\end{equation}
\subsection{Computation of \texorpdfstring{$\boldsymbol{\mu}_w$}{}}
Consider $\boldsymbol{\mu}_w$ defined in Eq.~\eqref{eq:tildemu}. Call $R=\Sigma_{xx}$ and $S=\Sigma_{x|c}$ and assume they commute.
Then

\begin{align}
\int_{t_1}^{t_2} Z^{-1}(t') B(t')dt'&=(1+w)\int_{t_1}^{t_2} (R + tI_{d_2})^w (S + tI_{d_2})^{-2 - w}\,dt\\
&=P\;\mathrm{diag}(e_1(t_1,t_2),\dots,e_{d_2}(t_1,t_2))\;P^{-1},
\end{align}
where $P$ is the unitary matrix collecting the common eigenvectors to $R$ and $S$ and 
\begin{equation}
e_i(t_1,t_2) = 
\begin{cases}
\dfrac{1}{(s_i - r_i)} \left[\left(\frac{r_i + t_2}{s_i + t_2}\right)^{w+1} - 
\left(\frac{r_i + t_1}{s_i + t_1}\right)^{w+1}\right],
& 
s_i \neq r_i,
\\[8pt]
(1+w)\left[\frac{1}{\,r_i + t_1\,} - \frac{1}{\,r_i + t_2\,}\right],
& 
s_i = r_i.
\end{cases}
\end{equation}

This is the sought-after closed-form solution, valid as long as \((s_i + t)\neq 0\) on \([t_1,t_2]\). 
By substituting $t_1 = t$ and $t_2 = T$ and plugging the last calculation into Eq.~\eqref{eq:tildemu} we obtain
\begin{equation}
    \label{eq:mutilde_w}
    \boldsymbol{\mu}_{w}(t,T) = \sum_{i=1}^{d_2} \lambda_i(t,T)\;( \boldsymbol{v}^{(i),\top} \boldsymbol{\mu})\;\boldsymbol{v}^{(i)},
\end{equation}
where $\{ \boldsymbol{v}^{(i)}\}_{i=1}^{d_2}$ is the basis of eigenvectors shared between  $\Sigma_{xx}$ and $\Sigma_{x|c}$ and 
\begin{equation}
\lambda_i(t,T) = 
\begin{cases}
\frac{1}{(s_i - r_i)}\left[\left(\frac{r_i + T}{s_i + T}\right)^{w+1}\frac{(s_i+t)^{w+1}}{(r_i+t)^{w}}-r_i-t\right],
& 
s_i \neq r_i,
\\[8pt]
(1+w)\left[1 - \frac{r_i+t}{\,r_i + T\,}\right],
& 
s_i = r_i.
\end{cases}
\end{equation}
In the limit $T\to \infty$ the same variables become
\begin{equation}
    \label{eq:new_mutilde_w}
    \boldsymbol{\mu}_{w}(t) = \sum_{i=1}^{d_2} \lambda_i(t)\;( \boldsymbol{v}^{(i),\top} \boldsymbol{\mu})\;\boldsymbol{v}^{(i)},
\end{equation}
\begin{equation}
\lambda_i(t) = 
\begin{cases}
\frac{1}{(s_i - r_i)}\left[\frac{(s_i+t)^{w+1}}{(r_i+t)^{w}}-(r_i+t)\right],
& 
s_i \neq r_i,
\\[8pt]
1+w
& 
s_i = r_i.
\end{cases}
\end{equation}


Since $s_i/r_i < 1$ $\forall i$ it can be deduced that $\lambda_i(t) > 1$ $\forall i, t$. The behavior of $\lambda_i(0)$ as a function of $s_i/r_i$ is showed in Figure \ref{fig:eig_vs_d}.
By rewriting the conditional mean as
\begin{equation}
    \boldsymbol{\mu} = \sum_{i = 1}^{d_2}(\boldsymbol{v}^{(i),\top},\boldsymbol{\mu})\;\boldsymbol{v}^{(i)},
\end{equation}
and comparing it with the expression of $\boldsymbol{\mu}_w$ in Eq. \eqref{eq:mutilde_w} we notice that $\|\boldsymbol{\mu}_w\| > \|\boldsymbol{\mu}\|$ $\forall w$, that means that means are always expanded by the CFG in this framework at $t = 0$. Numerical simulations of the guided process, reported in Figure \ref{fig:cont_mean_cov}, show good agreement with the expected trend from theory. 

\subsection{Computation of \texorpdfstring{$\Sigma_w$}{}}
\begin{figure}[t]
    \centering
    \begin{subfigure}{0.48\textwidth}
        \centering
        \includegraphics[width=0.49\linewidth]{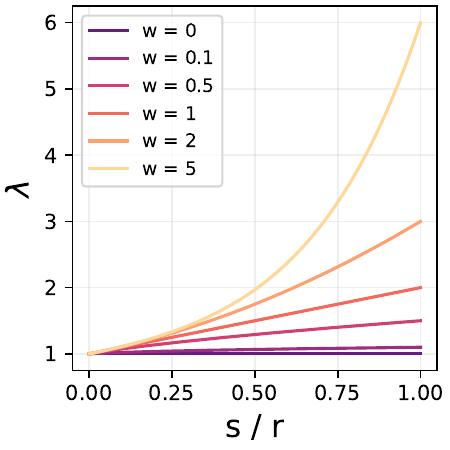}\includegraphics[width=0.49\linewidth]{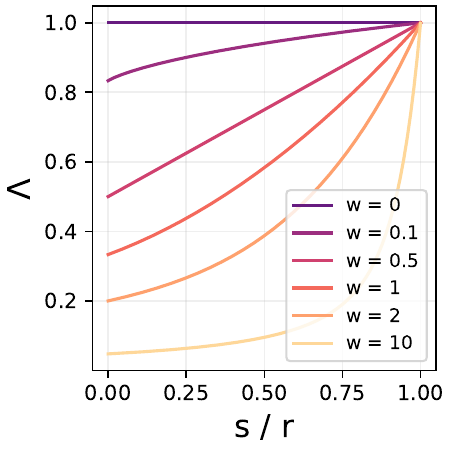}
        \caption{The coefficients $\lambda$ and $\Lambda$ governing the distortion of CFG in the Gaussian setting, at $t = 0$ and as a function of $w$ and the ratio $s/r$, where $(s,r)$ are eigenvalues, respectively, of $\Sigma_{x|c}$ and $\Sigma_{xx}$. Since $\lambda \geq 1$, the mean of the conditional target distribution is always expanded, while $\Lambda \leq 1$ implies a systematic contraction of the covariance matrix.}
        \label{fig:eig_vs_d}
    \end{subfigure}
    \hfill
    \begin{subfigure}{0.51\textwidth}
        \centering
        \includegraphics[width=\linewidth]{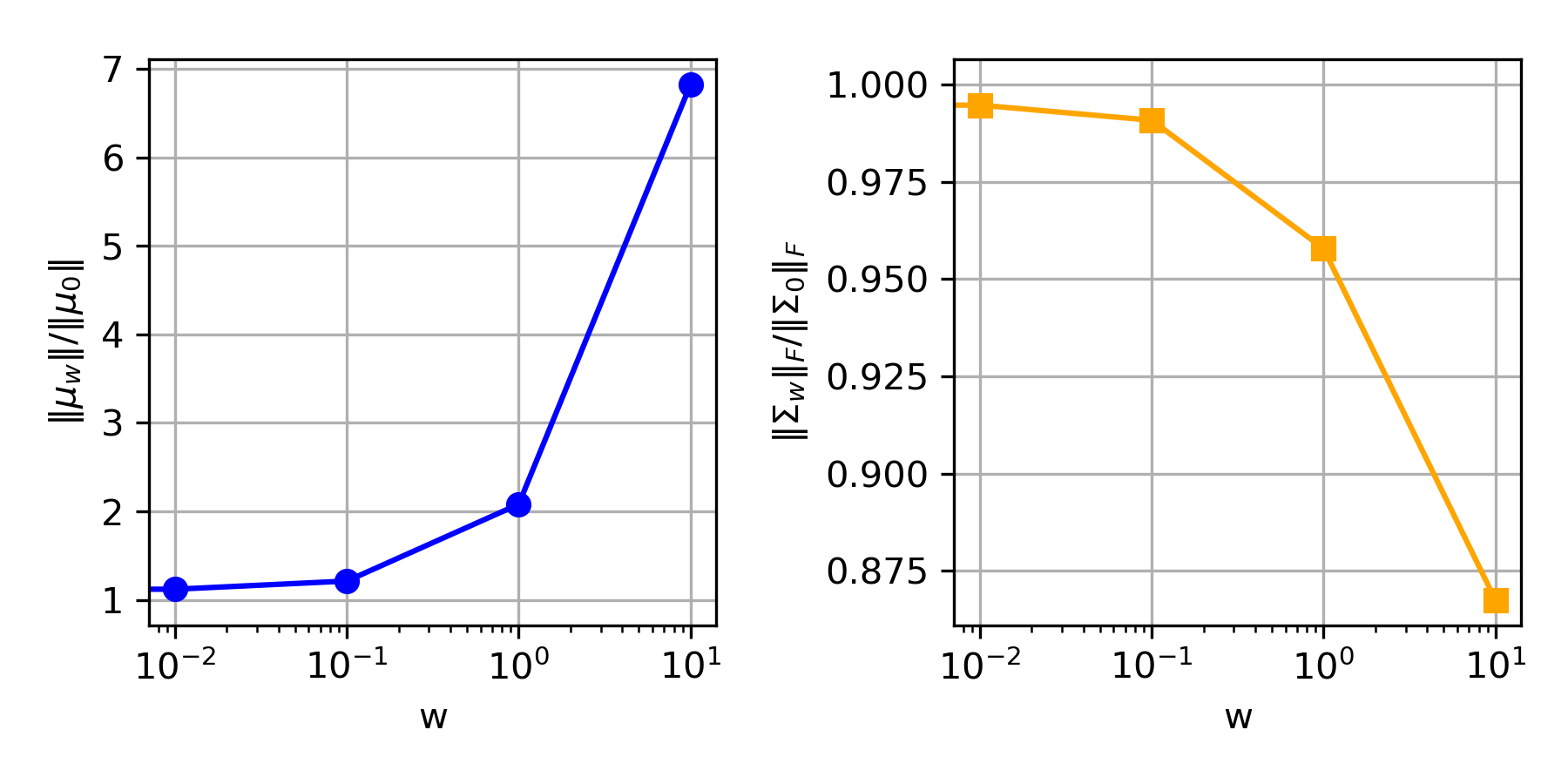}
        \caption{Measure of distortion from numerical simulations for CFG  on jointly Gaussian classes and data, showing increased class separation and decreased diversity with increasing $w$. Dimensions are $d_1 = 1$, $d_2 = 9$. Left: the norm of the CFG mean divided by the true conditional mean. Right: Frobenius norm of $\Sigma_w$ divided by the true conditional covariance matrix.} 
        \label{fig:cont_mean_cov}
    \end{subfigure}
    \caption{CFG induced distortions from theory and numerical simulations in the infinite continuous classes regime.}
    \label{fig:cont_main}
\end{figure}

Consider $\Sigma_w$ defined in Eq.~\eqref{eq:tildesigma}. Call $R=\Sigma_{xx}$ and $S=\Sigma_{x|c}$ and assume they commute.
For each eigenvalue pair $(r_i,s_i)$ of $(R,S)$, the integral is diagonal in the basis that diagonalizes $R$ and $S$. Hence
\begin{align}
\int_{t_1}^{t_2} Z^{-2}(t')dt'&=\int_{t_1}^{t_2} (R + tI_{d_2})^{2w}\ (S + tI_{d_2})^{-2 - 2w}\ dt\\
&=P\ \mathrm{diag}(e_1(t_1,t_2),\dots,e_{d_2}(t_1,t_2)))\ P^{-1},
\end{align}
where $P$ is the unitary matrix collecting the common eigenvectors to $R$ and $S$ and 
\begin{equation}
e_i(t_1,t_2)
=
\begin{cases}
\frac{1}{(2w+1)\ (s_i - r_i)}\left[\left(\frac{r_i + t_2}{s_i + t_2}\right)^{2w+1} - \left(\frac{r_i + t_1}{s_i + t_1}\right)^{2w+1}
\right],
& \text{if }s_i \neq r_i, \\[1.5em]
\displaystyle
\frac{1}{s_i + t_1} -
\frac{1}{s_i + t_2},
& \text{if }s_i = r_i.
\end{cases}
\end{equation}
That is 
valid as long as $(s_i + t)\neq 0$ over $[t_1,t_2]$. By substituting $t_1 = t$ and $t_2 = T$ and plugging the last calculation into Eq.~\eqref{eq:tildesigma} we obtain
\begin{equation}
    \label{eq:sigma tilde_w_app}
    \Sigma_{w}(t,T) = \sum_{i=1}^{d_2} \Lambda_i(t,T) (s_i + t)\ \boldsymbol{v}^{(i)}\boldsymbol{v}^{(i),\top},
\end{equation}
where $\{ \boldsymbol{v}^{(i)}\}_{i=1}^{d_2}$ is the basis of eigenvectors shared between  $\Sigma_{xx}$ and $\Sigma_{x|c}$ and 
\begin{equation}
\Lambda_i(t,T)
=
\begin{cases}
\frac{1}{(2w+1)(s_i - r_i)}\left[\left(\frac{r_i + T}{s_i + T}\right)^{2w+1}\frac{(s_i+t)^{1+2w}}{(r_i+t)^{2w}}-(r_i+t)\right],
& \text{if }s_i \neq r_i, \\[1.5em]
1-\frac{s_i+t}{s_i+T},
& \text{if }s_i = r_i.
\end{cases}
\end{equation}
In the limit $T\to \infty$ the same variables become
\begin{equation}
    \label{eq:sigma tilde_w2}
    \Sigma_{w}(t) = \sum_{i=1}^{d_2} \Lambda_i(t) (s_i + t)\ \boldsymbol{v}^{(i)}\boldsymbol{v}^{(i),\top},
\end{equation}
and
\begin{equation}
\Lambda_i(t)
=
\begin{cases}
\frac{1}{(2w+1) (s_i - r_i)}\left[\frac{(s_i+t)^{1+2w}}{(r_i+t)^{2w}} - (r_i+t) \right],
& \text{if }s_i \neq r_i, \\[1.5em]
1
& \text{if }s_i = r_i.
\end{cases}
\end{equation}
One can notice that this quantity is smaller than unity for any choice of $w > 0$ and $s_i/r_i < 1$. The behavior of $\Lambda_i(0)$ as a function of $s_i/r_i$ is reported in Figure \ref{fig:eig_vs_d}.
By writing the conditional covariance according to its spectral decomposition
\begin{equation}
    \Sigma_{x|c} = \sum_{i = 1}^{d_2}s_i\;\boldsymbol{v}^{(i)}\boldsymbol{v}^{(i),\top},
\end{equation}
and comparing it with the expression of the guided covariance $\Sigma_{w}$ in Eq. \eqref{eq:sigma tilde_w2} we realize that covariances always undergo contraction under CFG in this setup at $t = 0$. Numerical simulations of the guided process, reported in Figure \ref{fig:cont_mean_cov}, show good agreement with the expected trend from theory. 

\subsection{Common Principal Components Assumption for Latent Diffusion Models}
\label{app:cpca}
In order to integrate the backward guided SDE in Eq. \ref{eq:SDE2_app}, we assumed co-diagonalization between the covariance matrix of the image space $\Sigma_{xx}$ and the conditional covariance matrix $\Sigma_{x|c}$ of the images given the class representations. This simple assumption, referred to as Common Principal Components Assumption (CPCA) and also employed by \cite{li_towards2025}, dates back to \cite{flury_84} and it is employed in data analysis for comparing multiple datasets (e.g., across different groups or populations) in a more interpretable fashion \cite{cpca_practical}. 

\begin{figure}[t]

  \centering
  \includegraphics[width=0.6\linewidth]{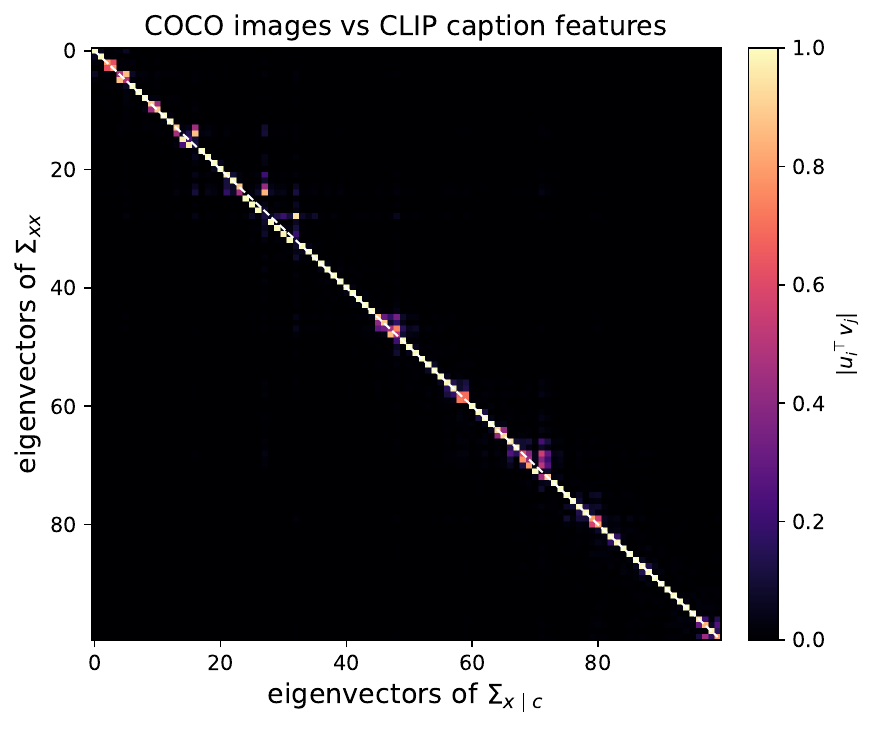}
  \caption{Matrix of the overlaps between the first $100$ eigenvectors of matrices $\Sigma_{xx}$ and $\Sigma_{x|c}$ empirically estimated from COCO dataset embedded in a latent space via CLIP.}

\label{fig:cpca}
\end{figure}

We now provide  numerical evidence showing that CPCA could be approximately correct in real-world datasets and CFG applications. 
We consider $30$k images from the COCO dataset \cite{coco}, of dimension $d_2 = 3 \times 32\times 32 = 3072$ and computed the covariance matrix $\Sigma_{xx}$. Then we took one text caption per image and embedded it in a latent space of dimension $d_1 = 512$ via CLIP \cite{clip} semantic embedding. This is a standard procedure employed by latent DMs, i.e. the state-of-the-art of text-to-image generation. We empirically estimated the covariance matrices $\Sigma_{cc}$ across the latent caption representations, and then $\Sigma_{xc}$ and $\Sigma_{x|c}$ as prescribed in Section \ref{sec:cont}. We finally confronted the first $100$ eigenvectors of $\Sigma_{xx}$ with the ones of $\Sigma_{x|c}$ and reported the overlap matrix in Figure \ref{fig:cpca}: evidently, the overlap matrix is strongly diagonal, suggesting that these bases are similar.
The motivation between CPCA to be valid is most probably due to the fact that captions are embedded so to share, in latent space, the same semantic content of the relative images. 

\section{Data from a Gaussian Mixture \label{app:rem}}

Consider the Gaussian Mixture 
\begin{equation}
\label{eq:target_uncond}
   p_{0}(\mathbf{x}) = \frac{1}{M}\sum_{\mu = 1}^M \mathcal{N}(\mathbf{x};\mathbf{c}^{\mu}, \sigma^2) 
\end{equation}
as the target distribution of our diffusion model. The conditional distribution with respect to one class $\mathbf{c}^1$ is going to be one Gaussian from the mixture, namely 
\begin{equation}
    p_0(\mathbf{x}|\mathbf{c}^1) = \mathcal{N}(\mathbf{x};\mathbf{c}^1,\sigma^2).
\end{equation}
The time evolution of the target distribution according to a variance-exploding forward process will read
\begin{align}
    \label{eq:samp_pdf}p_t\left(\mathbf{x}_t\right) &= \frac{1}{M}\sum_{\mu=1}^M \mathcal{N}\left(\mathbf{x}_t; \mathbf{c}^{\mu},\sigma^2+t\right),\\
    p_t\left(\mathbf{x}_t | \mathbf{c}^1 \right) &= \mathcal{N}\left(\mathbf{x}_t; \mathbf{c}^1,\sigma^2+t \right).
\end{align}
The guided backward process is described by the following SDE 
\begin{equation}
    \label{eq:guided_SDE}\text{d}\boldsymbol{x}_t = -\tilde{\boldsymbol{s}}_t(\boldsymbol{x}_t |\mathbf{c}^1)\;\text{d}t+\text{d}\boldsymbol{W}_t,
\end{equation}
where the guided score function is given by
\begin{equation}
    \label{eq:guide}\tilde{\boldsymbol{s}}_t(\boldsymbol{x}) = (1+w)\boldsymbol{s}_t(\boldsymbol{x}|\boldsymbol{c}^1)-w\;\boldsymbol{s}_t(\boldsymbol{x}),
\end{equation}
where $\text{d}\boldsymbol{W}_t$ is Brownian noise, and $w$ being our guidance level.
The same SDE can be rewritten in terms of an effective diffusion potential as
\begin{equation}
\label{eq:SDE_pot2}
    \text{d}\boldsymbol{x}_t = \nabla_{\boldsymbol{x}}V_{\text{eff}}(\boldsymbol{x}_t)\text{d}t + \text{d}\boldsymbol{W}_t,
\end{equation}
where
\begin{align}
    V_{\text{eff}}(\boldsymbol{x}_t)&=-\log\left[\frac{p_t(\boldsymbol{x}_t|\boldsymbol{c}^{1})^{1+w}}{p_t(\boldsymbol{x}_t)^w}\right] \\
    &= \log\left[e^{\frac{(1+w)}{2(\sigma^2+t)}\|\boldsymbol{x}_t-\boldsymbol{c}^1\|^2}\left(\frac{1}{M}\sum_{\mu = 1}^Me^{-\frac{1}{2(\sigma^2+t)}\|\boldsymbol{x}_t-\boldsymbol{c}^{\mu}\|^2}\right)^w\right]+\frac{d}{2}\log(2\pi t),
\end{align}
Let us define the name of modes in the Gaussian mixture as $M = e^{\beta(d)\cdot d}$, where $\beta(d)$ can arbitrarily scale with the dimension $d$. The same potential can be expressed in terms of a conditional and a guided part as
\begin{align}
    V_{\text{eff}}(\boldsymbol{x}_t)&=V_{\text{cond}}(\boldsymbol{x}_t)+V_{\text{guided}
    }(\boldsymbol{x}_t) \\
    &=\frac{1}{2}\frac{\|\boldsymbol{x}_t-\boldsymbol{c}^1\|^2}{\sigma^2+t}+\frac{d}{2}\log(2\pi t)\\
    &-w\left[d\beta(d)-\log\left(1+\sum_{\mu > 1}^{M}e^{-\frac{1}{2(\sigma^2+t)}\left(\|\boldsymbol{x}_t-\boldsymbol{c}^{\mu}\|^2-\|\boldsymbol{x}_t-\boldsymbol{c}^1\|^2\right)}\right)\right].
\end{align}

\subsection{The Random Energy Model (REM) formalism}
\label{sec:formalism}
Let us introduce the tools needed to solve a generic REM, following \cite{lucibello_exponential_2023, Derrida1981}. 

Let us consider $M=e^{\beta d}$ (or equivalently $M=e^{\beta d}-1$) i.i.d. energy levels $\epsilon^\mu \sim p(\epsilon \,|\, \omega)$, where we extend the typical REM setting allowing for a common source of quenched disorder $\omega \sim p_\omega$. 
The goal is to compute the average asymptotic free energy of the system, defined by
\begin{equation}
\label{eq:free_ene}
\phi_\beta(\lambda) = \lim_{d\to\infty}  \frac{1}{\lambda d} \mathbb{E}\log \sum_\mu e^{\lambda d\epsilon^\mu}     
\end{equation}

We shall assume that the probability distribution of the energy levels is such that, with probability one over the choice of $\omega$ when $d\to \infty$ the cumulant generating function has a well defined limit:
$ \lim_{d\to\infty}\frac{1}{N}\log\mathbb{E}_{\epsilon|\omega}\, e^{\lambda d \epsilon}$ exists, and the distribution over the choices of $\omega$ concentrates around its mean. Then we define the typical cumulant generating function and its Legendre transform::
\begin{align}
\zeta(\lambda)&=\lim_{d\to\infty}\frac{1}{d}\mathbb{E}_\omega\log\mathbb{E}_{\epsilon|\omega}\, e^{\lambda d \epsilon},\label{eq:zeta}\\
s(\epsilon)&=\sup_{\lambda}\ \epsilon\lambda-\zeta(\lambda).
\label{eq:s}
\end{align}

The total entropy of the system is $\Sigma(\epsilon)=\alpha-s(\epsilon)$.
Depending on the value of $\Sigma(\epsilon)$, the REM displays a separation into two thermodynamic phases: an \emph{uncondensed} phase where the system can populate an exponential number of energy levels, at lower values of $\lambda$; a \emph{condensed} phase where the system is able to populate a unique energy state, at higher values of $\lambda$.

Let us define the quantities $\epsilon_{*}(\alpha)$ and $\lambda_{*}(\beta)$
respectively as the 
maximum value of the energy levels 
in the uncondensed phase, obtained as the
largest root of $\Sigma(\epsilon_{*})=0$, and the condensation
threshold. Notice that we are seeking for the maximum energy, by definition of the free-energy function in Eq. \eqref{eq:free_ene}. In the uncondensed phase, i.e. when $\lambda<\lambda_{*}(\beta)$,
the dominating energy level $\tilde{\epsilon}(\lambda)$ is obtained
as the stationary point of $\lambda\epsilon-s(\epsilon)$,
and by the Legendre transform definition of $\zeta(\lambda)$
this is equivalent to $\tilde{\epsilon}(\lambda)=\zeta'(\lambda)$.
The entropy of the dominating state can be rewritten as $\Sigma(\tilde{\epsilon}(\lambda))=\beta-s(\tilde{\epsilon}(\lambda))=\beta+\zeta(\lambda)-\lambda\zeta'(\lambda)$,
so the condensation threshold $\lambda_{*}(\alpha)$ is obtained from
the condensation condition
\begin{equation}
  \beta+\zeta(\lambda_{*})-\lambda_{*}\zeta'(\lambda_{*})=0.  
\end{equation}
Finally, the free energy is given by
\begin{equation}
\phi_{\beta}(\lambda)=\begin{cases}
\frac{\beta+\zeta(\lambda)}{\lambda} & \lambda<\lambda_{*}(\beta),\\
\epsilon_{*}(\alpha) & \lambda\geq\lambda_{*}(\beta).
\end{cases}\label{eq:phi-rem}
\end{equation}

\subsection{REM analysis of the Guided Potential}

The guided contribution can be re-expressed in terms of the free-energy of a Random Energy Model (REM) \cite{Derrida1981}. The guided potential now reads
\begin{equation}
    V_{\text{guided}}(\boldsymbol{x}_t)\approx -w\left[d\beta(d)-\log\left(1+e^{d\phi_t(\boldsymbol{\boldsymbol{x}_t|\boldsymbol{c}^1)}}\right)\right].
\end{equation}
where the REM free-energy is given by
\begin{equation}
\phi_{t}(\boldsymbol{x}_t|\boldsymbol{c}^1)=\begin{cases}
\beta(d)+\zeta_{t,1}(\boldsymbol{x}_t|\boldsymbol{c}^1) & 1<\lambda_{*}(\beta,d,\sigma^2,t),\\
\zeta_{t,\lambda_*}^{'}(\boldsymbol{x}_t|\boldsymbol{c}^1)& 1\geq\lambda_{*}(\beta,d,\sigma^2,t).
\end{cases}\label{eq:phi-rem-guided}   
\end{equation}
The moment-generating function reads
\begin{align}
\zeta_{t,\lambda}(\boldsymbol{x}_t|\boldsymbol{c}^1) &= \lim_{d \to \infty}\frac{1}{d}\log\left(\mathbb{E}_{\boldsymbol{c}}\;e^{-\frac{\lambda}{2(\sigma^2+t)}\left(\|\boldsymbol{x}_t-\boldsymbol{c}\|^2-\|\boldsymbol{x}_t-\boldsymbol{c}^1\|^2\right)} \right)  \\&= \frac{\lambda}{2(\sigma^2+t)}\lim_{d \to \infty}\frac{\|\boldsymbol{c}^1\|^2}{d} -\frac{\lambda}{\sigma^2+t}\lim_{d \to \infty}\frac{\boldsymbol{x}_t\cdot \boldsymbol{c}^1}{d}
\\&+\lim_{d \to \infty}\log\left(\int \frac{\text{d}\boldsymbol{c} }{(2\pi)^{d/2}}e^{-\frac{1}{2}\left(1+\frac{\lambda}{\sigma^2+t}\right)\|\boldsymbol{c}\|^2+\frac{\lambda}{\sigma^2+t}\boldsymbol{x}_t\cdot \boldsymbol{c}}\right)\\
&= \frac{\lambda}{2(\sigma^2+t)}\lim_{d\to \infty}\frac{\|\boldsymbol{x}_t-\boldsymbol{c}^1\|^2}{d}-\frac{1}{2}\log\left(1+\frac{\lambda}{\sigma^2+t} \right)\\
&-\frac{\lambda}{2(\sigma^2+t+\lambda)}\lim_{d \to \infty}\frac{\|\boldsymbol{x}_t\|^2}{d},
\end{align}
while its derivative is
\begin{equation}
\zeta'_{t,\lambda}(\boldsymbol{x}_t|\boldsymbol{c}^1)
= \frac{1}{2(\sigma^2+t)}\lim_{d\to \infty}\frac{\|\boldsymbol{x}_t-\boldsymbol{c}^1\|^2}{d}-\frac{1}{2(\sigma^2+t+\lambda)}-\frac{\sigma^2+t}{2(\sigma^2+t+\lambda)^2} \lim_{d\to\infty}\frac{\|\boldsymbol{x}_t\|^2}{d}.
\end{equation}
The condition for finding the threshold variable $\lambda_* (d)$ reads
\begin{align}
\label{eq:condensation}
\beta(d) + \zeta_{t,\lambda_*}-\lambda_*\zeta^{'}_{t,\lambda_*}&=\beta(d)-\frac{1}{2}\,\ln\!\left(1+\frac{\lambda_*}{\mathrm{\sigma^2}+t}\right)\\
&+\frac{\lambda_* }{2(\mathrm{\sigma^2}+t+\lambda_*)}\left(1-\frac{\lambda_* }{\mathrm{\sigma^2}+t+\lambda_*}\lim_{d\to\infty}\frac{\|\boldsymbol{x}\|^2}{d}\right)
= 0.
\end{align}
According to the physics of REM, the model enters a \textit{conditional} phase as soon as $\phi_t(\boldsymbol{x}|\boldsymbol{c}^1) \leq 0$. Otherwise, the model explores an \textit{guided} phase if $\lambda_* > 1$ or a \textit{condensed} one if $\lambda_* \leq 1$.
The expression for the diffusion potential in terms of the REM moment-generating function reads
\begin{equation}
V_{\text{eff}}(\boldsymbol{x}_t)=\begin{cases}
\frac{1}{2}\frac{\|\boldsymbol{x}_t-\boldsymbol{c}^1\|^2}{\sigma^2+t}+\frac{d}{2}\log(2\pi t)+dw \;\zeta_{t,1}(\boldsymbol{x}_t|\boldsymbol{c}^1)& \text{(guided phase)}\\
-\frac{1}{2}\frac{\|\boldsymbol{x}-\boldsymbol{c}^1\|^2}{\sigma^2+t}-\frac{N}{2}\log(2\pi t)+w\beta N & \text{(conditional phase)}\\
-\frac{1}{2}\frac{\|\boldsymbol{x}-\boldsymbol{c}^1\|^2}{\sigma^2+t}-\frac{N}{2}\log(2\pi t)+wN\left[\beta -\zeta^{'}_{t,\lambda_*}(\boldsymbol{x}|\boldsymbol{c}^1)\right]& \text{(condensed phase)},
\end{cases}\label{eq:diff_pot}   
\end{equation}
while full effective potential becomes (in the relative phases, that we do not indicate for space limitations):
\begin{equation}
V_{\text{eff}}(\boldsymbol{x}_t)=\begin{cases}
-\frac{(1+w)}{2}\frac{\|\boldsymbol{x}-\boldsymbol{c}^1\|^2}{\sigma^2+t}+\frac{w \|\boldsymbol{x}\|^2}{2(\sigma^2+t+1)}-\frac{wN}{2}\log\left(1+\frac{1}{\sigma^2+t}\right)-\frac{N}{2}\log(2\pi t) \\
-\frac{1}{2}\frac{\|\boldsymbol{x}-\boldsymbol{c}^1\|^2}{\sigma^2+t}-\frac{N}{2}\log(2\pi t)+w\beta N  \\
-\frac{(1+w)}{2}\frac{\|\boldsymbol{x}-\boldsymbol{c}^1\|^2}{\sigma^2+t}-\frac{N}{2}\log(2\pi t)+w \left[\beta N
+\frac{\sigma^2+t}{2(\sigma^2+t+\lambda_*)^2} \|\boldsymbol{x}\|^2 +\frac{1}{2(\sigma^2+t+\lambda_*)}\right]
\end{cases}\label{eq:diff_pot2_app}   
\end{equation}
The local minimum of the potential is unique at any time, and it is given by
\begin{equation}
    \boldsymbol{x}^*(t)=
    \begin{cases}
    \frac{(1+w)(\sigma^2+t+1)}{w+\sigma^2+t+1}\;\boldsymbol{c}^1& \text{(guided phase)}\\
\boldsymbol{c}^1 & \text{(conditional phase)}\\
\left[1+\frac{w}{1+w}\left(\frac{\sigma^2+t}{\sigma^2+t+\lambda_*}\right)^2\right]^{-1}\boldsymbol{c}^1
& \text{(condensed phase)}.
\end{cases}
\label{eq:minimum}
\end{equation}
In absence of correlated centroids, we expect the system to never enter the condensed phase of the potential, as a consequence of the isotropic nature of the stochastic process: the condensation and collapse transitions of the relative REM are going to coincide (see \cite{achilli2025memorization} for a more rigorous justification based on the Nishimori conditions applied to the problem).
Hence the transition between the guided phase and the conditional one occurs when the REM free-energy of the model changes sign. 
We name the transition time, in continuity with the literature, \textit{speciation time} $t_s$ \cite{biroli_dynamical_2024}. Our transition condition thus reads
\begin{equation}
\label{eq:collapse_CFG2_bis}
    \lim_{d \to \infty}\left[\beta(d) +\zeta_{t_s}(\sigma^2,w)\right] = 0,
\end{equation}
where $\zeta_{t}$ depends on the specific trajectory $\mathbf{x}_t$ and also on the norm of $\mathbf{c}^1$.

\subsection{REM analysis of the Unconditional Potential}
\label{app:uncond}
\begin{figure}[ht!]
    \centering
    \includegraphics[width=0.33\linewidth]{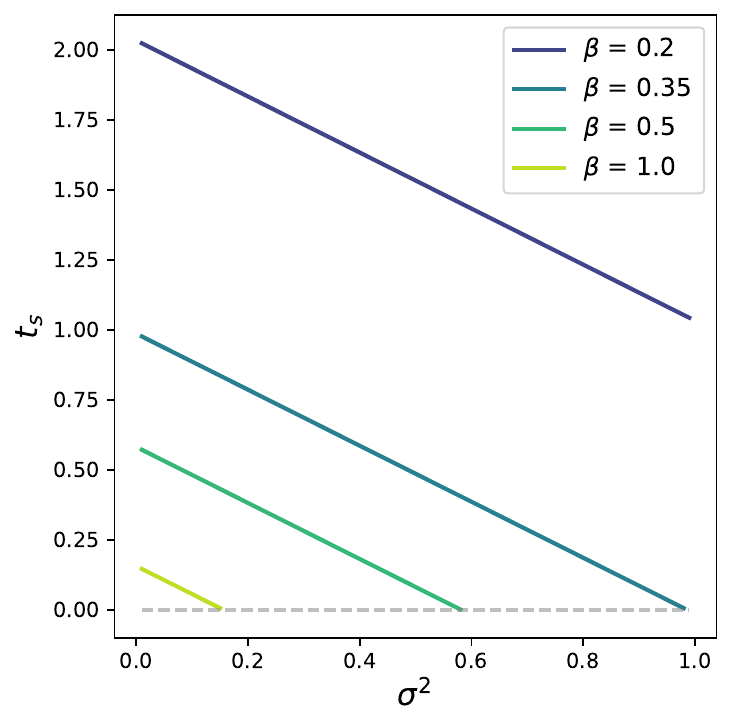}\includegraphics[width=0.33\linewidth]{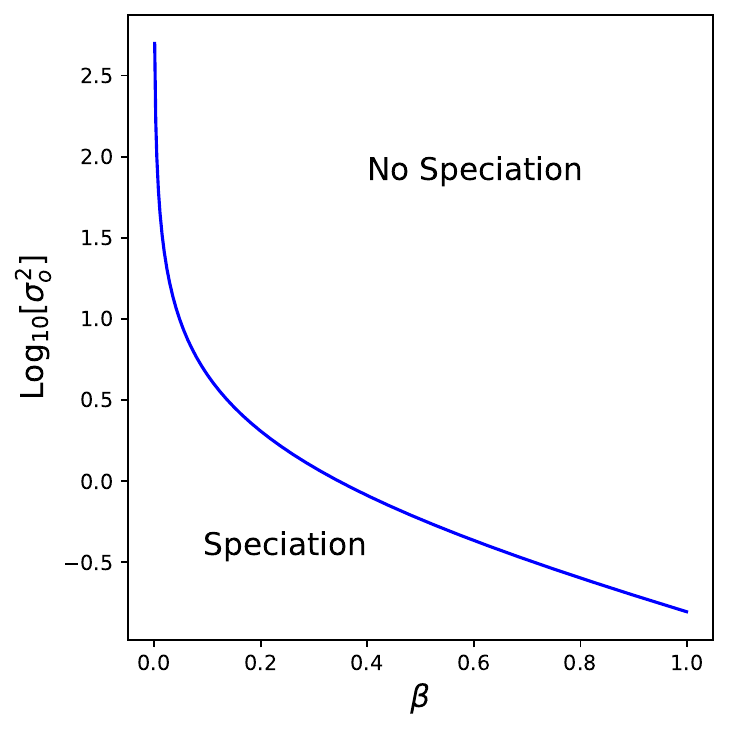}\includegraphics[width=0.33\linewidth]{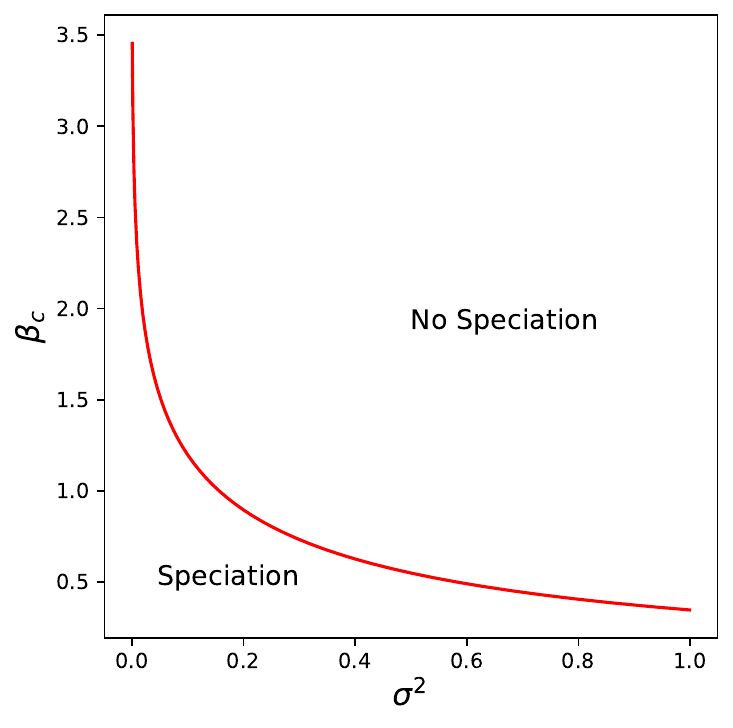}
    \caption{Left: speciation time $t_s$ as a function of the variance of the target distribution for multiple amounts of centroids in the target Gaussian mixture. For each value of $\beta$ there exists one $\sigma_o^2$ such that $t_s = 0$. Center: onset variance $\sigma_o^2$ in log-scale as a function of $\beta$. The more centroids there are, the lower is $\sigma_o^2$, i.e. the easier is for the centroids not to be local maxima of the Gaussian mixture. Right: critical entropy of centroids $\beta_c$ as a function of $\sigma^2$. The larger is $\sigma^2$, the lower is $\beta_c$, i.e. the easier is for the centroids not to be local maxima of the Gaussian mixture.}
    \label{fig:spec}
\end{figure}

Let us consider a diffusion model with a target distribution being the distribution in Eq. \eqref{eq:target_uncond}, as it would be without any conditioning on a class. 
One can now study the exact diffusion model represented by the density function in Eq.\eqref{eq:samp_pdf} in terms of a REM. Moreover, we can define the \textit{speciation time} $t_s$, i.e. the time the backward diffusive trajectory gets trapped into one of the modes of the target distribution, as the collapse time of the relative REM.
Thus, we aim at computing the moment generating function of $p_t(\boldsymbol{x})$, namely the observable $\zeta_t(\lambda)$, and impose the following collapse condition
\begin{equation}
    \label{eq:collapse}
    \zeta_{t_s}(1)+\beta = -\frac{1}{2}, 
\end{equation}
where
\begin{equation}
    \zeta_t(\lambda) = \lim_{d\to\infty}\frac{1}{d}\mathbb{E}_{\boldsymbol{c}^1, \boldsymbol{\omega}}\log{\mathbb{E}_{\boldsymbol{c}}\;e^{-\frac{\lambda}{2(\sigma^2 + t)}\|\boldsymbol{c}^1 + \boldsymbol{\omega}\sqrt{\sigma^2 + t} -\boldsymbol{c}\|^2}}.
\end{equation}
The final expression  of the moment generating function reads
\begin{equation}
\label{eq:zeta2}
\zeta_t(\lambda) = -\frac{1}{2}\log\left(1+\frac{\lambda}{\sigma^2+t}\right)-\frac{\lambda}{2}\frac{1+\sigma^2+t}{\lambda+\sigma^2+t}   
\end{equation}
and the full collapse condition (that equals the condensation condition in this case) is
\begin{equation}
2\beta=\log\left(1+\frac{1}{\sigma^2+t_s} \right),
\end{equation}
from which we get
\begin{equation}
\label{eq:ts}
t_s(\beta,\sigma^2) = \frac{1}{e^{2\beta}-1}-\sigma^2.
\end{equation}
The speciation time is plotted in Figure \ref{fig:spec}, left panel. By imposing $t_s = 0$ we can extract two important quantities: 
\begin{itemize}
    \item The onset variance 
    \begin{equation}
    \label{eq:sigma_o}
     \sigma^2_o(\beta) = \frac{1}{e^{2\beta}-1}. 
     \end{equation}
     Given $\beta$, we can observe speciation for $\sigma^2 < \sigma^2_o(\beta)$. For larger values of the variance, $p_0$ does not admit any separation among the centroids (see Figure \ref{fig:spec}, central panel).
    \item The critical entropy of the centroids 
    \begin{equation}
        \beta_c(\sigma^2) = \frac{1}{2}\log\left(1+\frac{1}{\sigma^2}\right).
    \end{equation}
    Given the variance $\sigma^2$, we can observe speciation for $\beta < \beta_c(\sigma^2)$. Beyond this critical value of $\beta$, $p_0$ does not admit any separation among the centroids (see Figure \ref{fig:spec}, right panel). 
    
\end{itemize}

\subsubsection{Comparison between Guided \& Unconditional Phase Transitions}
We have seen that the unconditional model undergoes an ergodicity breaking event in sampling time that prior literature has named \textit{speciation transition} \cite{biroli_dynamical_2024}. According to this event, the diffusive trajectory spontaneously chooses one among the possible $M$ classes of the target distribution, and this is related to the collapse / condensation of the partition function of a REM into a sub-exponential number of energy levels.

On the other hand, the transition underwent by the guided model shares similarities with the traditional speciation transition, yet it does not signal an ergodicity breaking the diffusion potential, because the diffusion potential, in our setting, has one single minimum. 
The analogy derives from the structure of the relative REMs between the two setups, that are very similar. 
In both cases, the speciation transition separates a phase (large noise levels) where all the classes contribute to the potential, from another phase (small noise levels) where only the conditioning class $c^1$ participates to the sampling probability measure.
We then observe that the speciation times $t_s$ derived for guided and unconditional models share the same qualitative behavior with respect to the control parameters $\beta, \sigma^2$, specifically, the transition disappears when modes in the Gaussian mixture are fully merged. However, the guided speciation time also depends on the guidance level $w$, which makes the sampling phenomenology more complex, as showed in the next Sections. 

We stress that the work by \cite{biroli_meta} observe the same phenomenology for a simplest target data model, where they are capable of identifying a direct correspondence between the unconditional and guided phase transitions. In our setup drawing such exact correspondence is not possible. 

\section{Mean Field description of CFG}
\label{sec:mf_CFG}

We want to solve the system of $d$ independent SDEs represented by

\begin{equation}
    d\mathbf{x}_t = \mathbf{x}_t a(t) dt + b(t)dt + d\mathbf{w}_t,
\end{equation}
where functions $a(t), b(t)$ change from phase to phase in the REM analysis. Such SDEs can be solved component by component using the integrating factor method. The procedure corresponds to first define an auxiliary function
\begin{equation}
    \Phi(t) = \exp\left[ - \int_0^s ds\; a(s)\right].
\end{equation}
By simplicity, let us consider one single component of the vectors and drop the index label. Each SDE now reads

\begin{equation}
    x_t = \Phi^{-1}(t)\Phi(T)\;x_T + \Phi(t)^{-1}\int_T^t ds\;\Phi(s)b(s) +\Phi^{-1}(t)\int_{T}^t ds\;\Phi(s)\xi(s), 
\end{equation}
where we have rewritten the noise term as $d w_t = \boldsymbol{\xi}(t)dt$ with $\langle \xi(t)\rangle_{\xi} = 0$ and $\langle \xi(t)\xi(t')\rangle_{\xi}=\delta(t-t')$. 
Let us derive, in Sections \ref{sec:conditional}, \ref{sec:guided}, \ref{sec:condensed} the solution $\mathbf{x}_t$, and its first two moments, separately in the three dynamic phases obtained from the REM analysis of the diffusion potential. Such moments are defined with respect to the measure $\langle \;\cdot\; \rangle_{\xi}$ as 
\begin{align}
    \boldsymbol{\mu}(t) &= \langle \mathbf{x}(t)\rangle_{\xi}\\
    \sigma^2(t) &= \langle \| \mathbf{x}(t)\|^2\rangle_\xi - \|\boldsymbol{\mu}(t)\|^2.
\end{align}
Section \ref{sec:assemb} will be devoted to put all the pieces together and show how the entire trajectory, across the three phases, is derived. 
\subsection{Conditional phase \label{sec:conditional}}

In this phase SDEs read
\begin{equation}
    dx = \frac{x}{\sigma^2+t}dt-\frac{c^1}{\sigma^2+t}dt + dW_t
\end{equation}
The auxiliary functions read
\begin{equation}
    a(t) = \frac{1}{\sigma^2+t},\hspace{1cm}
    b(t) = -\frac{c^1}{\sigma^2+t},\hspace{1cm}
    \Phi(t) = \frac{\sigma^2}{\sigma^2+t}.
\end{equation}
The evolution equation reads

\begin{equation}
    x_t = \frac{\sigma^2+t}{\sigma^2 + T}x_T+\frac{T-t}{\sigma^2+T}c^1+(\sigma^2+t)\int_{T}^t ds\; \frac{\xi(s)}{\sigma^2 + s},
\end{equation}
where $x_T$ is the initial condition for the integration. From the isotropy of the process, the solution $x_t$ must be normally distributed with mean
\begin{equation}
    \mu(t) = \frac{\sigma^2+t}{\sigma^2 + T}\mu(T)+\frac{T-t}{\sigma^2+T}c^1,
\end{equation}
and variance 
\begin{equation}
    \sigma^2(t) = \left(\frac{\sigma^2+t}{\sigma^2+T}\right)^2\sigma^2(T)+ (T-t)\frac{\sigma^2+t}{\sigma^2+T}. 
\end{equation}

\subsection{Guided Phase \label{sec:guided}}

In this phase SDEs read
\begin{equation}
    dx = x\frac{\sigma^2+t+1+w}{(\sigma^2+t)(\sigma^2+t+1)}dt-c^1\frac{(1+w)}{(\sigma^2+t)}dt + dW_t
\end{equation}
The auxiliary functions for the overlap read as
\begin{equation}
    a(t) = \frac{\sigma^2+t+1+w}{(\sigma^2+t)(\sigma^2+t+1)},\hspace{0.6cm}
    b(t) = -c^1\frac{(1+w)}{(\sigma^2+t)},\hspace{0.6cm}
    \Phi(t) = \frac{\sigma^2}{\sigma^2+t}\left[\frac{\sigma^2+1+t}{\sigma^2+1}\frac{\sigma^2}{\sigma^2+t}\right]^w.
\end{equation}
The evolution equation for the overlap thus reads as follows
\begin{align}
    x_t &= x_T\left(\frac{\sigma^2+t}{\sigma^2+T}\right)\left[\frac{\sigma^2+t}{\sigma^2+T}\frac{\sigma^2+1+T}{\sigma^2+1+t}\right]^w \\ &+c^1\frac{(\sigma^2+t)^{(1+w)}}{(\sigma^2+t+1)^w} \left[\left(1+\frac{1}{\sigma^2+t}\right)^{1+w}-\left(1+\frac{1}{\sigma^2+T}\right)^{1+w}\right] \\&+\frac{(\sigma^2+t)^{w+1}}{(\sigma^2+1+t)^w} \int_{T}^t ds\;\xi(s)\frac{(\sigma^2+1+s)^w}{(\sigma^2+s)^{w+1}}
\end{align}
where $x_T$ is the initial condition for the integration. From the isotropy of the process, the solution $x_t$ must be normally distributed with mean
\begin{align}
    \mu(t) &= \left(\frac{\sigma^2+t}{\sigma^2+T}\right)\left[\frac{\sigma^2+t}{\sigma^2+T}\frac{\sigma^2+1+T}{\sigma^2+1+t}\right]^w \mu(T) \\
    &+ \frac{(\sigma^2+t)^{(1+w)}}{(\sigma^2+t+1)^w} \left[\left(1+\frac{1}{\sigma^2+t}\right)^{1+w}-\left(1+\frac{1}{\sigma^2+T}\right)^{1+w}\right]c^1,
\end{align}
and variance 
\begin{align}
    \sigma^2(t) &= \left(\frac{\sigma^2+t}{\sigma^2+T}\right)^2\left[\frac{\sigma^2+t}{\sigma^2+T}\frac{\sigma^2+1+T}{\sigma^2+1+t}\right]^{2w}\sigma^2(T)\\
    &+  \frac{(\sigma^2+t)^{2w+2}}{(\sigma^2+t+1)^{2w}}\int_{T}^t ds\;\frac{(\sigma^2+1+s)^{2w}}{(\sigma^2+s)^{2w+2}} \\ &=\left(\frac{\sigma^2+t}{\sigma^2+T}\right)^2\left[\frac{\sigma^2+t}{\sigma^2+T}\frac{\sigma^2+1+T}{\sigma^2+1+t}\right]^{2w}\sigma^2(T)&\\&+\frac{(\sigma^2+t)^{2w+2}}{(\sigma^2+t+1)^{2w}}\frac{1}{2w+1}\left[\left(1+\frac{1}{\sigma^2+t}\right)^{2w+1}-\left(1+\frac{1}{\sigma^2+T}\right)^{2w+1}\right]. 
\end{align}

\subsection{Condensed Phase \label{sec:condensed}}
In this phase, SDEs read
\begin{equation}
    dx = x\left[\frac{1+w}{\sigma^2+t}-\frac{w(\sigma^2+t)}{(\sigma^2+t+\lambda_*(t))^2}\right]dt-c^1\frac{(1+w)}{(\sigma^2+t)}dt + dW_t
\end{equation}

The auxiliary functions for the overlap read as
\begin{equation}
    a(t) = \frac{1+w}{\sigma^2+t}-\frac{w(\sigma^2+t)}{(\sigma^2+t+\lambda_*(t))^2},\hspace{0.6cm}
    b(t) = -c^1\frac{(1+w)}{(\sigma^2+t)},\hspace{0.6cm}
    \Phi(t) = \exp\left[-\int_0^t ds\; a(s)\right].
\end{equation}
The integrating factor cannot be computed in closed form, since $\lambda_*(t)$ is derived, for each time value, through the implicit function in Eq.~\eqref{eq:condensation}. As a consequence this ODE must be integrated numerically.
\subsection{Assembling the whole Diffusive Trajectory and Measuring Distortion \label{sec:assemb}}
In order to derive the full trajectory of the system in the ambient space, as well as the distortion as a function of time, one has to take into account the transition among different phases of the potential. \\
All trajectories will be initialized at the time horizon $T \to \infty$. We can infer that, in this limit, the system always starts from the guided phase. In fact, from Eq.~\eqref{eq:condensation}, one has

\begin{equation}
\lim_{T \to \infty}\frac{1}{2}\,\ln\!\left(1+\frac{\lambda_*}{\mathrm{\sigma^2}+T}\right)
-\frac{\lambda_*}{2(\mathrm{\sigma^2}+T+\lambda_*)}\left(1-\frac{\lambda_* }{\mathrm{\sigma^2}+T+\lambda_*}\lim_{d\to\infty}\frac{\|\mathbf{x}\|^2}{d}\right)
= \beta.
\end{equation}
By choosing the variance for $\mathbf{x}(T)$ to be $\mathcal{O}(T)$, then the same condition becomes
\begin{equation}
    \lim_{T \to \infty} \frac{\lambda_*(\lambda_*-1)}{2T} = \beta,
\end{equation}
As a consequence, the system always starts diffusing from the guided phase. 

At this point we need to compute the transition time from guided phase to the conditional one, by solving the implicit equation \eqref{eq:collapse_CFG2_bis}. The explicit expression of the moment-generating function in the guided phase reads
\begin{align}
\zeta_t(\sigma^2,w)
&= \frac{1}{2}\Biggl(
\frac{(\sigma^2+t)^{2w+1}}{(\sigma^2+t+1)^{2w}}
-\frac{(\sigma^2+t)^{2w+2}}{(\sigma^2+t+1)^{2w+1}}
\Biggr) \notag\\
&\quad{}\times
\Biggl[
\frac{1}{2w+1}
\biggl[
\biggl(1+\frac{1}{\sigma^2+t}\biggr)^{2w+1}-1
\biggr]
+
\biggl[
\biggl(1+\frac{1}{\sigma^2+t}\biggr)^{1+w}-1
\biggr]
\Biggr] \notag\\
&\quad{}
+\frac{1}{2(\sigma^2+t)}
-\frac{(\sigma^2+t)^w}{(\sigma^2+t+1)^w}
\biggl[
\biggl(1+\frac{1}{\sigma^2+t}\biggr)^{1+w}-1
\biggr] \notag\\
&\quad{}
-\frac{1}{2}\log\biggl(1+\frac{1}{\sigma^2+t}\biggr).
\end{align}

Generally speaking, the order of magnitude of such time with respect to the ambient dimension $d$ will depend on the nature of $\beta(d)$. 
We can recover the scaling behavior of $t_s$ Vs $d$ from a simple dimensional reasoning. Let us notice that $\zeta_{t}(\sigma^2,w)$ decreases monotonically in time and $\zeta_{t}(w) = -\frac{1+w}{t}+\mathcal{O}\left(\frac{1}{t^2}\right)$ at large times $t$. As a consequence, if $\lim_{d \to \infty}\beta(d) = 0$, then we must have 
\begin{equation}
    t_s(w,d) = \mathcal{O}\left(\frac{1+w}{\beta(d)}\right).
\end{equation}
This expression implies that the speciation time diverges when the number of modes in the mixture is sub-exponential. If the speciation time diverges, the system never effectively visits the guided phase, and we can envisage no final distortion of the target distribution. On the other hand, when the number of possible classes is exponentially large in $d$, $t_s = \mathcal{O}(1)$ and distortion is ensured. 

The assembly of the whole trajectory can be done by integrating the backward SDE in the following order:
\begin{enumerate}
    \item The \textbf{guided} phase SDE for $t \in [\;\text{max}(0,t_s(w,\beta,\sigma^2)),\; T\;]$ with initial conditions $\boldsymbol{\mu}(T) = \boldsymbol{0}$, $\sigma^2(T) = T$ and $T \to \infty$. 
    \item The \textbf{conditional} phase SDE for $t \in [\;0,\;\text{max}(0,t_s(w,\beta,\sigma^2))\;]$ with initial conditions $\boldsymbol{\mu}(T) = \boldsymbol{\mu}(t_s)$ and $\sigma^2(T) = \sigma^2(t_s)$.
\end{enumerate}
The operator $\text{max}(0,t_s)$ means that, when $t_s < 0$ from the analysis, the system displays no speciation, because the modes of the Gaussian mixture mutually merged, due to $\sigma^2$ or $\beta$ being too large. 

\begin{figure}[ht!]
    \centering
    \includegraphics[width=0.7\linewidth]{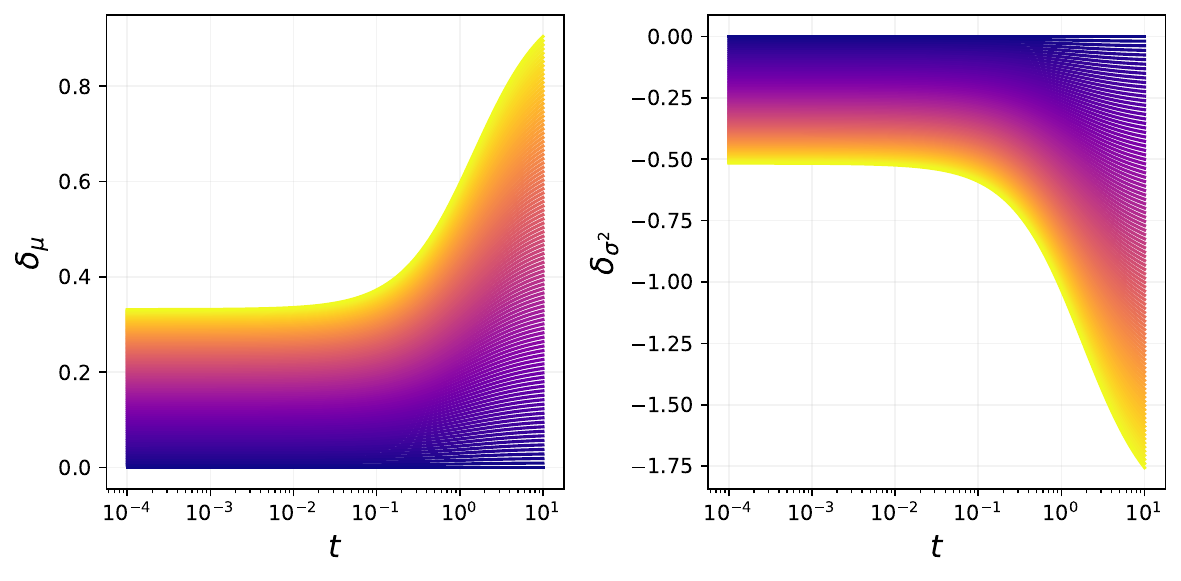}
     \caption{Distortion observables for a choice of $\sigma^2, \beta$ that do not allow for the transition from the guided to the conditional phases of the potential. The system ends its run inside the guided phase. The color-map for the curves indicates different values of the guidance level: the darkest curve indicates $w = 0$, the lightest one $w = 1$.}\label{fig:distort2}
\end{figure}

\begin{figure}[ht!]
    \centering

    \begin{subfigure}{0.3\linewidth}
        \centering
        \includegraphics[width=\linewidth]{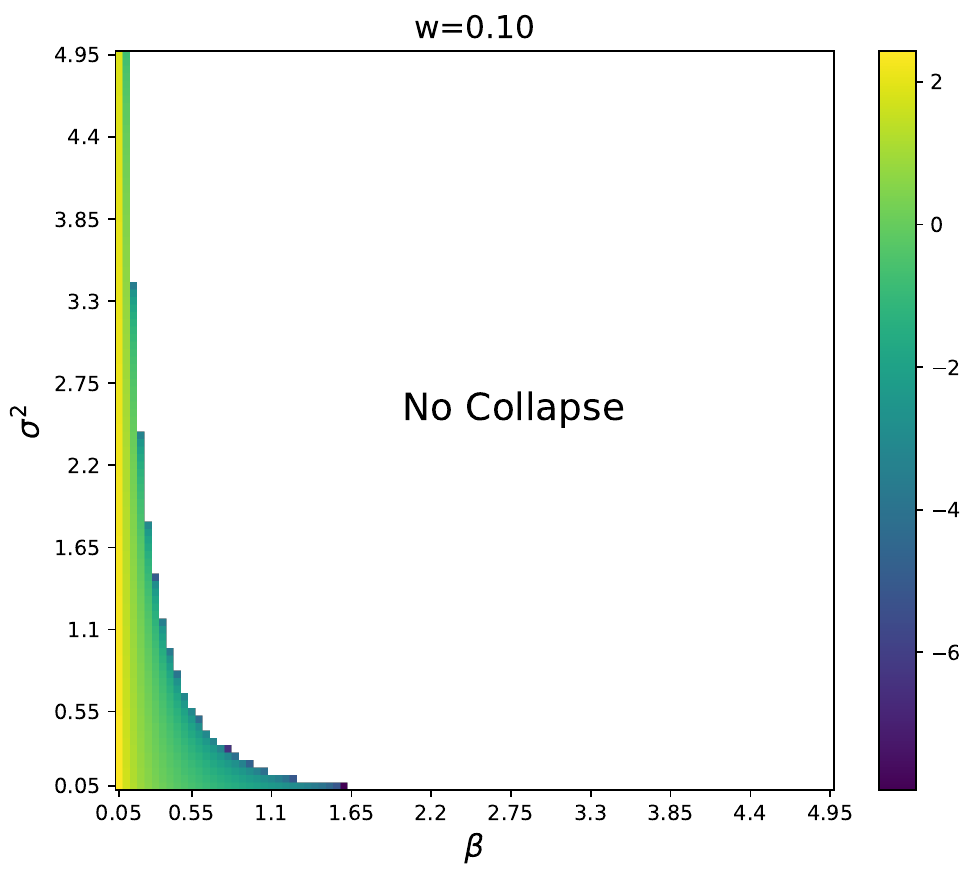}
        \caption{speciation time (log), $w=0.1$}
    \end{subfigure}
    \begin{subfigure}{0.3\linewidth}
        \centering
        \includegraphics[width=\linewidth]{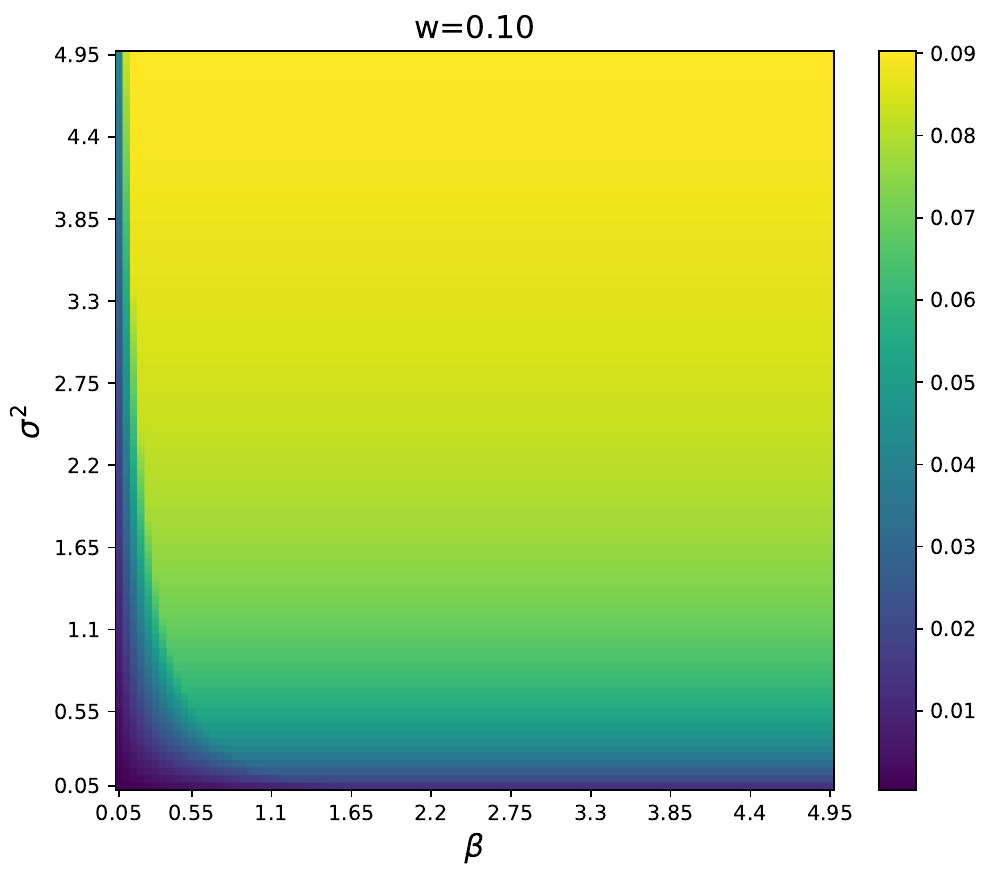}
        \caption{$\delta_{\mu}(0)$, $w=0.1$}
    \end{subfigure}
    \begin{subfigure}{0.3\linewidth}
        \centering
        \includegraphics[width=\linewidth]{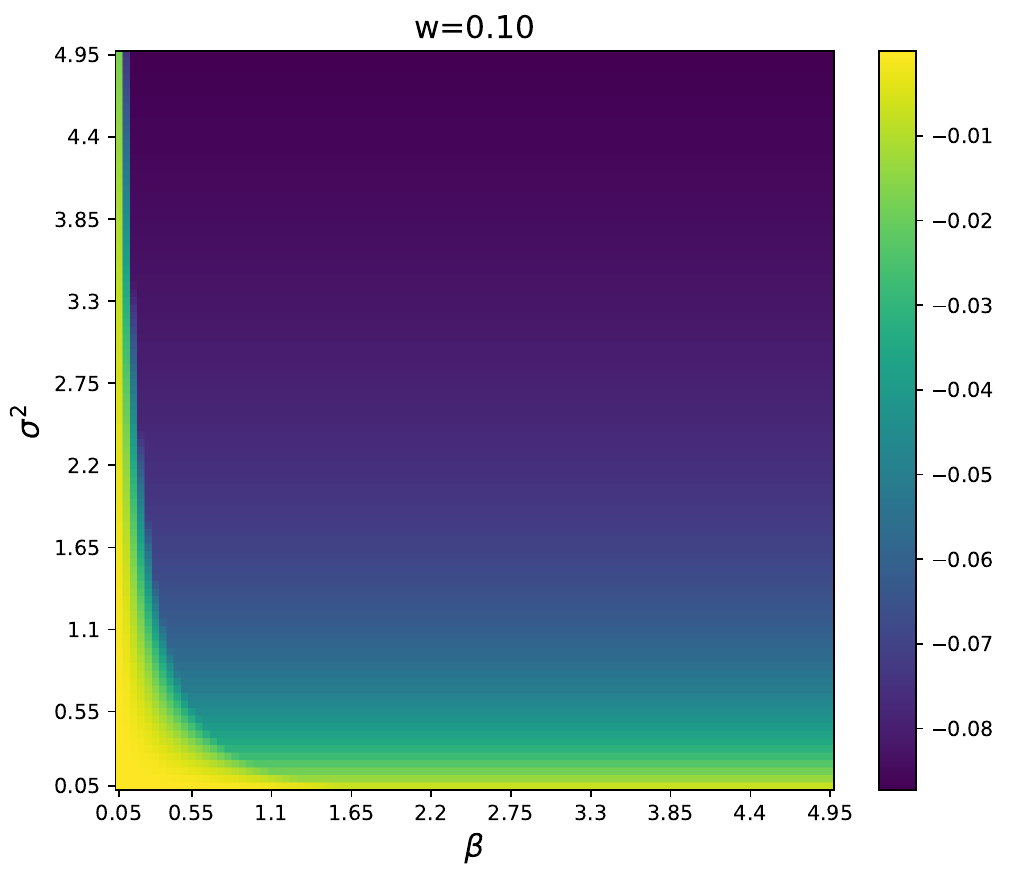}
        \caption{$\delta_{\sigma^2}(0)$, $w=0.1$}
    \end{subfigure}

    \vspace{0.5em}

    \begin{subfigure}{0.3\linewidth}
        \centering
        \includegraphics[width=\linewidth]{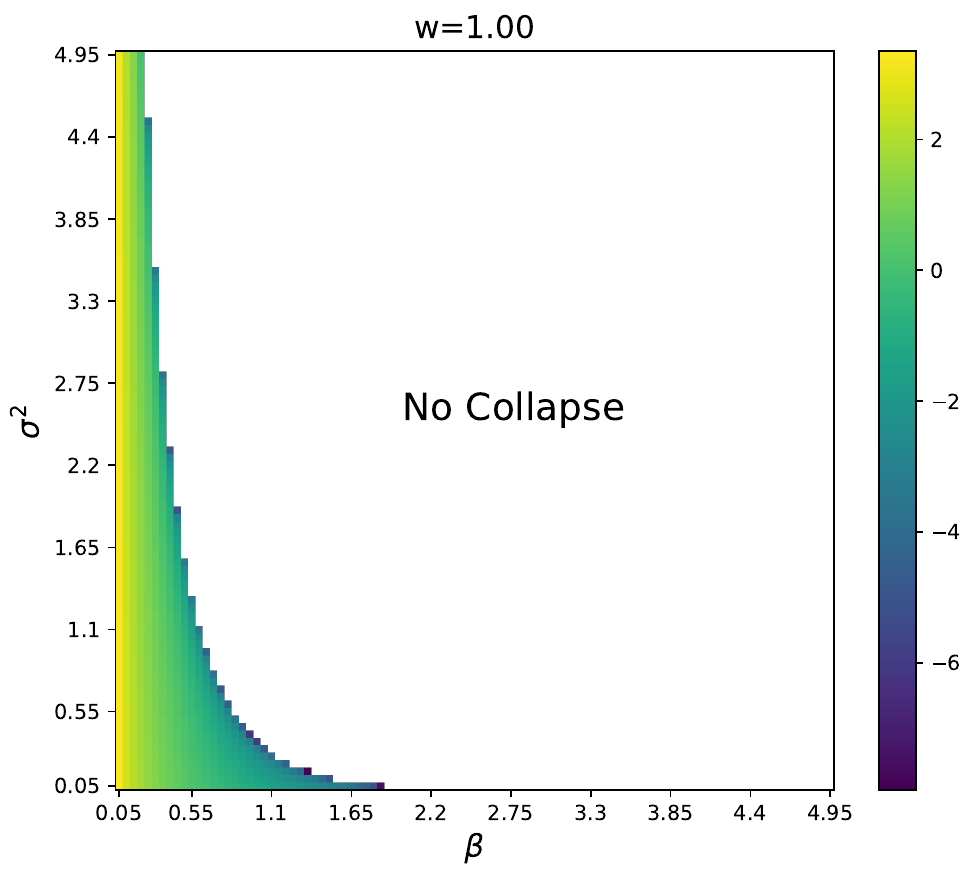}
        \caption{speciation time (log), $w=1$}
    \end{subfigure}
    \begin{subfigure}{0.3\linewidth}
        \centering
        \includegraphics[width=\linewidth]{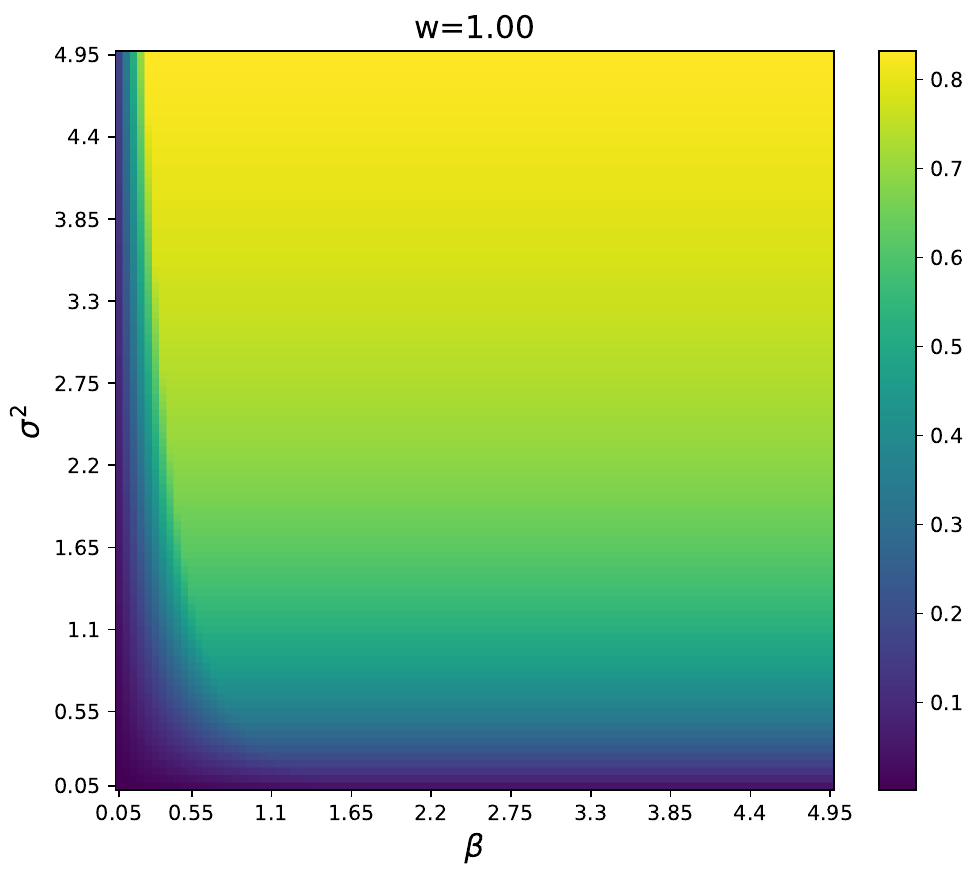}
        \caption{$\delta_{\mu}(0)$, $w=1$}
    \end{subfigure}
    \begin{subfigure}{0.3\linewidth}
        \centering
        \includegraphics[width=\linewidth]{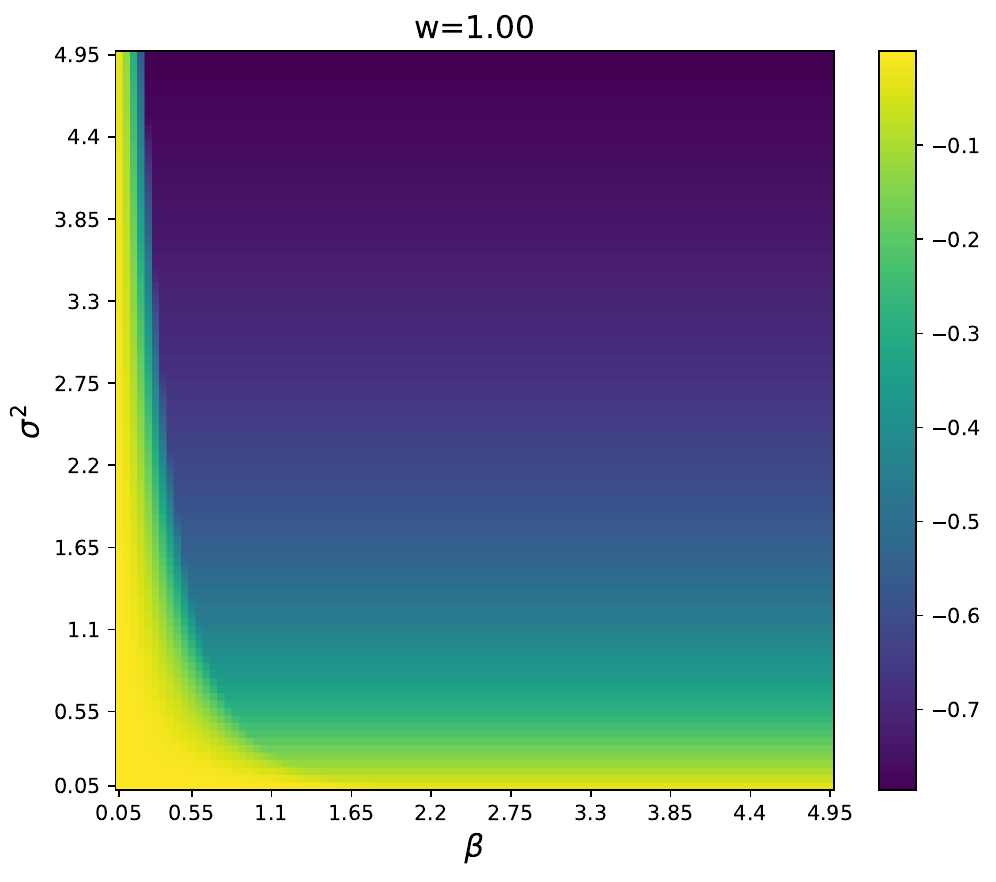}
        \caption{$\delta_{\sigma^2}(0)$, $w=1$}
    \end{subfigure}

    \vspace{0.5em}

    \begin{subfigure}{0.3\linewidth}
        \centering
        \includegraphics[width=\linewidth]{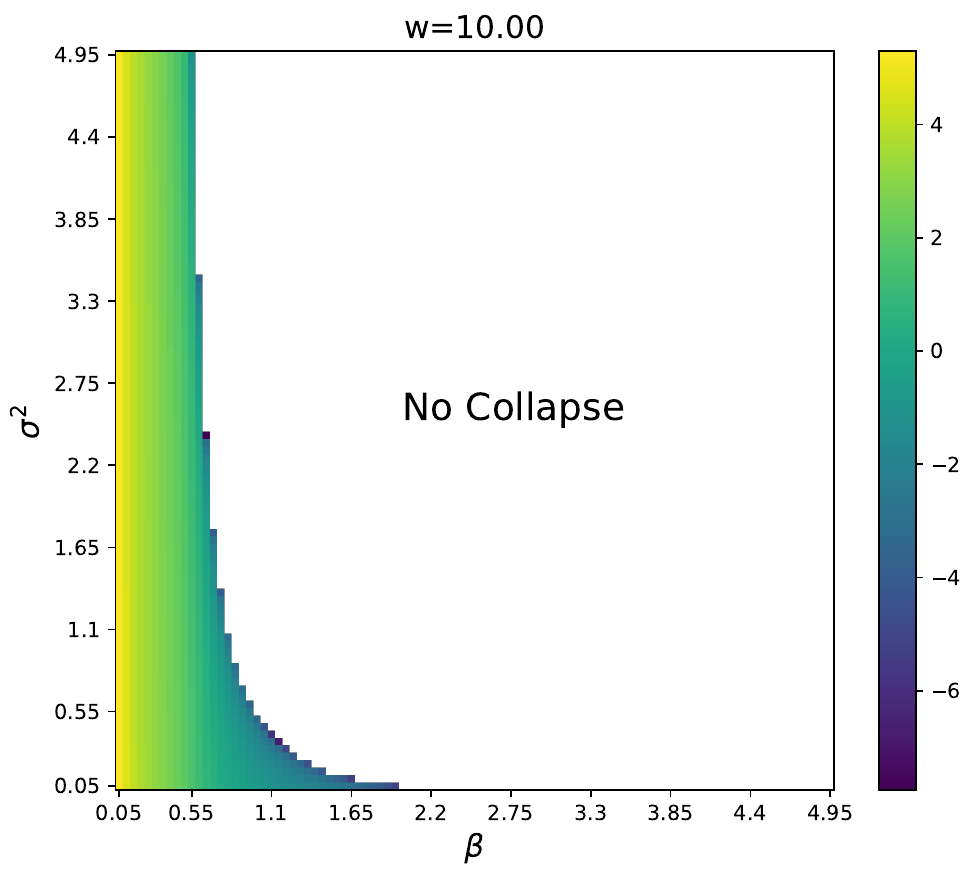}
        \caption{speciation time (log), $w=10$}
    \end{subfigure}
    \begin{subfigure}{0.3\linewidth}
        \centering
        \includegraphics[width=\linewidth]{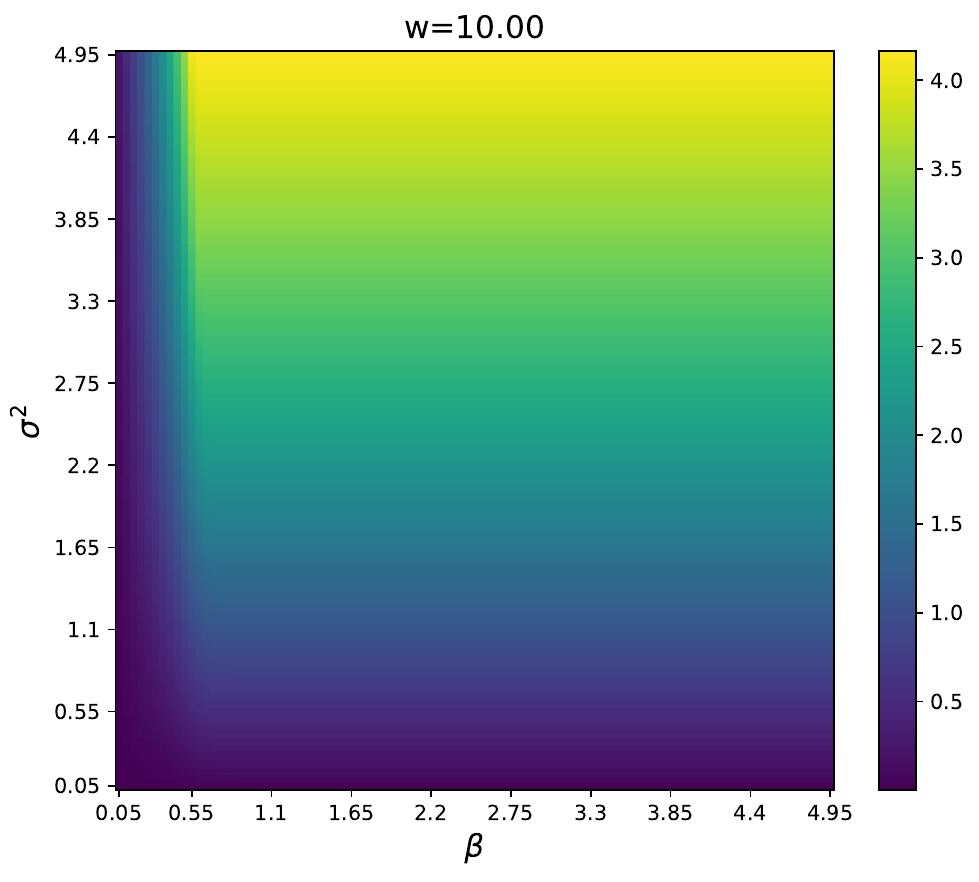}
        \caption{$\delta_{\mu}(0)$, $w=10$}
    \end{subfigure}
    \begin{subfigure}{0.3\linewidth}
        \centering
        \includegraphics[width=\linewidth]{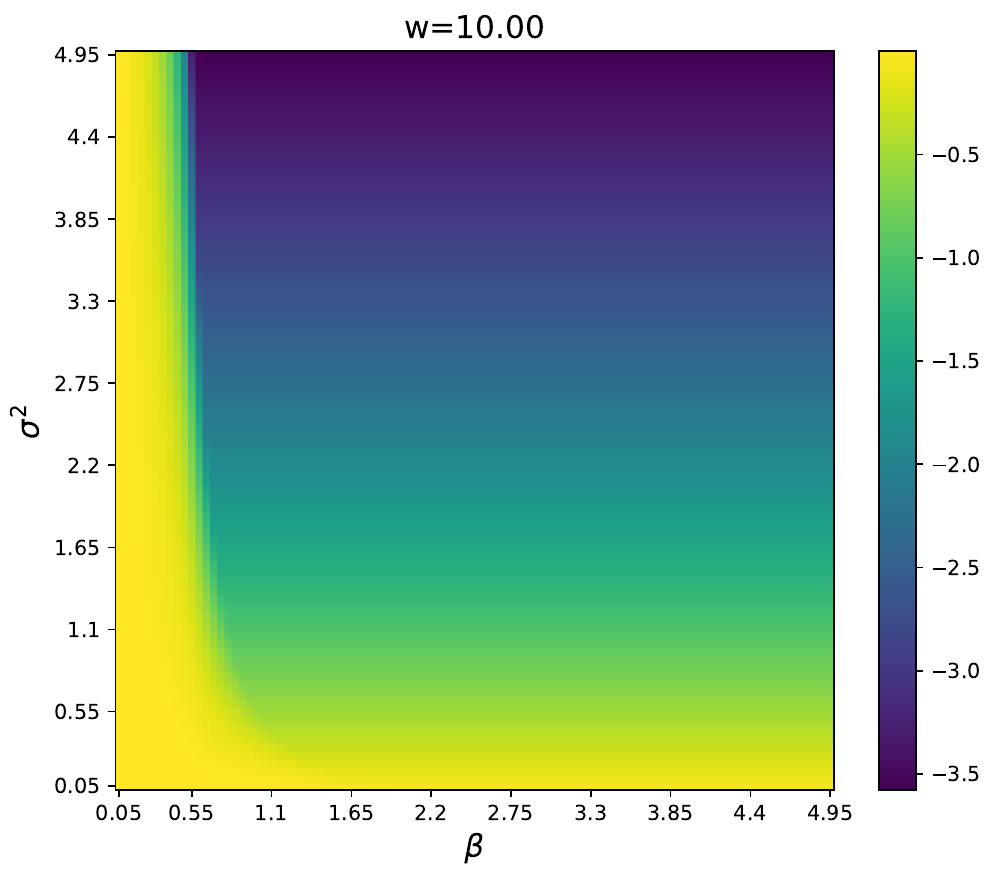}
        \caption{$\delta_{\sigma^2}(0)$, $w=10$}
    \end{subfigure}

    \caption{Heat-maps representing the distortion at $t = 0$ for various combinations of $\sigma^2, \beta, w$ when the target distribution is a Gaussian Mixture with an exponential mode count. The blank region in the speciation time panel indicates $t_s < 0$, i.e. no transition from the guided to the conditional phase.}
    \label{fig:combined_w}
\end{figure}

\begin{figure*}[t]
\centering
  \includegraphics[width=\linewidth]{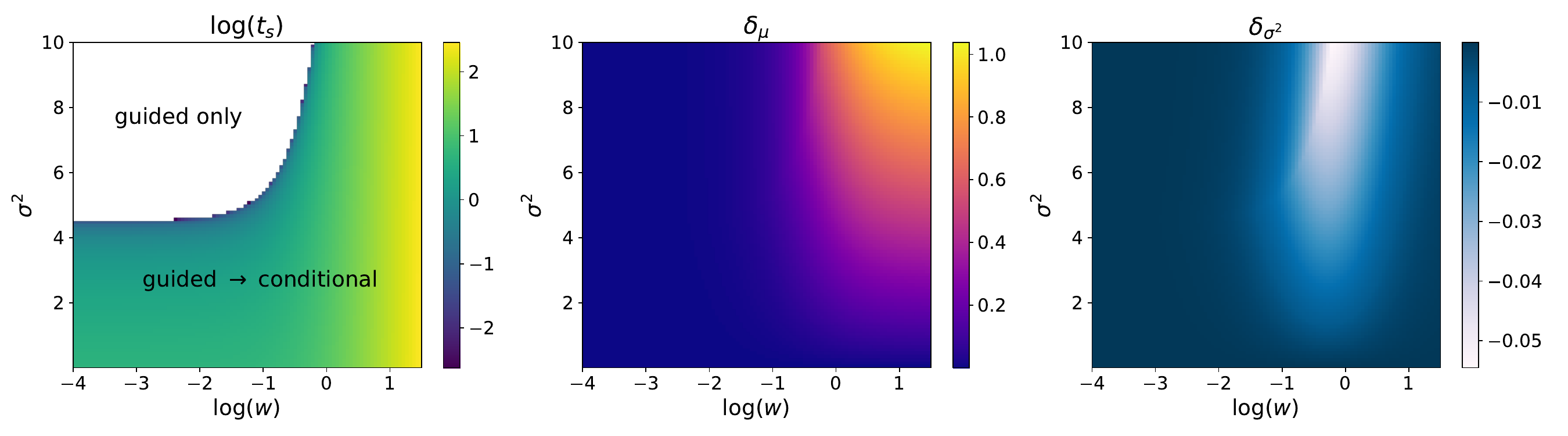}
\caption{Measure of the speciation time $t_s$ and the distortion estimators in the exponential regime as a functions of the control parameters $\sigma^2$ and $w$, for $\beta = 0.1$ at sampling time $t = 0$. The white region in the $t_s$ panels signal that a negative speciation time, or equivalently, the absence of transition to the conditional phase. While this aspect is correlated to a stronger deformation of the target distribution, the occurrence of such transition appears to imply with a weak deformation, as testified by the behavior of $\delta_{\mu}$ and $\delta_{\sigma^2}$.}
\label{fig:distor_exp_app}
\end{figure*}

Eventually, to quantify the distortion performed in time by CFG on the data target distribution we introduce the following two observables
\begin{equation}
\label{eq:delta_mu}
    \delta_{\mu}(t) = \lim_{d \to \infty}\frac{\mathbf{c}^1\cdot(\boldsymbol{\mu}_w(t)-\mathbf{c}^1)}{d},
\end{equation}
and 
\begin{equation}
\label{eq:delta_sigma}
    \delta_{\sigma^2}(t) =\frac{\sigma^2(t)-(\sigma^2+t)}{\sigma^2+t}.
\end{equation}
The observable in Eq.~\eqref{eq:delta_mu} quantifies the distortion of the mean, while observable in Eq.~\eqref{eq:delta_sigma} focuses on the distortion of the variance, always with respect to the conditional target distribution.    
Given these considerations, the starting distortion observables measure
\begin{align}
    \delta_{\mu}(t) &= \frac{(\sigma^2+t)^{1+w}}{(\sigma^2+t+1)^w}\left[\left(1+\frac{1}{\sigma^2+t}\right)^{1+w}-1 \right]-1,\\
    \delta_{\sigma^2}(t) &=\frac{1}{\sigma^2+t}\left[\frac{(\sigma^2+t)^{2w+2}}{(\sigma^2+t+1)^{2w}}\frac{1}{2w+1}\left[\left(1+\frac{1}{\sigma^2+t}\right)^{2w+1}-1\right]-(\sigma^2-t)\right].
\end{align}
The system might then enter the conditional phase at a time $t = t_s > 0$, that generally depends on the guidance level $w$. Whether this occurs, the distortion variables when $t \leq t_s$ will read
\begin{align}
    \delta_{\mu}(t) &= \frac{\sigma^2+t}{\sigma^2+t_s}\frac{(\sigma^2+t_s)^{1+w}}{(\sigma^2+t_s+1)^{w}}\left[\left(1+\frac{1}{\sigma^2+t_s}\right)^{1+w}-1\right]+\frac{t_s-t}{\sigma^2+t_s}-1,\\
    \delta_{\sigma^2}(t) &=\frac{\sigma^2+t}{(\sigma^2+t_s)^2}\frac{(\sigma^2+t_s)^{2w+2}}{(\sigma^2+t_s+1)^{2w}}\frac{1}{2w+1}\left[\left(1+\frac{1}{\sigma^2+t_s}\right)^{2w+1}-1\right]\\
    &+\frac{t_s-t}{\sigma^2+t_s}\left(\frac{\sigma^2+t}{\sigma^2+t_s}\right)-1.
\end{align}
Figure \ref{fig:distort2} represents the distortion estimators as evolving in time for a given realization of the control parameters: the two observables keep on decreasing in time. 
Figure \ref{fig:combined_w} reports the entity of the distortion at sampling time $t = 0$, as well as the magnitude of the speciation time $t_s$ for different combinations of the parameters. 
Generally speaking, large speciation times imply small distortion, and viceversa. Interestingly, the speciation time increases with the guidance level $w$, as guidance would reduce the distortion effects when $w$ is very high. 
In fact, as also showed by Figures \ref{fig:distor_exp_app} and \ref{fig:distor_exp_numer}, the trend of $\delta_{\sigma^2}(0)$ presents a minimum at a given value of $w$ and then increases. Also $\delta_{\mu}(0)$, according to our analytical computations, must decrease after reaching a maximum, that can be found at large values of $w$.




\subsubsection{Consistency between variance-exploding (VE) and variance-preserving (VP) diffusion frameworks}

Let us now consider a variance-preserving (VP) framework the forward process, i.e. the following SDE
\begin{equation}
    \label{eq:vp}
    \text{d}\mathbf{x}_t= - \mathbf{x}_t \text{d}t+ \sqrt{2}\text{d}\mathbf{W}_{t}.
\end{equation}
The solution of the forward process reads
\begin{equation}
    \mathbf{x}_t = e^{-t}\mathbf{x}_0 + \sqrt{1-e^{-2t}}\;\boldsymbol{\epsilon}_t,
\end{equation}
where $\boldsymbol{\epsilon}_t \sim \mathcal{N}(0,I_d)$. 
By repeating the REM analysis  one gets the same phase structure as in the variance-exploding (VE) formulation. The guided-to-conditional transition is still determined by
\begin{equation}
    \beta(d) + \zeta^{VP}_{t_s}(1) = 0,
\end{equation}
where $\zeta_{t}(1)$ depends on $\sigma^2$ and $w$ and it is the VP free-energy contribution evaluated along the guided trajectory. At large diffusion times we would obtain
\begin{equation}
    \zeta_t^{VP}(1) = -\left(w + \frac{1}{2}\right)e^{-2t} + \mathcal{O}\left(e^{-4t}\right),
\end{equation}
which leads to
\begin{equation}
    t_s^{VP} \simeq \frac{1}{2}\log \left[\frac{w+\frac{1}{2}}{\beta(d)}\right]. 
\end{equation}
Thus, when the number of classes is sub-exponential in the dimension, i.e. 
$M = \mathcal{o}(d)$, one still has 
$t_s \to \infty$, as in the VE setting. For instance, for a finite number of classes we recover $t_s \simeq \frac{1}{2}\log d$
in agreement with the VP speciation scale identified by \cite{biroli_dynamical_2024, achili_speciation26}. Conversely, when $M = e^{\beta d}$, the speciation time remains finite, i.e. $t_s = \mathcal{O}(1)$, but its exact value must be computed by the implicit guided-to-conditional transition equation \eqref{eq:collapse_CFG2}, as it was for the VE setting. We hence conclude that, despite a difference in the functional scaling in the VP regime, the phenomenology of distortions remains qualitatively the same.

\section{CFG prescriptions: time-dependent guidance level \label{app:neg}}

Let us now evaluate different CFG prescription obtained through changing the guidance level $w$ in time. Generally speaking we will solve our model by considering $w = w(t)$. 

\subsection{The goal of CFG}

From our mean-field analysis of the guided diffusion potential we could conclude that, inside the guided phase, any $w \geq 0$ caused an expansion of the conditional mean, as well as a shrinkage of the variance. 

Previous studies of CFG underline how this technique can be used to reach a favorable sampling condition where: 
\begin{itemize}
    \item Class-related means are more separated from each other, implying a better class separation. 
    \item The variance of sampled examples inside each class is not too small, to avoid loss of sample diversity. 
\end{itemize}
Usually, for a given training dataset, reaching such condition requires a lot of empirical trials, and so far a theoretical criterion is lacking. 

Our theory suggests that, when $w < 0$, which is a peculiar choice in actual CFG, would imply an opposite trend: a shrinkage of the conditional mean and an expansion of the conditional variance (see Eq.~\eqref{eq:min} from the main text).
We hence suggest the following simple guidance prescription 
\begin{equation}
\label{eq:neg_cfg2}
    w(t) = w_0 + \omega\cdot t,\quad\quad w_0\geq -1,\quad \omega > 0,
\end{equation}
that allows to have a controllable time window where the guidance level can be negative, boosting sample diversity.
Let us analyze the effect of such prescription under the lens of tour theory, both in the case of \textit{continuous classes} (see Section \ref{sec:cont} and Appendix \ref{app:multi}) and \textit{separated classes} (see Section \ref{sec:mix} and Appendix \ref{app:rem}) . 
\subsection{Multivariate Gaussian with Continuous Classes \label{app:multi_linearCFG}}
Let us a consider, as data distribution, the $d$-dimensional multivariate Gaussian introduced and analyzed in Appendix \ref{app:multi}. The matrix kernel that solves the backward SDE for this new guidance prescription reads
\begin{align}
M(t_1,t_2) = e^{\int_{t_1}^{t_2} A(t')dt'},    
\end{align}
with
\begin{align}
    \int_{t_1}^{t_2}A(t)dt &= \int_{t_1}^{t_2}\left[-(1+w_0+\omega t)(\Sigma_{x|c}^{t})^{-1}+(w_0+\omega t)(\Sigma_{xx}^{t})^{-1}\right]dt\\ &=\int_{t_1}^{t_2}\left[-\frac{1+w_0+\omega t}{\Sigma_{x|c}+tI_{d_2}}+\frac{w_0+\omega t}{\Sigma_{xx}+t I_{d_2}} \right] dt 
    \\ 
    &= \left(\omega \Sigma_{x|c}-(1+w_0)I_{d_2}\right)\log\left[\frac{\Sigma_{x|c}+t_2 I_{d_2}}{\Sigma_{x|c}+t_1I_{d_2}}\right]\\
    &-\left(\omega \Sigma_{xx}-w_0I_{d_2}\right)\log \left[\frac{\Sigma_{xx}+t_2 I_{d_2}}{\Sigma_{xx}+t_1 I_{d_2}}\right],
\end{align}
Since we assumed that $\Sigma_{xx}$ and $\Sigma_{x|c}$ commute, we can rewrite the matrix kernel as
\begin{align}
  M(t_1,t_2) &= Z(t_1)Z(t_2)^{-1}\\
  Z(t) &= \left(\Sigma_{x|c}+tI_{d_2}\right)^{(1+w_0)I_{d_2}-\omega\Sigma_{x|c}} \;\left(\Sigma_{xx}+tI_{d_2}\right)^{-\left(w_0 I_{d_2}-\omega \Sigma_{xx}\right)},
\end{align}
and we are allowed to substitute matrices with the relative eigenvalues to compute integrals.
More in the specific, one computes
\begin{equation}
\boldsymbol{\mu}_w(t) =\int_t^{+\infty}M(t,t')B(t')\boldsymbol{\mu}\,dt'
= Z(t)\int_t^{+\infty}Z^{-1}(t')B(t')\boldsymbol{\mu}\,dt',
\end{equation}
\begin{equation}
    B(t) = (1+w_0+\omega t)\left(\Sigma_{x|c}+t I_{d_2}\right)^{-1}
\end{equation}
\begin{equation}
    \Sigma_w(t) = \int_{t}^{+\infty}M^2(t,t')\, d t' = Z^{2}(t)\int_{t}^{+\infty}Z^{-2}(t')\, d t'.
\end{equation}
Call $R=\Sigma_{xx}$ and $S=\Sigma_{x|c}$ with eigenvalues $(r_i, s_i)$, then
\begin{equation}
\boldsymbol{\mu}_w(t)=
P\;\mathrm{diag}(\lambda_1(t),\dots,\lambda_{d_2}(t))\;P^{-1},
\end{equation}
where $P$ is the unitary matrix collecting the common eigenvectors to $R$ and $S$ and 
\begin{equation}
\lambda_i(t) = \frac{\left(s_i+t\right)^{(1+w_0)-\omega s_i}}{\left(r_i+t\right)^{w_0-\omega r_i}}\left[\int_{t}^{+\infty}\left(\frac{s_i+t^{'}}{r_i+t^{'}}\right)^{-w_0} \frac{(s+t^{'})^{\omega s_i-2}}{(r_i+t^{'})^{\omega r_i}}\left(1+w_0+\omega t^{'}\right)dt^{'}\right],
\end{equation}
that I can re-write in terms of incomplete Beta functions as
\begin{multline}
\lambda_i(t) = \frac{(s_i+t)^{(1+w_0)-\omega s_i}}{(r_i+t)^{w_0-\omega r_i}}(r_i-s_i)^{-\omega(r_i-s_i)-1}\Bigg[(1+w_0-ws_i)\times \\\times\Bigg(B_1(\omega s_i-w_0-1, \omega(r_i-s_i))-B_{\frac{s_i+t}{r_i+t}}(\omega s_i-w_0-1, \omega(r_i-s_i))\Bigg)+(\omega r_i-(1+w_0))\times \\ \times\Bigg(B_1(\omega s_i-w_0, \omega (r_i-s_i))-B_{\frac{s_i+t}{r_i+t}}(\omega s_i-w_0, w(r_i-s_i))\Bigg)\Bigg]. 
\end{multline}
where the incomplete Beta function $B_f(a,b)$ is defined as
\begin{equation}
    B_{f}(a,b) = \int_0^f dr\; r^{a-1}(1-r)^{b-1},
    \label{eq:inc_beta}
\end{equation}
Beta functions are easy to compute numerically. 
At the same way, the covariance matrix reads
\begin{equation}
\Sigma_w(t)=
P\;\mathrm{diag}(e_1(t),\dots,e_{d_2}(t))\;P^{-1},
\end{equation}
where
\begin{equation}
    e_i(t) = (s_i+t)\cdot \Lambda_i(t), 
\end{equation}
with
\begin{equation}
    \Lambda_i(t) = (s_i+t)^{1+2(w_0-\omega s_i)}(r_i+t)^{-2w_0+2\omega r_i}\int_{t}^{+\infty} \left(\frac{s_i+t^{'}}{r_i + t^{'}}\right)^{-2w_0}\left[\frac{(s_i+t^{'})^{s_i}}{(r_i+t^{'})^{r_i}}\right]^{2\omega}\frac{dt^{'}}{(s_i+t^{'})^2},
\end{equation}
which can be expressed in terms of incomplete Beta functions as
\begin{multline}
    \Lambda_i(t) = (s_i+t)^{1+2(w_0-\omega s_i)}(r_i+t)^{-2w_0+2\omega r_i}(r_i-s_i)^{-2\omega (r_i-s_i)-1}\times \\ \times\left[B_1(2(\omega s_i-w_0)-1, 2\omega(r_i-s_i)+1)-B_{\frac{s_i+t}{r_i+t}}(2(\omega s_i-w_0)-1, 2\omega (r_i-s_i)+1)\right].
\end{multline}

\begin{figure}[ht!]
\centering
\begin{subfigure}{0.32\textwidth}
  \centering
  \includegraphics[width=\linewidth]{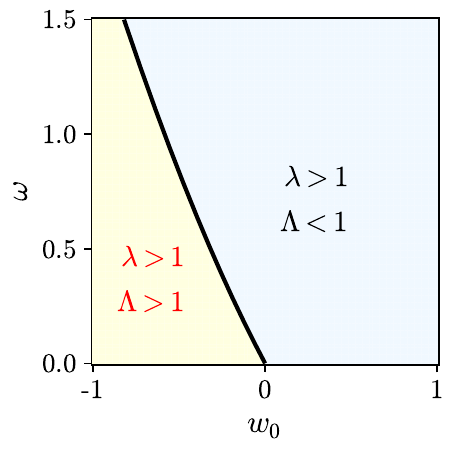}
  \caption{Phase Diagram}
  \label{fig:negCFG_distor_multi_a}
\end{subfigure}
\begin{subfigure}{0.65\textwidth}
  \centering
  \includegraphics[width=\linewidth]{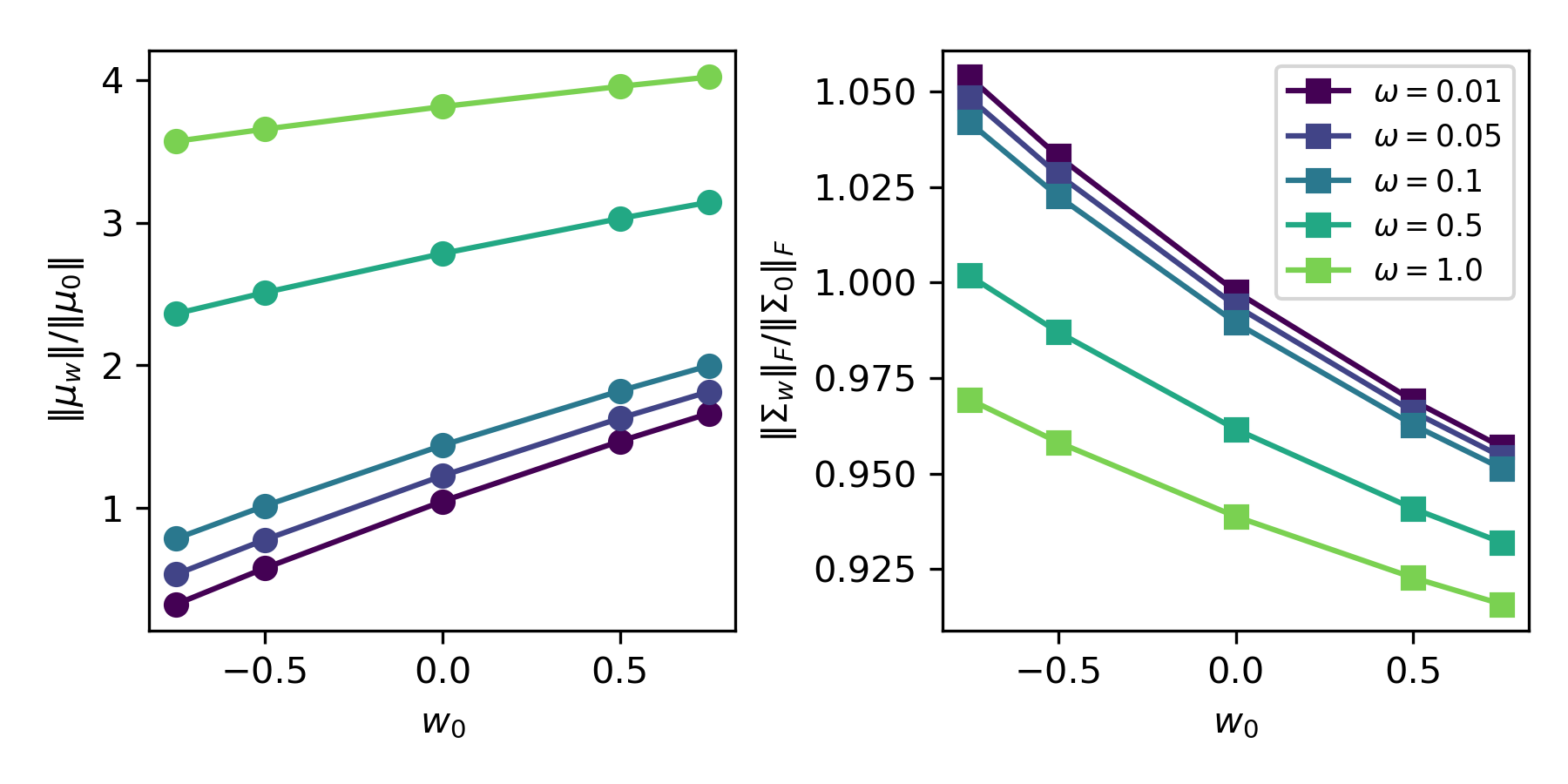}
  \caption{Distortion Measures}
  \label{fig:negCFG_distor_multi_b}
\end{subfigure}
\caption{The effect of linear early-high CFG schedules  where the guidance level is allowed to become negative on a multivariate Gaussian target, depending on two parameters $w_0$ and $\omega$. Panel (a) represents a phase diagram reporting the beneficial region in yellow: here both distortion weights are larget than unity. The outer region contains schedules where the mean is expanded and covariances are contracted. Panel (b) reports distortion estimators at $t = 0$ measured from numerical simulations at different values of $w_0$ and $\omega$, for $d_1 = 1$ and $d_2 = 9$. Specifically: the norm of the mean divided by the norm of the conditional mean on the left; the norm of the covariance matrix divided by the norm of the conditional covariance matrix on the right. As we can notice, for small values of $w_0$ and small values of $\omega$ (i.e. a large negative guidance time window) we reach the yellow region of the phase diagram. The general effect is mild with this target distribution, because positive distortion is enhanced only along directions where the difference $r_i - s_i$ is large.}
\label{fig:negCFG_distor_multi_rs}
\end{figure}

\begin{figure}[ht!]
\centering
\begin{subfigure}{0.4\textwidth}
  \centering
  \includegraphics[width=\linewidth]{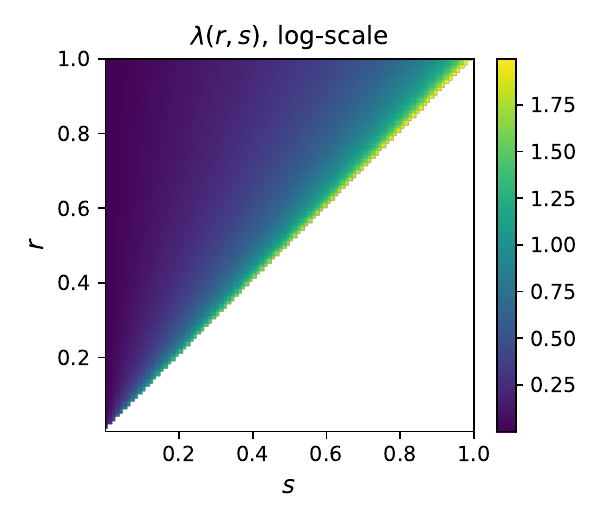}
  \caption{}
\end{subfigure}
\begin{subfigure}{0.4\textwidth}
  \centering
  \includegraphics[width=\linewidth]{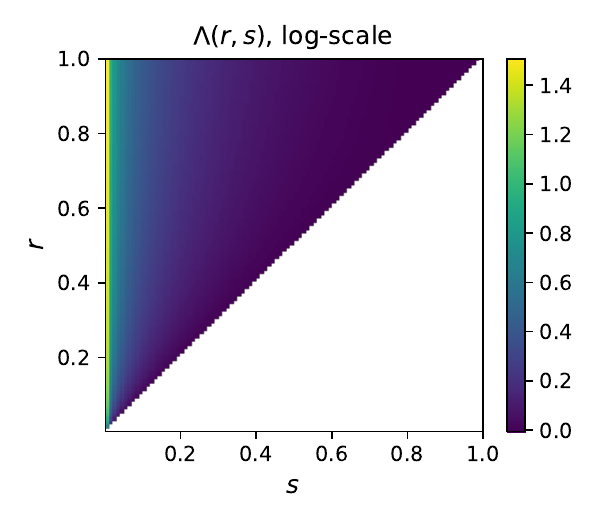}
  \caption{}
\end{subfigure}
\caption{Distortion weights $\lambda$ and $\Lambda$ as functions of the eigenvalues $(r,s) \in [0,1]^2$, with $s \leq r$, for a linear ealry-high CFG schedule with $w_0 = -0.75$ and $\omega = 1$. The values are plotted in log-scale for enhancement. Panel (a) shows that the mean is expanded significantly only when $s_i \simeq r_i$. Panel (b) displays a significant distortion of covariances in the opposite case, i.e. $r_i - s_i$ large. This behavior implies that the effects of such CFG prescription must be mild in general.}
\label{fig:negCFG_distor_multi_rs2}
\end{figure}

Let us consider one direction $i$ of the ambient space and one choice of the eigenvalues of $\Sigma_{xx}$ and $\Sigma_{x|c}$, respectively $r_i = 1$ and $s_i = 0.6$. 
At this point we can trace a phase-diagram of distortion as a function of $w_0$ and $\omega$, that is represented in Figure \ref{fig:negCFG_distor_multi_a}. We identify one region where we have $\lambda > 1$ and $\Lambda > 1$ simultaneously, and it occurs when $w_0 < 0$ and $\omega$  is smaller than a characteristic value which depends on the data distribution. Outside this region, the width of the negative-guidance window is not large enough to improve sample diversity and allow $\Lambda > 1$. We also simulated the system for one choice of the class and covariance matrices at different values of $w_0$ and $\omega$. 
As we can notice from Figure \ref{fig:negCFG_distor_multi_b} there is a little improvement of the performance, that becomes more evident when the time window is larger, i.e. when $w_0$ is largely negative and $\omega$ is small. 
As showed in Figure \ref{fig:negCFG_distor_multi_rs2}, we find that for this class of data distributions the effects of the schedule must be mild, since we need a large difference between the eigenvalues $r_i - s_i$ to achieve significant expansions of the covariances, and this might not occur along all the directions of the data space. 

\subsection{Mixture of an Exponential number of Gaussians \label{app:mog_linear_CFG}}
Let us consider a target distribution being a mixture of $M$ Gaussians where $M = e^{\beta d}$ and centroids are $d$-dimensional normal vectors. Then the REM theory developed in Appendix \ref{app:rem} applies and the backward SDE is
\begin{equation}
    dx = x\frac{\sigma^2 + t + 1 +w_0 + \omega t}{(\sigma^2+t)(\sigma^2+t+1)}dt- c^1 \frac{1+w_0+\omega t}{(\sigma^2+t)}dt + dW_t 
\end{equation}
The auxiliary functions read
\begin{equation}
    a(t) = \frac{\sigma^2 + t + 1 +w_0 + \omega t}{(\sigma^2+t)(\sigma^2+t+1)},\hspace{0.5cm}
    b(t) = - c^1 \frac{1+w_0+ \omega t}{\sigma^2+t},
\end{equation}
\begin{equation}
    \Phi(t) = \left(\frac{\sigma^2}{\sigma^2+t}\right)^{1+w_0-\omega\sigma^2}\left(\frac{1+\sigma^2}{\sigma^2+t+1}\right)^{\omega(1+\sigma^2)-w_0}.
\end{equation}
The evolution equation reads
\begin{align}
    x_t &= \left(\frac{\sigma^2+t}{\sigma^2+T}\right)^{1+w_0-\omega\sigma^2}\left(\frac{\sigma^2+t+1}{\sigma^2+T+1}\right)^{\omega(1+\sigma^2)-w_0}x_T-c^1\frac{\left(\sigma^2+t\right)^{1+w_0-\omega\sigma^2}}{\left(\sigma^2+t+1\right)^{w_0-\omega(1+\sigma^2)}}\\
    &\times\Bigg[(1+w_0-\omega\sigma^2)\left[B_{f(t)}\left(\omega\sigma^2-w_0-1, \omega\right)-B_{f(T)}\left(\omega\sigma^2-w_0-1, \omega\right)\right]\\
    &+\left(\omega(1+\sigma^2)-w_0-1\right)\left[B_{f(t)}\left(\omega\sigma^2-w_0, \omega)\right)-B_{f(T)}\left(w\sigma^2-w_0, \omega\right)\right]\Bigg]\\
    &+\frac{\left(\sigma^2+t\right)^{1+w_0-\omega\sigma^2}}{\left(\sigma^2+t+1\right)^{w_0-\omega(1+\sigma^2)}}\int_{T}^t ds\;\xi(s)\frac{(\sigma^2+s+1)^{w_0-\omega(1+\sigma^2)}}{(\sigma^2+s)^{1+w_0-\omega\sigma^2}},
\end{align}
where the incomplete Beta function $B_f(a,b)$ are defined in Eq. \eqref{eq:inc_beta} 
and the function $f(x)$ appearing in our version of the Beta function reads
\begin{equation}
    f(x) = \frac{\sigma^2+x}{\sigma^2+x+1}.
    \label{eq:f}
\end{equation}
Incomplete Beta functions are easy to compute compute numerically. 
Instead, $x_t$ is the initial condition for the integration. From the isotropy of the process, the solution $x_t$ must be normally distributed with mean
\begin{align}
    \mu(t) &= \left(\frac{\sigma^2+t}{\sigma^2+T}\right)^{1+w_0-\omega\sigma^2}\left(\frac{\sigma^2+t+1}{\sigma^2+T+1}\right)^{\omega(1+\sigma^2)-w_0}\mu(T)-c^1\frac{\left(\sigma^2+t\right)^{1+w_0-\omega\sigma^2}}{\left(\sigma^2+t+1\right)^{w_0-\omega(1+\sigma^2)}}\\
    &\times\Bigg[(1+w_0-\omega\sigma^2)\left[B_{f(t)}\left(\omega\sigma^2-w_0-1, \omega\right)-B_{f(T)}\left(\omega\sigma^2-w_0-1, \omega\right)\right]\\
    &+\left(\omega(1+\sigma^2)-w_0-1\right)\left[B_{f(t)}\left(\omega\sigma^2-w_0, \omega)\right)-B_{f(T)}\left(\omega\sigma^2-w_0, \omega\right)\right]\Bigg]
\end{align}
and variance 
\begin{align}
    \sigma^2(t) &= \left(\frac{\sigma^2+t}{\sigma^2+T}\right)^{2(1+w_0-\omega\sigma^2)}\left(\frac{\sigma^2+t+1}{\sigma^2+T+1}\right)^{2\omega(1+\sigma^2)-2w_0}\sigma^2(T) \\
    &+\frac{(\sigma^2+t)^{2(1+w_0-\omega\sigma^2)}}{(\sigma^2+t+1)^{2w_0-2\omega(1+\sigma^2)}}\\
    &\times \left[B_{f(t)}\left(2\omega\sigma^2-2w_0-1,1+2\omega\right)-B_{f(T)}\left(2\omega\sigma^2-2w_0-1,1+2\omega\right)\right]. 
\end{align}
An initial condition $\sigma^2(T) = \mathcal{O}(T)$ should require $w > -1/2$, that is always satisfied in this case. 
At this point one can perform the $T \to \infty$ obtaining 
\begin{align}
    \mu(t) &= -c^1\frac{\left(\sigma^2+t\right)^{1+w_0-\omega\sigma^2}}{\left(\sigma^2+t+1\right)^{w_0-\omega(1+\sigma^2)}}\\
    &\times\Bigg[(1+w_0-\omega\sigma^2)\left[B_{f(t)}\left(\omega\sigma^2-w_0-1, \omega\right)-B_{1}\left(\omega\sigma^2-w_0-1, \omega\right)\right]\\
    &+\left(\omega(1+\sigma^2)-w_0-1\right)\left[B_{f(t)}\left(\omega\sigma^2-w_0, \omega)\right)-B_{1}\left(\omega\sigma^2-w_0, \omega\right)\right]\Bigg],
\end{align}
\begin{align}
    \sigma^2(t) &= \frac{(\sigma^2+t)^{2(1+w_0-\omega\sigma^2)}}{(\sigma^2+t+1)^{2w_0-2\omega(1+\sigma^2)}}\\
    & \times\left[B_{f(t)}\left(2\omega\sigma^2-2w_0-1,1+2\omega\right)-B_{1}\left(2\omega\sigma^2-2w_0-1,1+2\omega\right)\right]. 
\end{align}
At this point the distortion estimators become
\begin{align}
    \delta_{\mu}(t) &=-\frac{\left(\sigma^2+t\right)^{1+w_0-\omega\sigma^2}}{\left(\sigma^2+t+1\right)^{w_0-\omega(1+\sigma^2)}}\\
    &\times\Bigg[(1+w_0-\omega\sigma^2)\left[B_{f(t)}\left(w\sigma^2-w_0-1, \omega\right)-B_{1}\left(\omega\sigma^2-w_0-1, \omega\right)\right]\\
    &+\left(\omega(1+\sigma^2)-w_0-1\right)\left[B_{f(t)}\left(\omega\sigma^2-w_0, \omega)\right)-B_{1}\left(\omega\sigma^2-w_0, \omega\right)\right]\Bigg]-1,
\end{align}
and 
\begin{align}
    \delta_{\sigma^2}(t) &= \frac{(\sigma^2+t)^{1+2(w_0-\omega\sigma^2)}}{(\sigma^2+t+1)^{2w_0-2\omega(1+\sigma^2)}}\\
    &\times\left[B_{f(t)}\left(2\omega\sigma^2-2w_0-1,1+2\omega\right)-B_{1}\left(2\omega\sigma^2-2w_0-1,1+2\omega\right)\right]-1, 
\end{align}

\begin{figure}[ht!]
\centering
\begin{subfigure}{0.3\textwidth}
  \centering
  \includegraphics[width=\linewidth]{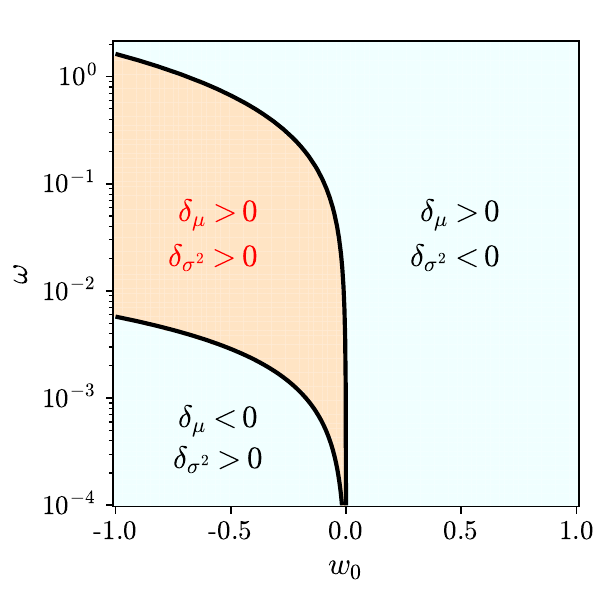}
  \caption{Phase Diagram}
\end{subfigure}\hfill
\begin{subfigure}{0.32\textwidth}
  \centering
  \includegraphics[width=\linewidth]{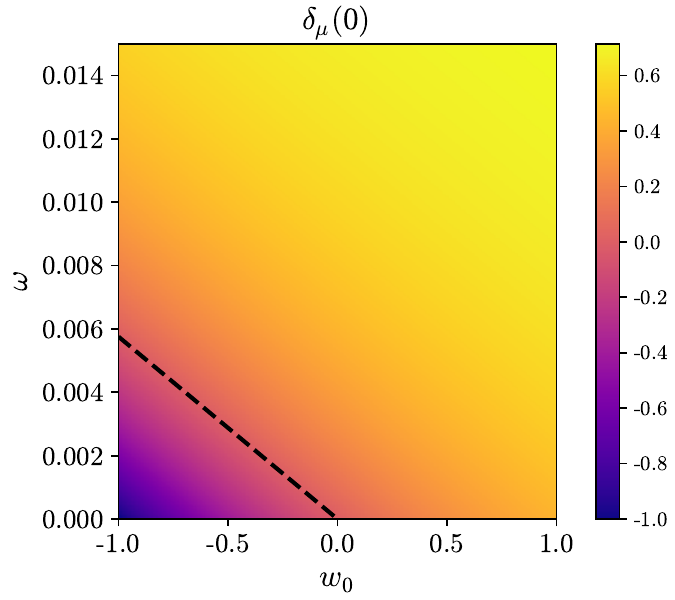}
\caption{Distortion of the Mean}
\end{subfigure}\hfill
\begin{subfigure}{0.32\textwidth}
  \centering
  \includegraphics[width=\linewidth]{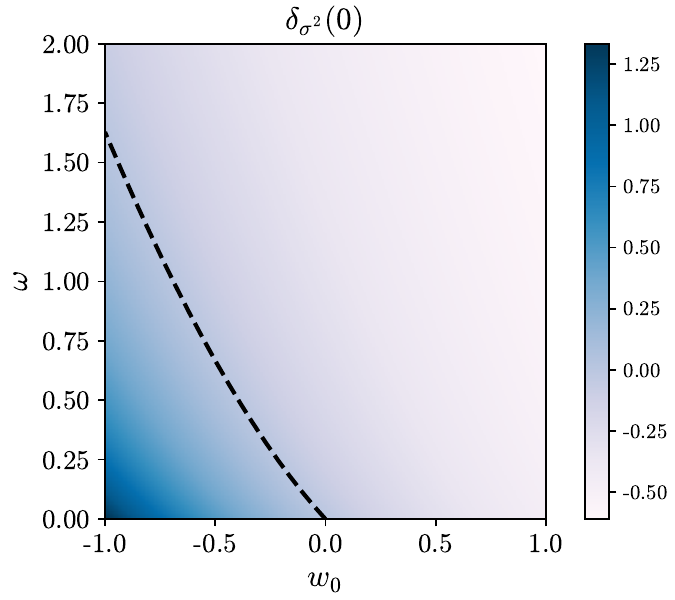}
  \caption{Distortion of the Variance}
\end{subfigure}

\caption{The effect of linear late-high CFG schedule where the guidance level is allowed to become negative, depending on two parameters $w_0$ and $\omega$. The variance of the mode is $\sigma^2 = 0.75$ and the mode density $\beta$ has been chosen to be large, so that the system never enters the conditional phase in the displayed domain. The first panel from the left is a phase diagram reporting the beneficial region in orange: here both means and variances are enlarged. The other regions either show loss of separability or tendency to shrink the mode variance. The following panels report distortion estimators. Dashed lines signal the passage from a negative distortion at time $t = 0$, to a positive one. 
}
\label{fig:negCFG_distor_exp_app}
\end{figure}

\subsection*{Sanity check: a guidance prescription that is solvable in closed form}

Let us consider a late-high linear prescription where  $w_0 = \sigma^2-1$ and $\omega = 1$. Then one has
\begin{equation}
    w(t) = \sigma^2+t-1.
\end{equation}
The guidance level can become negative at $t^* = 1-\sigma^2$, whether $\sigma^2 < 1$. 
This choice for $w(t)$ simplifies the SDE per component of the guided phase of the potential into 
\begin{equation}
    dx_ = \frac{2x}{\sigma^2+t+1}dt-c^1dt+ dW_t.
\end{equation}
The auxiliary functions read
\begin{equation}
    a(t) = \frac{w}{\sigma^2+t},\hspace{1cm}
    b(t) = -c^1,\hspace{1cm}
    \Phi(t) = \left(\frac{\sigma^2+1}{\sigma^2+t+1}\right)^2.
\end{equation}
The evolution equation reads
\begin{align}
    x_t &= \left(\frac{\sigma^2+t+1}{\sigma^2 + T + 1}\right)^2x_T+\left[(\sigma^2+t+1)-\frac{(\sigma^2+t+1)^2}{\sigma^2+T+1}\right]c^1\\
    &+(\sigma^2+t+1)^2\int_{T}^t ds\; \frac{\xi(s)}{(\sigma^2 + s+1)^2},
\end{align}
where $x_t$ is the initial condition for the integration. From the isotropy of the process, the solution $x_t$ must be normally distributed with mean
\begin{equation}
    \mu(t) = \left(\frac{\sigma^2+t+1}{\sigma^2 + T + 1}\right)^2\mu(T)+\left[(\sigma^2+t+1)-\frac{(\sigma^2+t+1)^2}{\sigma^2+T+1}\right]c^1,
\end{equation}
and variance 
\begin{equation}
    \sigma^2(t) = \left(\frac{\sigma^2+t+1}{\sigma^2 + T + 1}\right)^4 \sigma^2(T)+\frac{(\sigma^2+t+1)^4}{3}\left[\frac{1}{(\sigma^2+t+1)^3}-\frac{1}{(\sigma^2+T+1)^3}\right]. 
\end{equation}
When $T \gg 1$ the same moments become
\begin{equation}
    \boldsymbol{\mu}(t) = (\sigma^2+t+1)\;\mathbf{c}^1,
\end{equation}
\begin{equation}
    \sigma^2(t) = \frac{(\sigma^2+t+1)}{3}.
\end{equation}
As a consequence, the deformation observable evolve in time according to
\begin{equation}
    \delta_{\mu}(t) = \sigma^2+t,
\end{equation}
\begin{equation}
    \delta_{\sigma^2}(t) = \frac{1}{3\sigma^2}\left(t+1-2\sigma^2\right).
\end{equation}
If we choose $\sigma^2 \leq 1/2$ we can obtain the desired condition  $\delta_{\mu}(0) = \sigma^2 > 0$ and $\delta_{\sigma^2}(0) = \frac{1-2\sigma^2}{3} \geq 0$.
In this setup, we can still have collapse of the potential to the phase where its local minimum is centered on $\mathbf{c}^1$. In this phase the diffusion potential is centered on the class, so diffusion will be drifted towards a situation where $\delta_{\mu} > 0$ and $\delta_{\sigma^2} < 0$. This transition occurs when the collapse time $t_s$ satisfies the following condition
\begin{equation}
    t_s = \frac{1}{e^{2\beta-1}-1}-\sigma^2 > 0,\quad \quad \beta > 1/2.
\end{equation}
Specifically, when $\beta \to 1/2$ we have $t_s \rightarrow \infty$, implying zero distortion of the target measure, regardless of the value of $\sigma^2$. When $\beta \leq 1/2$ the REM free-energy of the model is strictly negative implying the system to always live in the conditional phase, implying no distortion.


\section{Experiments with Stable Diffusion}\label{app:experiments}

In this Appendix we first describe and motivate our choice of the metrics to evaluate the generative performance of CFG. Then we provide the reader with further details about the experimental procedure employed along our manuscript. 

\subsection{Experimental details}

We generated a synthetic image dataset using Stable Diffusion v1.5 (runwayml/stable-diffusion-v1-5) with the Hugging Face diffusers implementation. Images were generated at resolution 512×512 using the Euler Ancestral Discrete scheduler for 30 denoising steps. The model was used only for inference; no fine-tuning was performed.
All experiments were run on NVIDIA TITAN RTX with 24 GB memory using CUDA. 

All the details about CFG schedules are detailed in Section~\ref{app:add_exp}.

\subsection{Evaluation Metrics}
\label{app:metrics}

Our analysis separates the effect of CFG onto two components: \textit{class separation} and \textit{sample diversity}. 

The first property is showed to be related to a good prompt-alignment, i.e. the capability of the model to generate samples that are coherent with the conditioning class. Empirically talking, class separation is boosted by CFG techniques through an expansion of the guided mean with respect to the conditional one.

The second property is instead relative to the variance of the samples generated from a given prompt, or also to the entropy of the guided distribution. A very small variance / entropy indicates the collapse of the generated samples on a single configuration. 

We will use estimators that are capable of disentangling the deformation of the mean from the deformation of the variance. As a consequence we avoid using metrics that, instead, aggregate such effects, such as FID, Inception Score or Saturation (in pixel or feature space). 
We will mainly apply such metrics to representations of the data in the latent feature space, where features are extracted via CLIP \cite{clip} or DINOv2 \cite{dinov2} models. 

\subsubsection{Class Separation}

In order to evaluate class separation in the samples generated via CFG we will adopt the following estimators: $L_2$ distance, Cosine Similarity between the means, and CLIP score. These quantities only depend on the magnitudes of the means of the distorted and the conditional distributions, without involving higher order cumulants. 
\paragraph{$L_2$ distance.} 
This quantity is defined as the mean squared distance between the mean of the CFG sampled data at a given guidance level $w$ and the mean measured in the conditional setting, i.e. at $w = 0$: 
\begin{equation}
\label{eq:L2}
    L_2(w) = \mathbb{E}_p\;  \|\bar{f}_{p,w} -  \bar{f}_{p,0}\|^2,\quad\quad\text{with}\quad \bar{f}_{p,w} = \mathbb{E}_{f|w,p}\,\left[ \;f\;\right].
\end{equation}
\paragraph{Cosine Similarity.} 
This observable measures the alignment between the mean feature vectors at guidance level $w$ and in the conditional setting. It is defined as
\begin{equation}
    \text{Cos}(w) = \mathbb{E}_p \left[ \frac{\langle \bar{f}_{p,w}, \bar{f}_{p,0} \rangle}{\|\bar{f}_{p,w}\|\;\|\bar{f}_{p,0}\|} \right].
\end{equation}
Differently from the $L_2$ distance, this metric is insensitive to the norm of the mean vectors and only captures directional changes induced by CFG. A decrease in cosine similarity signals that guidance alters the semantic direction of the generated features with respect to the conditional distribution.

\paragraph{CLIP Score.} 
We evaluate prompt alignment by computing the CLIP similarity between generated images and their corresponding textual prompts. Given a prompt $p$ and generated sample in pixel space $x_{p,w}$, we define
\begin{equation}
    \text{CLIP}(w) = \mathbb{E}_p \left[ \mathbb{E}_{f|w,p} \; \langle f_{p,w}, \phi_{\text{text}}(p) \rangle \right],
\end{equation}
where $\phi_{\text{text}}$ denotes the normalized CLIP embedding of the text prompt. This metric directly quantifies semantic consistency between images and prompts, i.e. prompt-alignment of the generated sample.

\subsubsection{Sample Diversity}

\paragraph{Features Variance.} 
This observable is most natural quantity to quantify the diversity of generated samples and it corresponds to
\begin{equation}
    \Sigma^2(w) = \mathbb{E}_p\left[\mathbb{E}_{f|w,p}\;\|f_{p,w} - \overline{f}_{p,w} \|^2\right],
\end{equation}
where $\overline{f}$ is the mean defined in Eq. \eqref{eq:L2}. 
In the particular setting where the CFG distorted distribution is Gaussian, we know that this quantity controls the entropy of such distribution. 

\paragraph{Participation Ratio.} 
This quantity is defined as  
\begin{equation}
 \text{PR}_{w} = \mathbb{E}_{p}\left[\frac{\left(\sum_i \lambda_{i}^{p,w}\right)^2}{\sum_i (\lambda_{i}^{p,w})^2}\right],   
\end{equation}
where $\lambda_{i}^{p,w}$ are PCA eigenvalues of the features for each prompt and guidance level. Assume that the empirical covariance matrix of the data features belong to a certain random matrix class. Notice that $PR_w$ is strongly related to the entropy of the probability distribution of the eigenvalues $\lambda_i$. 
In fact we can rewrite $\text{PR}_{w} = \mathbb{E}_p \frac{1}{\sum_i \omega_i^2}$ with $\omega_i = \lambda^{p,w}_i / \sum_i \lambda^{p,w}_i$ is a probability distribution over eigenvalue indices. We also define the Renyi entropy of order $2$ of a probability distribution as 
$$S = -\log\left(\sum_i \omega_i^2\right),$$
with $\omega_i^2$ being the probability that two samples assume the same value. 
From our definition of $\text{PR}_w$, if we assume that $d$ is large enough for the fluctuations of $\sum_i \omega_i^2 / d$ to be small, we can conclude that $\text{PR}_w \approx e^{S_w}$ where $S_w$ is the Renyi entropy of the distribution of the eigenvalues of the covariance of the data. 
When $S_w$ decreases also $\text{PR}_w$ decreases, and the spectrum becomes more concentrated, meaning that variance is dominated by a small number of principal components. A vanishing $\text{PR}_w$ could signal that the data distribution has become spiked, killing diversity. 

The covariance matrix is computed with the Ledoit–Wolf shrinkage estimator, which is more stable and reliable than the usual sample covariance when considering high-dimensional data or limited samples.

\paragraph{Trace Ratio.}
The ratio of the trace of the covariance matrix with a certain guidance level to to purely conditional one 
\begin{equation}
    TR_w = \frac{\mathrm{Tr(\Sigma(w))}}{\mathrm{Tr(\Sigma(0))}}
\end{equation}
is a scalar summary of total spread: a value below 1 means that CFG has compressed the distribution.

\paragraph{Pairwise Distance.} 
This observable is defined as
\begin{equation}
    D(w) = \mathbb{E}_p \frac{1}{2N_p}\sum_{i,j} \| f_{p,w}^{(i)} - f_{p,w}^{(j)} \|^2, 
\end{equation}
where $N_p$ is the number of samples generated per prompt. Intuitively, whenever the samples distribution collapses on one spike at one $w$, the quantity $D(w)$ must vanish.   

\paragraph{Vendi Score.}
The Vendi score \cite{friedman2023vendi} measures the effective number of distinct samples generated from a given prompt. Given the $N_p$ samples generated from prompt $p$ at guidance level $w$, we build the similarity matrix $K^{p,w}_{ij} = k(f_{p,w}^{(i)}, f_{p,w}^{(j)})$, with $k$ the cosine similarity between feature vectors, and define
\begin{equation}
    \text{VS}(w) = \mathbb{E}_p \left[ \exp\left( - \sum_i \mu_i^{p,w} \log \mu_i^{p,w} \right) \right],
\end{equation}
where $\mu_i^{p,w}$ are the eigenvalues of $K^{p,w}/N_p$. Since $\sum_i \mu_i^{p,w} = 1$, the Vendi score is the exponential of the Shannon entropy (i.e. the Renyi entropy of order $1$) of the eigenvalue distribution, and it is therefore closely related to the Participation Ratio, which corresponds to the order $2$. The Vendi score ranges from $1$, when all samples are identical, to $N_p$, when all samples are orthogonal in feature space. In the following we report the ratio $\text{VS}(w)/\text{VS}(0)$, where values below $1$ indicate that CFG has reduced the effective number of distinct samples with respect to the conditional distribution.

\paragraph{LPIPS.}
This estimator represents a perceptual metric computed directly in image space. LPIPS evaluates the average perceptual distance between pairs of generated samples by comparing deep feature activations across multiple layers of a pretrained network. For samples generated from the same prompt, we define
\begin{equation}
    \text{LPIPS}(w) = \mathbb{E}_p \left[ \frac{1}{2N_p} \sum_{i,j} d_{\text{LPIPS}}(x_{p,w}^{(i)}, x_{p,w}^{(j)}) \right],
\end{equation}
where $d_{\text{LPIPS}}$ denotes the learned perceptual similarity metric. This quantity captures variations in texture and structure that may not be fully reflected in latent feature representations. A decrease in LPIPS indicates perceptual collapse toward similar image configurations.

\subsubsection{Human Preference}

\paragraph{Human Preference Score.}
Besides class separation and diversity, we evaluate the perceived quality of the generated images through the Human Preference Score (HPS) \cite{wu2023hpsv2}, 
a model fine-tuned on large-scale human preference data to predict which image humans would favour for a given prompt. Given a prompt $p$ and generated sample $x_{p,w}$, we define
\begin{equation}
    \text{HPS}(w) = \mathbb{E}_p \left[ \mathbb{E}_{x|w,p} \; s_{\text{HPS}}(x_{p,w}, p) \right],
\end{equation}
where $s_{\text{HPS}}$ denotes the score returned by the preference model. Since this score accounts jointly for prompt-alignment and aesthetic quality, we use it as a complementary check that gains in diversity are not obtained at the cost of degraded images. We report its variation with respect to the conditional setting, $\Delta\text{HPS}(w) = \text{HPS}(w) - \text{HPS}(0)$.
\subsection{Additional Experimental Results}
\label{app:add_exp}

\subsubsection{Measure of Generative Distortions}

\begin{figure}
    \centering
    \includegraphics[width=\linewidth]{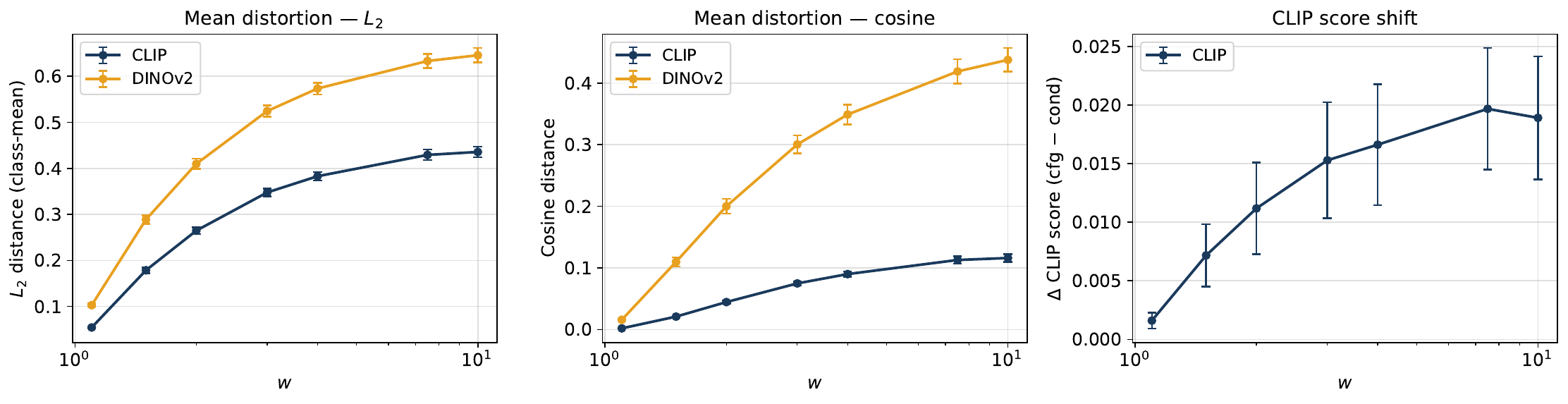}
    \caption{Evaluation of class separation metrics in feature space as a function of the guidance level $w$ averaged over a dataset from Stable Diffusion (v1.5). Points are averaged over 50 prompts and 20 samples per prompt, errors are standard deviations of the mean. Blue circles refer to CLIP feature extractor, yellow ones to DINOv2.}
    \label{fig:mean_metrics}
\end{figure}

\begin{figure}
    \centering
    \includegraphics[width=\linewidth]{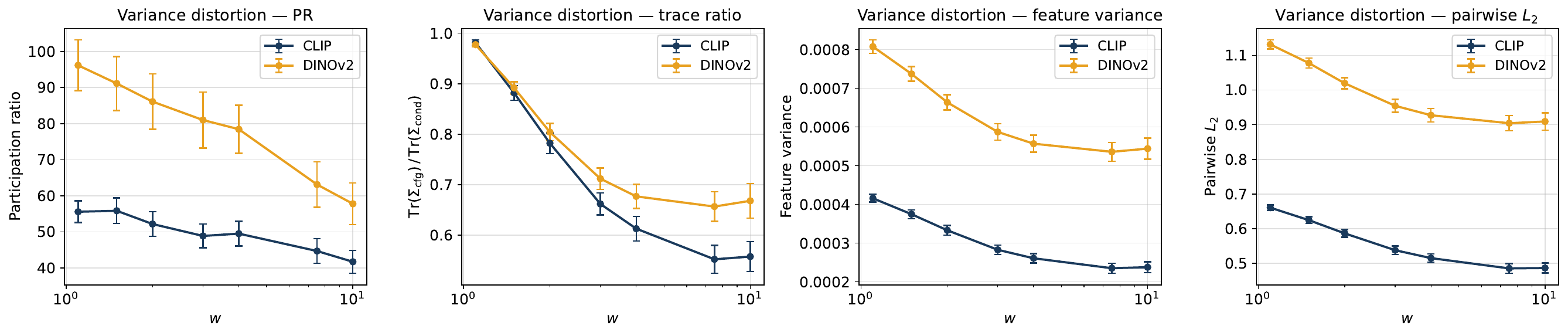}
    \caption{Evaluation of data dispersion metrics in feature space as a function of the guidance level $w$ averaged over a dataset from Stable Diffusion (v1.5). Points are averaged over 50 prompts and 20 samples per prompt, errors are standard deviations of the mean. Blue circles refer to CLIP feature extractor, yellow ones to DINOv2.}
    \label{fig:feature-variance}
\end{figure}

\begin{figure}
    \centering
    \includegraphics[width=0.4\linewidth]{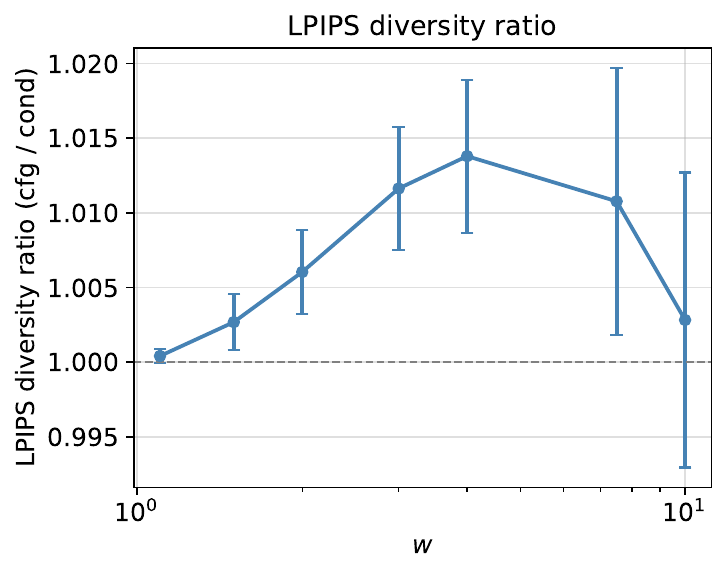}
    \caption{LPIPS as a function of the guidance level $w$ averaged over a dataset from Stable Diffusion (v1.5). Points are averaged over 50 prompts and 20 samples per prompt, errors are standard deviations of the mean.}
    \label{fig:pixel-variance}
\end{figure}

In this Section we provide additional results about the experiments described in Sec.~\ref{sec:sign}. We generate a dataset of images from Stable Diffusion (v1.5) \cite{rombach2022high}: for $50$ different prompts we generate $20$ images at different guidance levels. For each prompt and guidance level samples are then mean-centered, and we measure the metrics described in Sec.~\ref{app:metrics}, using using both CLIP and DINOv2 embeddings for the feature space. The metrics relative to class separation are shown in Figure~\ref{fig:mean_metrics}, the ones that quantify dispertion are in Figure~\ref{fig:feature-variance}. Finally we report LPIPS , which is a pixel-space metric, in Figure~\ref{fig:pixel-variance}.

Feature-space analysis exhibits clear and monotonic trends. 
Mean distortion metrics, $L_2$ distance and cosine similarity, monotonically increase for both encoders.
The fact that DINO consistently shows larger mean distortion suggests CFG has a stronger effect on low-level structural features than on high-level semantic ones.
CLIP score with respect to conditional also increases.
Both CLIP and DINOv2 participation ratio, trace ratio, feature variance and pairwise $L_2$ decrease steadily with increasing guidance. This indicates a genuine reduction of perceptual and semantic diversity rather than artifact-driven effects.
DINOv2 features show a faster initial reduction in variability than CLIP features, suggesting that guidance first suppresses structural and layout-level variation. CLIP features collapse more strongly at higher guidance values, indicating a subsequent reduction in semantic diversity. This ordering is further supported by PCA geometry: the participation ratio decreases monotonically in both spaces, but remains consistently higher for DINOv2 than for CLIP, while the explained variance of the first principal component grows more strongly in CLIP space.

Pixel-level metric LPIPS show a non-monotonic dependence on guidance. Perceptual similarity increases from low to intermediate guidance values, indicating suppression of stochastic noise and fine-scale variation, but decreases again at high guidance. This rebound is consistent with the emergence of sharpening and saturation artifacts rather than meaningful diversity.


\begin{figure}
    \centering
    \includegraphics[width=0.45\linewidth]{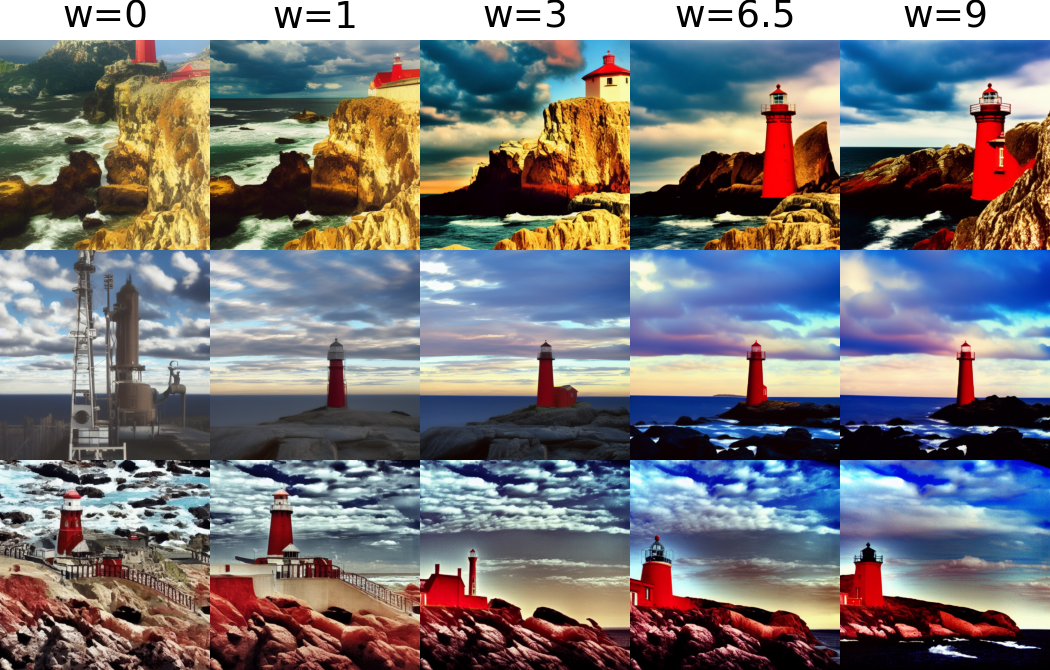}
    \includegraphics[width=0.45\linewidth]{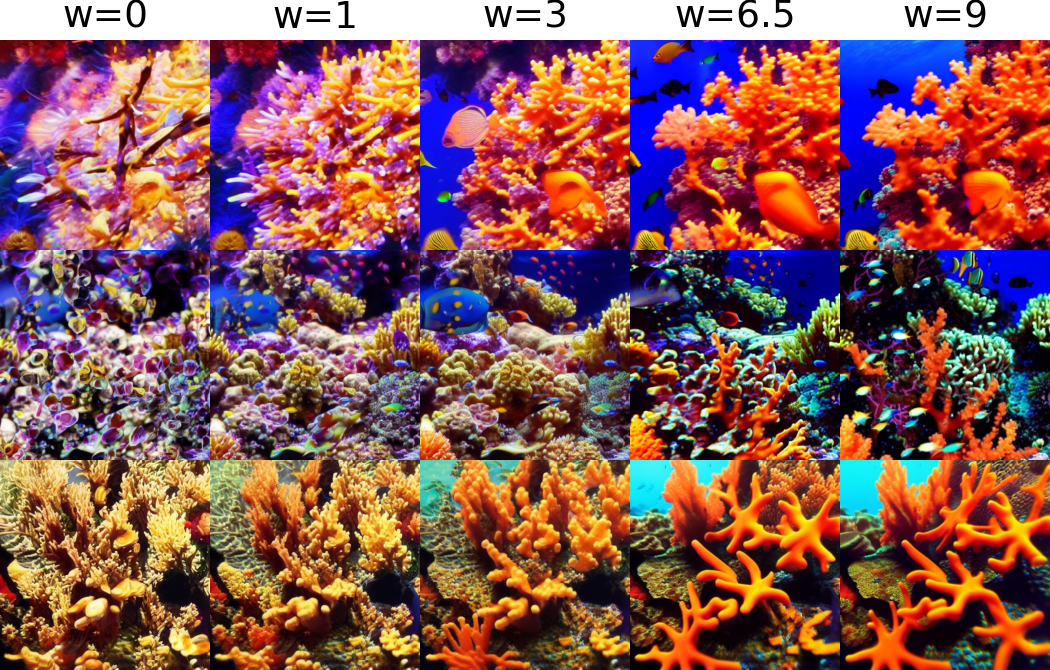}
    \caption{Samples of images generated with Stable Diffusion v1.5 at different guidance levels. Rows are different random seeds, columns refer to guidance levels. Prompts read, respectively: \textit{a photo of a red lighthouse on a rocky coast, dramatic sky, 35mm film}, and \textit{a vibrant coral reef teeming with colorful fish, underwater scene, bright lighting}.}
    \label{fig:sd-samples}
\end{figure}

Figure \ref{fig:sd-samples} reports further generation of images from prompts. 

\subsubsection{Testing the Negative-Window Guidance Schedule}

We now provide for experimental details behind the testing of the Negative-Window guidance schedule. 
We considered the following time schedules for the guidance levels:

\begin{itemize}
    \item \textbf{vanilla}: the guidance level is fixed along diffusion at ($w=1.0, 2.0, 3.0, 4.0, 5.0, 6.5$) 
    \item \textbf{$w=0$ (percentage)}: early-high schedules. Starting from a reference guidance level ($w=1.0, 2.0, 3.0, 4.0, 5.0, 6.5$), after a percentage (10, 30, 50, 70 \%) of inference steps of the backward process we set the guidance level to $w=0.0$.
    \item \textbf{early low 50 \%}: early-low schedule. Start from $w=0.0$ and after 50\% of inference steps we increase a reference guidance level to the nominal values ($w=1.0, 2.0, 3.0, 4.0, 5.0, 6.5$).
    \item \textbf{interval (percentage)}: guidance applied only in a limited interval of the backward process \cite{kynkaanniemi2024interval}. The reference guidance level ($w=1.0, 2.0, 3.0, 4.0, 5.0, 6.5$) is applied for a window placed in the middle  covering a percentage (30, 40, 50, 60, 70 \%) of the inference steps,
    while the guidance is set to $w=0.0$ outside of it.
    \item \textbf{interval 50\% off}:  complementary schedule where guidance is switched off in the central 50\% of the steps and applied elsewhere.
    \item Our schedule,  \textbf{negative (percentage)}: early-high schedule with negative window. Starting from a reference guidance level ($w=1.0, 2.0, 3.0, 4.0, 5.0, 6.5$), after a percentage (10, 30, 50, 70 \%) of inference steps of the backward process we set the guidance level to $w=-1.0$.
\end{itemize}

\begin{figure}
    \centering
    \includegraphics[width=\linewidth]{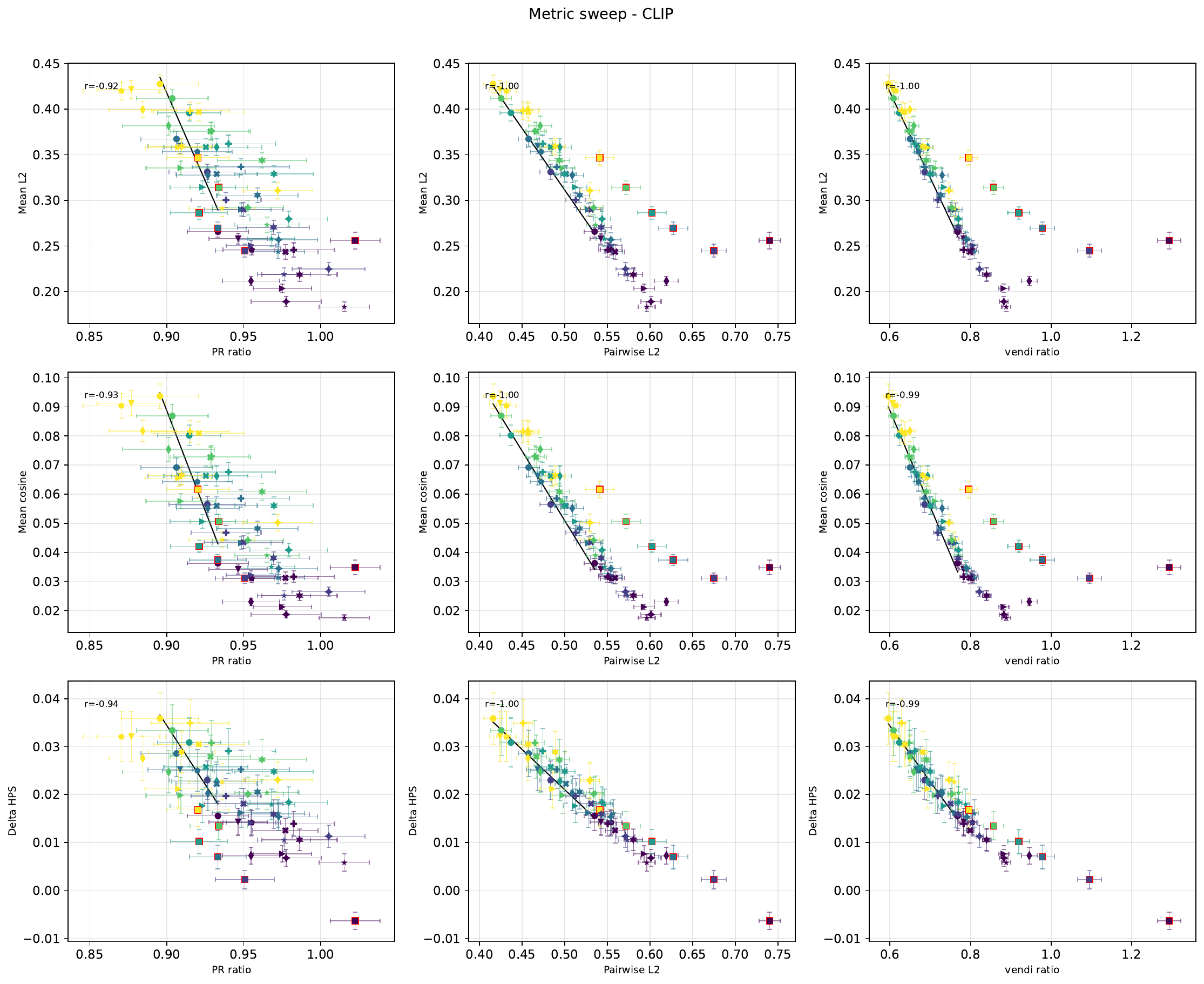}
    \caption{Evaluation of different pairs of metrics for mean and variance distortion on several guidance schedules, features extracted with encoder CLIP. The black line corresponds to the fitted curve for a fixed guidance schedule. Round marks correspond to vanilla CFG schedules, polygons to negative window schedule, triangles to early-high schedules, star to early-low. The markers are colored according to the reference guidance levels reported in Fig.~\ref{fig:schedules}.}
    \label{fig:schedule-sweep-clip}
\end{figure}

\begin{figure}
    \centering
     \includegraphics[width=\linewidth]{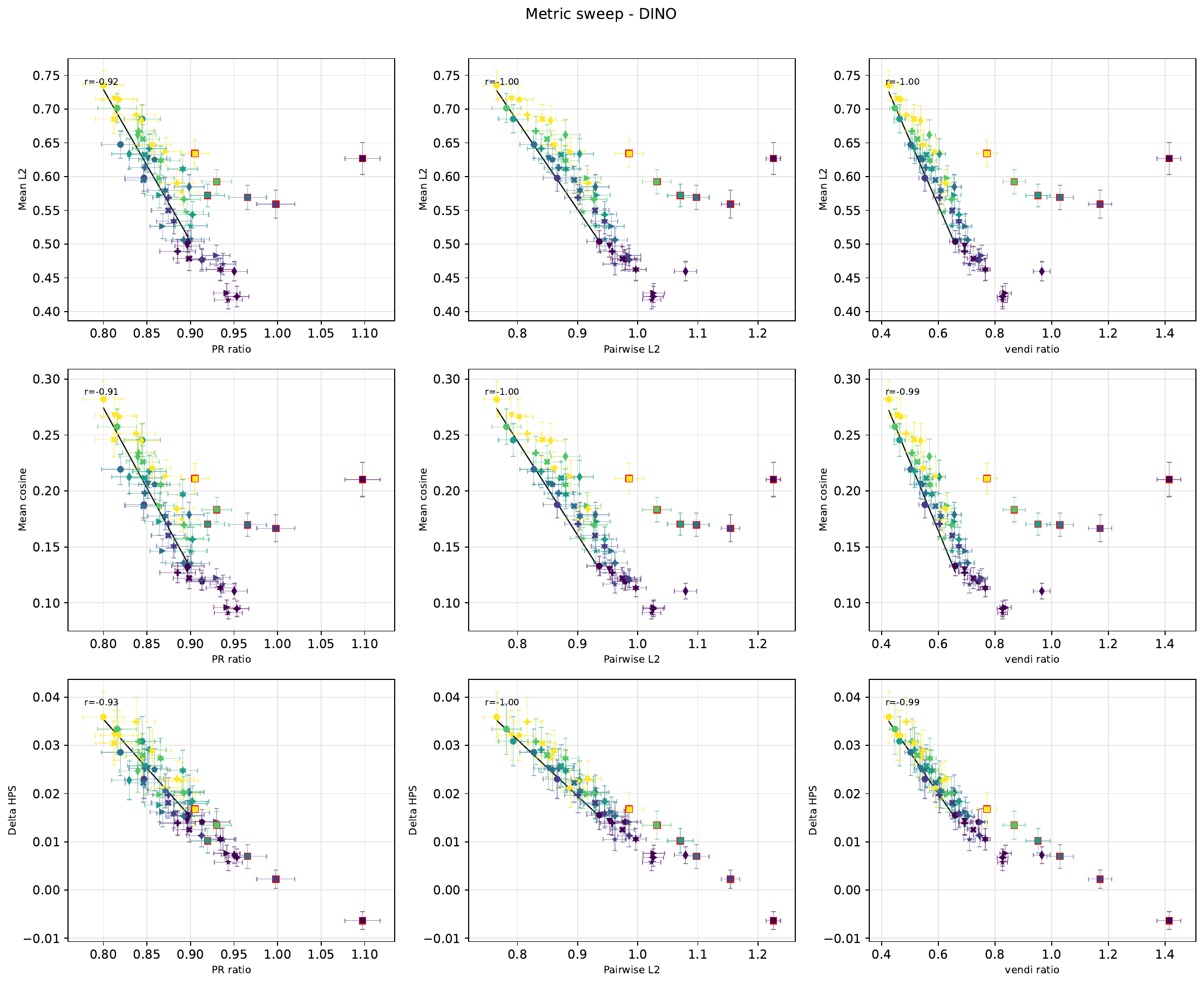}
    \caption{Evaluation of different pairs of metrics for mean and variance distortion on several guidance schedules, features extracted with encoder DINOv2. The black line corresponds to the fitted curve for a fixed guidance schedule. Round marks correspond to vanilla CFG schedules, polygons to negative window schedule, triangles to early-high schedules, star to early-low. The markers are colored according to the reference guidance levels reported in Fig.~\ref{fig:schedules}.}
    \label{fig:schedule-sweep-dino}
\end{figure}

\begin{figure}
    \centering
    \includegraphics[width=0.9\linewidth]{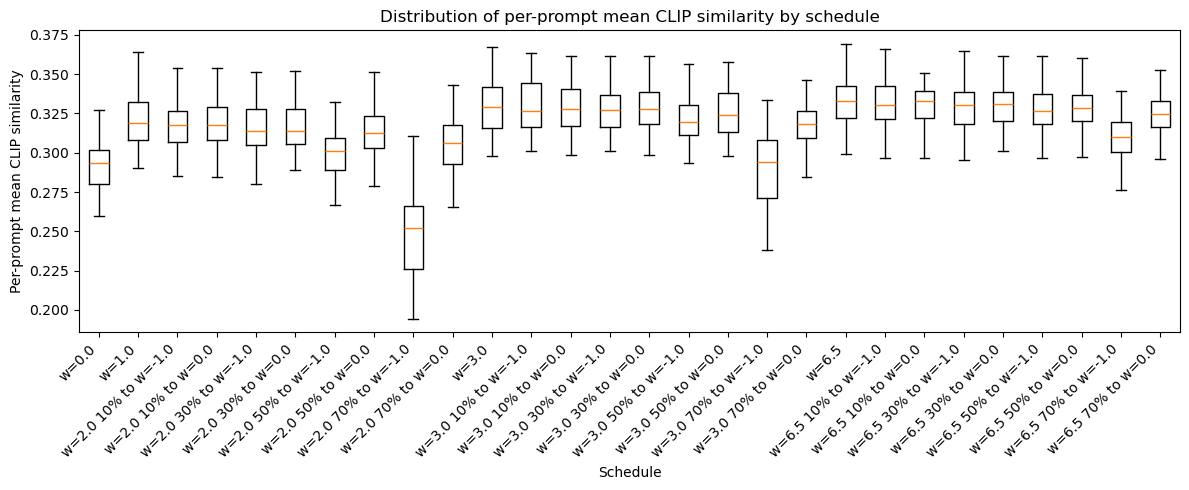}
    \caption{CLIP scores across several standard CFG schedules, quantifying the alignment of the data with the prompt representation. As predicted by the theory, a large gain in diversity, observed for larger negative windows, sacrifices the mean shift, i.e. prompt-alignment.}
    \label{fig:clip_scores}
\end{figure}

\begin{figure}
    \centering
    \includegraphics[width=0.9\linewidth]{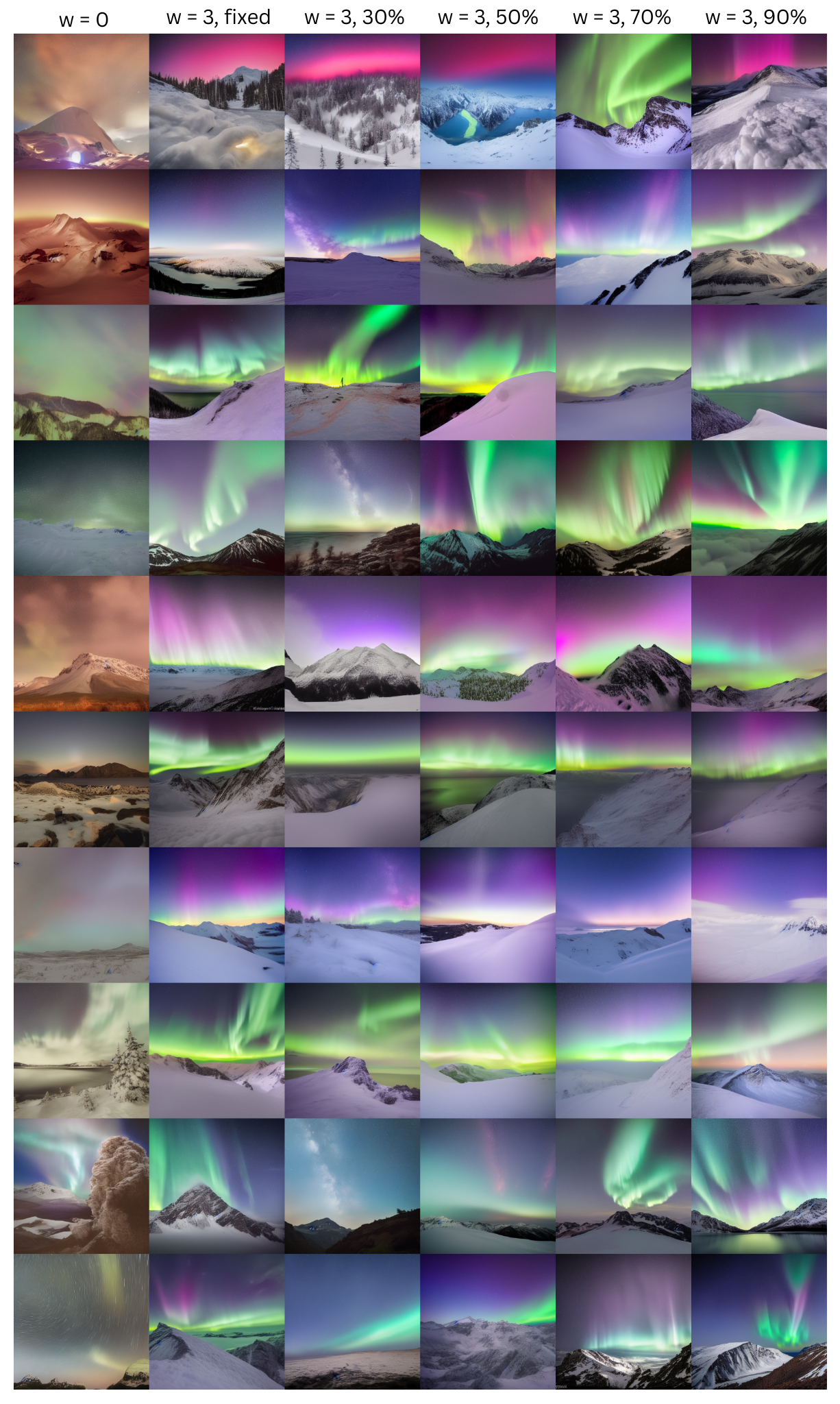}
    \caption{Comparison between images generated at $w = 0$ (conditional), fixed $w = 3$ (vanilla CFG), and negative-window guidance with different window widths, measured in diffusion time percentage. Intermediate windows allow to increase prompt alignment and the diversity of the samples. Prompt: \textit{a snowy mountain peak under a starry sky, northern lights, long exposure}.}
    \label{fig:auroras}
\end{figure}

\begin{figure}
    \centering
    \includegraphics[width=0.9\linewidth]{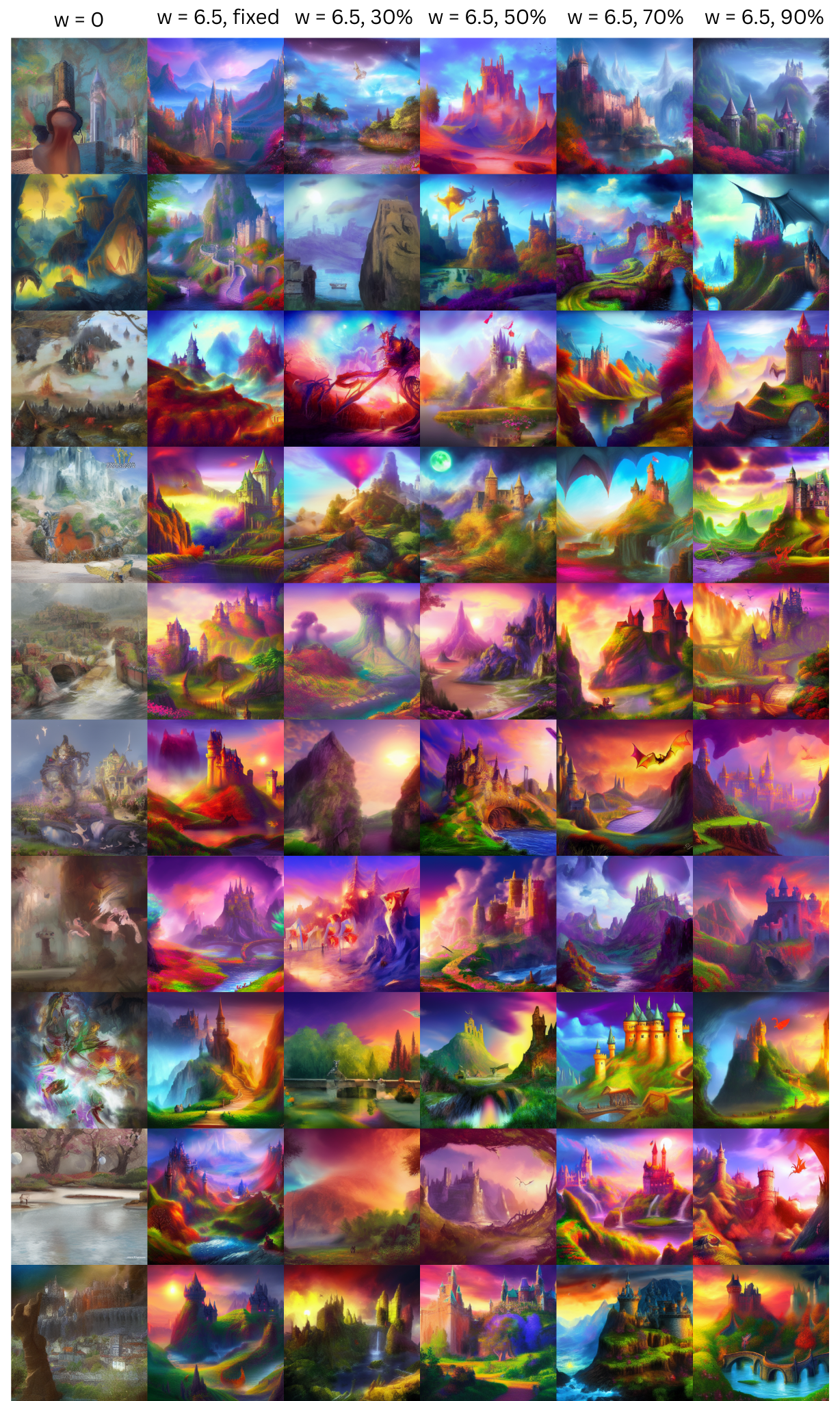}
    \caption{Comparison between images generated at $w = 0$ (conditional), fixed $w = 6.5$ (vanilla CFG), and negative-window guidance with different window widths, measured in diffusion time percentage. Intermediate windows allow to increase prompt alignment and the diversity of the samples. Prompt: \textit{a fantasy landscape with castles and dragons, vibrant colors, digital art}.}
    \label{fig:castles}
\end{figure}

For each schedule, we use Stable Diffusion (v1.5)  to generate a dataset consisting of 20 different samples of the same 50 prompts used for the first batch of experiments described in Sections \ref{app:add_exp} and \ref{sec:sign}. We extract features using CLIP and DINOv2, and measure different metrics of mean and variance distortion as described in Sec.~\ref{app:metrics}. Plotting these measures of distortions on the same plane results in Figures ~\ref{fig:schedule-sweep-clip} and \ref{fig:schedule-sweep-dino}. The black line corresponds to the fitted curve for a fixed guidance schedule. This curve partitions the plane into two indicative regions:
\begin{itemize}
    \item Below the curve: lower mean separation and diversity with respect to vanilla CFG.
    \item Above the curve: higher mean separation and diversity with respect to vanilla CFG.
\end{itemize}
One can observe that while most of the schedules are positioned very close to the solid black line, the ones indicated by a square mark and highlighted in red, which correspond to our early high schedule with a negative guidance window for 70\% of the steps, are positioned far in the region where we diversity is increased with respect to vanilla CFG. At the same time, they show a degree of class separation that is comparable with the other schedules having the same reference guidance level.  
Figure \ref{fig:clip_scores} report CLIP scores of different guidance schedules, showing that the schedules reaching highest diversity (e.g. negative-window with $70 \%$ width) must also sacrifice a bit of prompt-alignment, as it also results from the theory. 
Finally, Figures \ref{fig:auroras} and \ref{fig:castles} show real images generated from the negative-window schedule, for visualization purposes. 

\end{document}